\documentclass{article}

\usepackage[preprint]{neurips_2023}

\usepackage[utf8]{inputenc} 
\usepackage[T1]{fontenc}    
\usepackage{hyperref}       
\usepackage{url}            
\usepackage{booktabs}       
\usepackage{amsfonts}       
\usepackage{nicefrac}       
\usepackage{microtype}      
\usepackage{xcolor}         

\usepackage[english]{babel}
\usepackage{amsmath}
\usepackage{graphicx}
\usepackage{amssymb}
\usepackage{amsthm}
\usepackage{mathrsfs}
\usepackage{longtable}

\usepackage{booktabs}
\usepackage[load-configurations=version-1]{siunitx} 
\usepackage[normalem]{ulem}

\newcommand{\kl}{D_{KL}}
\newcommand{\E}{{\mathbb E}}
\newcommand{\R}{\mathbb{R}}
\renewcommand{\S}{\mathbb{S}}

\newcommand{\F}{\mathcal{F}}

\newtheorem{thm}{Theorem}[section]

\newtheorem{defn}[thm]{Definition}

\newtheorem{rmk}[thm]{Remark}

\newcommand{\ie }{\emph{i.e.}, }

\title{Probabilistic U-Net with Kendall Shape Spaces \\
for Geometry-Aware Segmentations of Images}

\author{%
  Jiyoung Park\\
  Texas A\&M University\\
  \texttt{wldyddl5510@tamu.edu} \\
  \And
  Günay Doğan \\
  National Institute of Standards and Technology \\
  \texttt{gunay.dogan@nist.gov} \\
}

\begin{document}

\maketitle

\begin{abstract}
  One of the fundamental problems in computer vision is image segmentation, the task of detecting distinct regions or objects in given images. Deep Neural Networks (DNN) have been shown to be very effective in segmenting challenging images, producing convincing segmentations. There is further need for probabilistic DNNs that can reflect the uncertainties from the input images and the models into the computed segmentations, in other words, new DNNs that can generate multiple plausible segmentations and their distributions depending on the input or the model uncertainties. While there are existing probabilistic segmentation models, many of them do not take into account the geometry or shape underlying the segmented regions. In this paper, we propose a probabilistic image segmentation model that can incorporate the geometry of a segmentation. Our proposed model builds on the Probabilistic U-Net of \cite{kohl2018probabilistic} to generate probabilistic segmentations, i.e.\! multiple likely segmentations for an input image. Our model also adopts the Kendall Shape Variational Auto-Encoder of \cite{vadgama2023kendall} to encode a Kendall shape space in the latent variable layers of the prior and posterior networks of the Probabilistic U-Net. Incorporating the shape space in this manner leads to a more robust segmentation with spatially coherent regions, respecting the underlying geometry in the input images.
\end{abstract}

\section{Introduction} \label{sec_intro}

Image segmentation, the task of detecting regions or objects in given images, is a fundamental problem in computer vision. Many different mathematical and algorithmic solutions have been proposed for this problem. Recently algorithms using deep neural networks (DNN) have been used successfully to segment images in very challenging scenarios \cite{ghosh2019understanding,minaee2021image}. In most applications, DNN models are applied to problems, such as image segmentation, in a deterministic manner. A single output (segmentation) is produced for a given input (image), without consideration of how the uncertainties in the input data and the DNN model affect the output, nor of what is a reasonable distribution of possible outputs based on these uncertainties. From a machine learning point of view, this is in contrast with some other conventional algorithms, for example logistic regression, which, when used for binary classification, can provide the exact decision boundary and the variance of the prediction under suitable assumptions. DNN models, in their most commonly used form, do not provide such information due to their black box nature and the massive number of parameters in the model. Nonetheless, in many segmentation scenarios, it is helpful and desirable to obtain multiple likely segmentations and quantify their distribution, depending on the uncertainties in the input data and the segmentation model. For example, a radiologist can produce multiple plausible tumor segmentations based on the limited amount of information available in a given CT scan. An automated system can analyze a range of segmentations and their likelihoods given the image of an outdoor scene that may include occlusions like smoke, rain, snow, and base its decisions on these likely segmentations. Thus probabilistic segmentation models that can generate multiple likely segmentations instead of a single deterministic segmentation are desirable in many image analysis scenarios. Such models can inform and enable uncertainty quantification and statistical inference, especially analyses using Monte Carlo sampling of the outcomes.

Despite the need for probabilistic segmentation models in applications, availability of such models have been limited. One natural approach to develop such methods is to adopt a suitable classification algorithm, including DNNs, and to compute pixel-wise probabilities, which can be used to estimate pixel-wise uncertainties in segmentations \cite{kendall2015bayesian,kendall2017uncertainties}. This approach neglects the spatial correlations between pixels and does not conform to the expected spatial coherency of a segmentation. Graphical models, e.g.\! junction chains, Markov Random Fields, have also been used to produce multiple segmentations from one image \cite{batra2012diverse,kirillov2015inferring,kirillov2015m,kirillov2016joint}, but these are more suitable for structured problems with dependencies that can be encoded in graphical models. DNNs, in contrast, are flexible and have broad representation capability. To enable multiple segmentations of images, ensembles of DNNs \cite{lakshminarayanan2017simple,guzman2012multiple,lee2015m} and multi-head models (i.e.\! one common neural network with M heads \cite{rupprecht2017learning,ilg2018uncertainty}) have been used. These models do not scale well to large numbers of segmentation hypotheses.

One essential point that we should emphasize is that a segmentation is not merely a collection of labeled pixels. A segmentation represents a detected object or region in an image. Hence, it is a spatially coherent region, in which the pixels are naturally correlated. The segmented region has a shape and topology. Segmentations of similar objects are expected to have similar shapes and topologies. A probabilistic model of segmentation should also be built on this principle. To our knowledge, only recently a probabilistic segmentation model that considers topology was proposed in \cite{hu2022learning}. However, their approach takes a one dimensional parameter space to encode the topology, and is hard to interpret and relate to the structure of the segmentation directly. In this work, we aim at the intrinsic geometry of the regions in the image, and focus on preserving the geometric information instead of topological information. In order to obtain a probabilistic segmentation model that encodes and respects the geometry in data, we build on the Probabilistic U-Net of \cite{kohl2018probabilistic}, and enhance it with Kendall Shape Variational Auto-Encoder (VAE) of \cite{vadgama2023kendall} to build our Kendall Shape Probabilistic U-Net. This new probabilistic segmentation model is the main contribution of our paper. Kendall Shape Probabilistic U-Net incorporates shape representations in its latent space to capture and represent the inherent geometry of the regions. This results in geometrically more coherent and more plausible segmentations generated for a given image.

\section{Main Components of Our Approach}\label{sec_prelim}

\subsection{Probabilistic U-Net}\label{sec_prob_unet}

U-Net \cite{ronneberger2015unet} is a widely used deep neural network (DNN) for image segmentation. U-Net approaches the image segmentation task as a two-stage problem: 1. Extract features from the image (Contraction), and 2. Recover the segmentation from the extracted feature (Expansion). In Contraction stage, U-Net takes an input image and passes it to convolutional neural network (CNN) layers to compress the image into features. Then, Expansion stage reverses this operation and reconstructs the image into segmentations, output as an image with assigned class labels as pixels. In Expansion stage, at every layer, instead of passing only features from the resulting vectors of the previous layer, we concatenate the resulting vectors of the corresponding Contraction stage with the previous layer’s resulting vectors and use this concatenated vector as an input for the next layer. By this, we are able to utilize the features from each downsampling stage when conducting upsampling stages. Lastly, we minimize the cross-entropy loss between predicted segmentations from U-Net with ground truth segmentations.


Probabilistic U-Net is a variant of U-Net, which has the capability to generate probabilistic segmentations \cite{kohl2018probabilistic}. The idea of Probabilistic U-Net is to append an additional neural network architecture independent of U-Net to inject the stochasticity. This additional neural network is implemented using Variational Autoencoder (VAE) \cite{kingma2014autoencoding} and is  called the prior network. In Expansion stage of U-Net, before the very last layer where we obtain the segmentation, we concatenate the resulting vector with a sample from the prior network, which results in a sample segmentation.

To train the latent space of VAE, Probabilistic U-Net appends one more auxiliary network during training, called posterior network. The architecture of prior network and posterior network are almost the same, except that the prior network takes only the input image as an input and encodes the input into latent space, while the posterior network takes both the input image and the ground truth segmentation as input and encodes both sets of information. Then, a Kullback-Leibler divergence (KL divergence) between the prior distribution and the posterior distribution is added to the original cross entropy loss of U-Net, to simultaneously train the latent space of the VAE network.

\subsection{Kendall Shape Space} \label{sec_kendall}

A Kendall shape space is a framework that originated from the field of shape analysis to define a space of shapes and to compute shape distances \cite{kendall1984shape}. In  Kendall shape space, two point clouds (or landmarks) are considered to be the same shape if one point cloud can be transformed into the other via rotations, translations, and/or scaling. Mathematically, it is defined as a space of normalized point clouds modulo rotation. 
\begin{defn}[Kendall shape space] \label{defn_kendall}
A Kendall shape space with $k$ landmarks $\in \R^m$ is denoted by $\Sigma_{m}^{k} = S_{m}^{k} / SO(m)$, where $S_{m}^{k} := \{X \in \mathbb{R}^{m \times k} \: \| \: \sum_{i = 1}^{k} X_i = 0, \: \|X\|_{F} = 1\}$ is called a `pre-shape' space.
\end{defn}
Since our goal is a probabilistic model incorporating geometries of segmentations, we focus on distributions in a Kendall shape space. One useful fact for this is that a pre-shape space $S^{k}_{m}$ can be identified with $\S^{(k-1)m - 1}$, a unit hypersphere in $\R^{(k-1)m}$. The identification between the two is done by a transformation $\psi: S_{m}^{k} \rightarrow \S^{(k-1)m - 1}$ defined by
\begin{align}\label{eq_preshape_sphere}
    \psi(X) = \frac{HX}{\|HX\|},
\end{align}
where $H$ is a Helmert sub-matrix of $(m-1) \times m$. This fact is crucial when we construct a distribution in a Kendall shape space, which we will describe in the next section.

\subsection{Von Mises-Fisher Distribution for Shapes}\label{sec_vmf}

Using the identification \eqref{eq_preshape_sphere} between $S^{k}_{m}$ and $\S^{(k-1)m - 1}$ mentioned above, a distribution in pre-shape space can be also constructed in terms of distributions on a sphere. Since Kendall shape space is a quotient space of pre-shape space, a distribution in Kendall shape space can be regarded as a distribution on the sphere. The most widely used distribution on sphere is von Mises-Fisher (vMF) distribution, which is analogous to a Gaussian distribution in Euclidean space. A vMF distribution on $\S^{d-1}$ consists of two parameters, a mean direction $\mu \in \S^{d-1}$, a concentration parameter $\kappa \in \R_{+}$. Its probability density function is defined by
\begin{align*}
    {} \qquad
    f(X;\mu, \kappa) = \frac{\kappa^{\frac{d}{2} - 1}}{(2\pi)^{\frac{d}{2}}I_{\frac{d}{2}-1}(\kappa)} exp(\kappa \mu^{T}X) 
    \ \text{ where } \ 
    I_{\alpha}(z) = \sum_{m=0}^{\infty} \frac{(z/2)^{2m + \alpha}}{m!\Gamma(m+\alpha+1)}.
\end{align*}
As discussed in the Section \ref{sec_prob_unet}, one of the key quantities that we need to train Probabilistic U-Net is the KL divergence between the prior distribution and the posterior distribution, which in our case will be two von Mises-Fisher distributions. While there is no closed form expression for KL divergence between two von Mises-Fisher distributions, \cite{diethe2015kl}[Theorem 3.1] shows that there exists an upper bound for such quantity, which can be computed very efficiently. 
%
\begin{thm}[KL divergence between von Mises-Fisher distributions]\label{thm_kl_vmf}
The KL divergence between two von Mises-Fisher $vMF(\mu_0, \kappa_0), vMF(\mu_1, \kappa_1)$ distributions on $\S^{d - 1}$ where $d$ is odd (with $d^{\bullet} = \frac{d-1}{2}$, $d^{\diamond} = d^{\bullet} - 1$) is bounded by the following quantity:
    \begin{align*}
        \kl \leq \kappa_0 - \kappa_1 \mu_1^{T}\mu_0
        + d^{\bullet} \log(\kappa_0) 
        + \sum_{i = 0}^{d^{\diamond}} \frac{\kappa_{0}^{i}}{i!} 
        - d^{\bullet} \log \kappa_1
        + d^{\bullet} d^{\diamond} \log d^{\diamond} - d^{\diamond 2} + 1.
    \end{align*}
\end{thm}
\begin{rmk}
When $d$ is even, just adding one null dimension and considering $d + 1$ as the new dimension will induce the upper bound. 
\end{rmk}
Now, we are ready to construct a distribution in Kendall shape space that corresponds to a vMF distribution on a sphere. Note that we need to have a mean direction $\mu$ and a concentration parameter $\kappa$. Therefore, an analogous distribution in $\Sigma_{m}^{k}$ should also consist of the mean shape $M_0$ and the concentration $\kappa$. Suppose $P(M_0, \kappa)$ is a probability distribution in Kendall shape space with parameters $(M_0, \kappa) \in \Sigma_{m}^{k} \times \R_{+}$. This is analogous formulation to vMF distribution, where $M_0$ stands for the mean shape and $\kappa$ denotes the concentration. Then, in fact we can regard $M_0$ as a standardized (with respect to one fixed orientation) element in $S_{m}^{k}$, \ie $M_0 = R M$ for some $R \in SO(m)$ and $M \in S_{m}^{k}$. Therefore, if we take the transformation in \eqref{eq_preshape_sphere} by $\psi(M_0) = \mu_0$, we can identify $P(M_0, \kappa)$ with $vMF(\mu_0, \kappa)$. We will use this $vMF(\mu_0, \kappa)$ as $P(M_0, \kappa)$, a distribution in Kendall shape space. 

\begin{rmk}\label{rmk_offset_normal}
We note that there is another (perhaps more widely used) way to model the distribution in a Kendall shape space. In statistical shape analysis, an Offset Normal Shape distribution (See \cite{Fontanella2018offset}[Section 2] for more details) is considered to be a distribution in Kendall shape space analogous to a normal distribution in Euclidean space. However, we found our approach computationally more tractable, as parameters of an Offset Normal Shape Distribution need to bring more complicated equivalence classes (see \cite{welling2010offset}[Section 2] for more details). On the other hand, our distribution can be identified to the vMF distribution directly, so that many computations (e.g. KL divergence) become more tractable. That said, an Offset Normal Shape Distribution has an advantage in that this distribution is more flexible in terms of the variance. Therefore, extending our methods to Offset Normal Shape distributions might be an interesting future direction to explore. 
\end{rmk}


\subsection{Equivariant Neural Networks}


A main takeaway from Section \ref{sec_vmf} is that a vMF distribution combined with rotational standardization can play the role of a distribution in the Kendall shape space. As a result, in order to implement the distribution in the Kendall shape space via a neural network, we need to have a neural network that enables us to perform the rotational standardization operations in the latent space domain. One of the way to accomplish this is to use a Group equivariant steerable CNN \cite{cohen2017steerable}.

\subsubsection{Group Equivariant Steerable CNN}

Group-equivariant steerable (G-steerable) CNN \cite{cohen2017steerable} is a framework to implement the equivariant neural network. G-steerable CNNs approach the problem by constructing equivariant filters and steerable feature spaces based on representation theory.
Consider $\F_i :=\{f\:|\: f: \R^{c_{i-1}} \rightarrow \R^{c_{i}}\}$, a set of feature maps from $(i-1)^{th}$ layer of the CNN to the $i$th layer. Let $\rho_i: G \rightarrow GL(\R^{c_i})$ and $\pi_i: G \rightarrow GL(\F_i)$ be group representations satisfying the  identity: 
\[
(\pi_i(g)f)(x) = \rho_i(g)f(g^{-1}x), \qquad \forall f \in \F_i, g \in G,
\]
which is true when $f$ is a vector field and $G = SO(c_i)$ (see \cite{weiler2019e2}[Fig.1]) 
A single convolution layer can be viewed as a mapping from the input feature map $\F_{in}$ to the output feature map $\F_{out}$, i.e. $\Phi: \F_{in} \rightarrow \F_{out}$ such that $\Phi(f) = f \star k$ where $\star$ is a convolution. Then, a steerable feature is defined as:

\begin{defn}[Steerable feature]\label{defn_steerable_filter}
For a given group representation $\pi_{in}: G \rightarrow GL(\F_{in})$, a convolution operator $\Phi: \F_{in} \rightarrow \F_{out}$ is said to be steerable if there exist a group representation $\pi_{out}: G \rightarrow GL(\F_{out})$ such that 
\[
\Phi(\pi_{in}(g)f) = \pi_{out}(g) \Phi(f), \qquad \forall f \in \F_{in}, g \in G.
\]
\end{defn}
This definition means that we can obtain the group representation in the output domain corresponding to that in the input domain. 
Now, we want to find a kernel $k$ inducing the steerable feature $\Phi$. Such a kernel needs to satisfy the following constraint:
\begin{align}\label{eq_steer_constraint}
k(gx) = \rho_{out}(g) k(x) \rho_{in}(g)^{-1}, \quad
\mathrm{for} \ x \in \R^c, g \in G.
\end{align}
This is due to the following computation: Under the above constraint,
\begin{align*}
\Phi(\pi_{in}(g)f)(x) 
&= (k \star \pi_{in}(g)f)(x) = \int_{X} k(y) \rho_{in}(g)f(g^{-1}(x - y))dy \\
&= \int_{X} \rho_{out}(g) k(g^{-1}y) \rho_{in}(g)^{-1}\rho_{in}(g)f(x- y)dy \\
&= \rho_{out}(g) \int_{X} k(g^{-1}y)f(g^{-1}x- g^{-1}y)dy
= \rho_{out}(g) (k \star f)(g^{-1}x) = \pi_{out}(g) \Phi(f)(x)
\end{align*}
meaning $\Phi$ is a steerable feature.

Since a kernel $k$ is a map from $\R^{c_{i-1}}$ to $\R^{c_i}$, it can be spanned by $\{k_j\}_{j = 1, \dots c_i c_{i-1}}$, a basis of linear maps from $\R^{c_{i-1}}$ to $\R^{c_i}$, \ie $k(x) = \sum_{j = 1}^{c_i c_{i-1}} w_j k_j(x)$. Then, since the constraint \eqref{eq_steer_constraint} is linear in $k$, the $G$-equivariant steerable kernel $k_G$ will be in a linear subspace of those linear maps. In particular, if we fix $\rho$ as an irreducible representation, we can choose the corresponding irreducible basis $\{k_j\}_{j = 1, \dots c_i c_{i-1}}$ as well. Using this fact, we are able to write equation \eqref{eq_steer_constraint} in the basis form to get the following:
\begin{align}\label{eq_steer_constraint_element}
    k_{a,b}(gx) = \rho_{a}(g) k_{a,b}(x) \rho_{b}(g)^{-1}
\end{align}
for all $a = 1, \dots, c_{i-1}, b = 1, \dots, c_i, g \in G$, and $x \in \R^{c_{i-1}}$. 
Note constraint \eqref{eq_steer_constraint_element} depends only on the choice of the group and the number of channels; it is independent of data. Therefore, we can obtain a pre-defined basis if we fix the group and the channels. To learn the $G$-equivariant steerable CNNs, we first calculate the basis satisfying \eqref{eq_steer_constraint_element} for given $G$ and channels. Then during training, we learn the weights $\{w_j\}$ to obtain the desired equivariant kernel. 

\subsection{Kendall Shape Variational Auto-Encoder (VAE)} \label{sec_kendall_vae}

Kendall Shape VAE \cite{vadgama2023kendall} is a variant of VAE, for which the latent space is designed to be a distribution on a Kendall shape space. It is implemented as follows: First, we preprocess the input image to have a fixed scale and center. By this, we obtain translation and scaling invariance. Now the problem is to obtain rotation invariance. To this end, instead of using plain CNNs in the VAE architecture, $SO(m)$-equivariant steerable CNNs \cite{weiler2018learning} are used. Under the $SO(m)$-equivariance, if we provide rotated input, we will obtain rotated output. As we want our latent space to be a Kendall shape space, we need for it to be invariant to rotation. For this, we make our neural network determine the orientation of the given input. Then, we act the inverse of this rotation matrix to our output of $SO(m)$-equivariant steerable CNNs. Since the output is from $SO(m)$-equivariant steerable CNNs, taking the inverse of the rotation corresponds to taking the inverse rotation of the input image as well. Therefore, by these procedures, each vector of our latent spaces will be a representative of one single shape. To model the prior and the posterior distribution, the approach in Hyperspherical VAE \cite{davidson2018svae} was used: the prior distribution is modeled by a uniform distribution on a sphere and the posterior distribution by the vMF distribution. Then KL divergence becomes feasible to compute.

\section{Proposed Algorithm: Kendall Shape Probabilistic U-Net}

In this section, we describe our novel Kendall Shape Probabilistic U-Net, which incorporates geometric information into Probabilistic U-Net. Our motivation for this model comes from the following idea: ``Segmentation samples of same object should share the same shape". The key idea in our approach is to replace the latent space of Probabilistic U-Net by Kendall Shape VAE. Kendall Shape Probabilistic U-Net encodes the geometry of the input image’s object into the latent space, which is a Kendall shape space. Then, we can interpret output samples from latent space as samples of different shapes from the distribution of shapes centered on the input region’s shape. Therefore, we expect the output segmentation samples as well as the latent space to remain close to the inherent shape of the input regions. Consequently, in contrast to Probabilistic U-Net, we expect to obtain more robust segmentation samples per image, with better preservation of the geometry. 

\begin{figure}[h]
    \centering
    \includegraphics[scale=0.4]{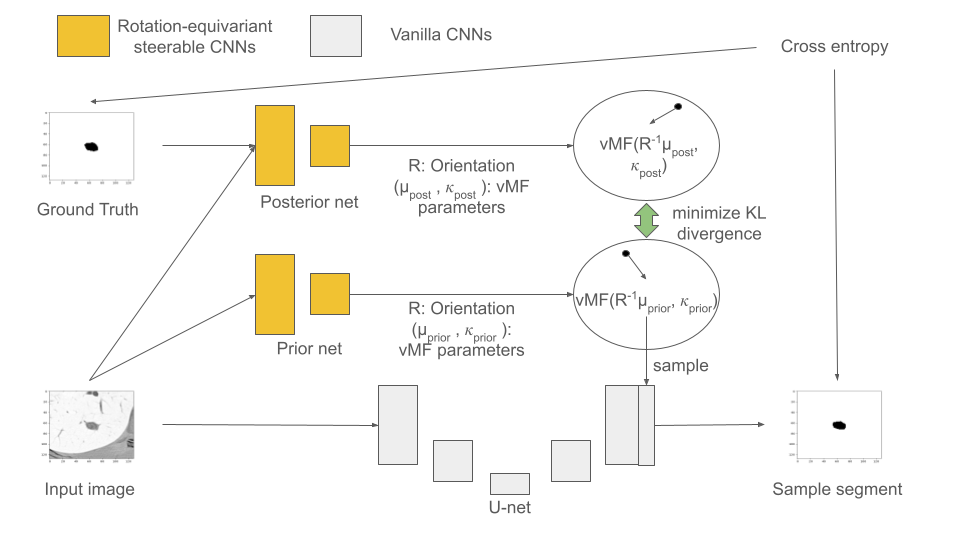}
    \caption{Architecture for Kendall Shape Probabilistic U-Net, consisting of U-Net, Prior Network, and Posterior Network. U-Net is the same as the one in Probabilistic U-Net. Prior Network takes an input image, and returns its orientation and the parameters of vMF distribution. Posterior Network works same, but also takes a ground-truth segmentation. Grey and yellow boxes are vanilla CNN and $SO(m)$-equivariant steerable CNNs layers respectively. 
    }
\label{fig_nn_figure}
\end{figure}


An overview of our DNN architecture is given in Fig.\ref{fig_nn_figure}. As in Probabilistic U-Net, we have three networks: Prior Network, Posterior Network, and U-Net. Prior Network and Posterior Network are implemented using Kendall Shape VAE: We pass the input image $X$ (also ground-truth segmentation if it is Posterior Network) into $SO(m)$-equivariant steerable CNNs. We construct our equivariant steerable CNNs to return $R \in SO(m), M \in \R^{m \times k}$, and $\kappa \in \R_{+}$. We then standardize $M$ (with respect to scale and mean) to be in $S_{m}^{k}$. $R$ will encode the information of current orientation, $M$ will serve as a mean pre-shape, and $\kappa$ will work as the concentration parameter. Once we have $M, \kappa$ and $R$, we make a representative shape by standardizing the orientation by $M_0 = R^{-1}M \in \Sigma_{m}^{k}$. Lastly, by conducting the transformation \eqref{eq_preshape_sphere}, we obtain $\mu_0 = \psi(M_0)$. Then, $vMF(\mu_0, \kappa)$ serves as our VAE latent space distribution for both prior and posterior distributions. As in probabilistic U-Net, we augment each sample of $vMF_{\text{prior}}(\mu_0, \kappa)$ into U-Net's last layer. For the implementation of $SO(m)$-equivariant steerable CNNs, we used the ESCNN package \cite{cesa2022escnn}, and for the rest of the architectures, we used probabilistic U-Net's architectures. 

\begin{rmk}\label{rmk_diff_btwn_kendall_vae}
    We emphasize one difference between the  Kendall VAE framework (Section \ref{sec_kendall_vae}) and our model. In VAE settings, prior distributions are only auxiliary distributions for training, and posterior distributions are the distributions that are used to reproduce the samples in practice. Therefore, it is possible to model the prior distribution as the uniform distribution and the posterior distribution as the vMF distribution in the VAE case. \\
    In contrast, in Probabilistic U-Net, the distribution that we eventually use to sample is a prior distribution as discussed in Section \ref{sec_prob_unet}, since this distribution does not require ground truth segmentations for the input. In this regard, choosing the uniform distribution as a prior is not desirable, as the uniform distribution will reproduce non-informative samples. We circumvent this problem by implementing both prior and posterior distributions to be vMF distributions.
\end{rmk}
We use the following loss function for  image $X$, ground-truth $Y$, and predicted segmentation $S(X,z)$:
\begin{align}\label{eq_exact_loss}
    L(X,Y) &= \E_{z \sim Q(\cdot | X,Y)}[-\log P(Y|S(X,z)] + \beta \kl(Q \| P) + \gamma L_{\text{weight.reg.}}
\end{align}
The first term corresponds to typical segmentation loss implemented as a cross-entropy in a probabilistic model sense. That being said, now latent variable $z$ is an element in a Kendall shape space, unlike the original probabilistic U-Net which takes $z$ in Euclidean space. This setting enables our samples to implicitly train $z$ to learn the underlying geometric structure.

The second term in \eqref{eq_exact_loss} is the KL divergence between the Prior and Posterior Network. As mentioned in Remark \ref{rmk_diff_btwn_kendall_vae}, we have to compute the KL divergence between two vMF distributions $Q$ and $P$. This term can be computed either by the sample estimate or substituted by the upper bound derived in Theorem \ref{thm_kl_vmf}. Note that the posterior network takes a ground truth segmentation as an additional input. Therefore, this term ensures that the prior network encodes the information of the ground-truth segmentation. In particular, the standardized orientation that we take in the prior distribution will coincide with the ground truth segmentation, by minimizing the KL divergence with the posterior distribution.
Lastly, we add suitable canonical weight regularizations for the prior network, posterior network, and U-Net network. These regularizations are aggregated in the last term $L_{\text{weight.reg.}}$ of \eqref{eq_exact_loss}.

\section{Experiments}\label{sec_experi}



We used $k = 4$, $m = 2$ for our Kendall shape space's hyperparameters. To implement $SO(2)$-steerable equivariant CNNs, we used an equivariance among a cyclic group of order 8. We used $\beta = 1$, $\gamma=1$ in Eqn.\! \eqref{eq_exact_loss}. For the rest of our Probabilistic U-Net setup, we maintained the same hyperparameters as those used with the original Probabilistic U-Net.


\begin{figure}[h]
    \centering
    \includegraphics[width=0.49\textwidth]{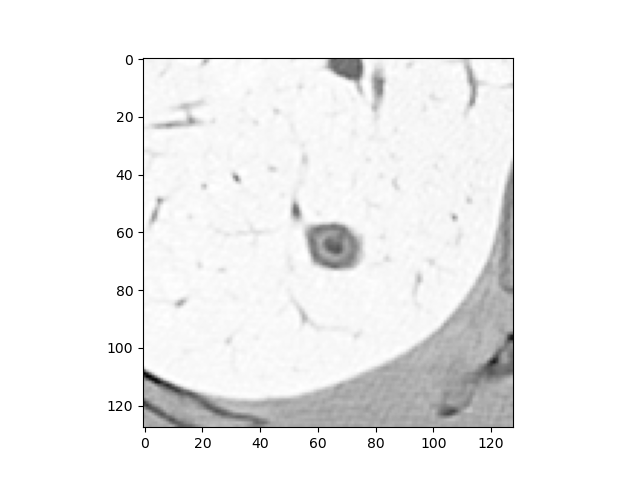}\hfill
    \includegraphics[width=0.49\textwidth]{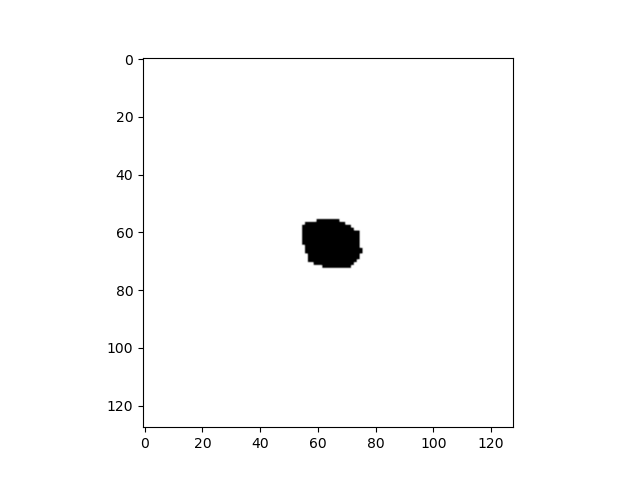}\hfill

    \hrule
    
    \includegraphics[width=0.2\textwidth]{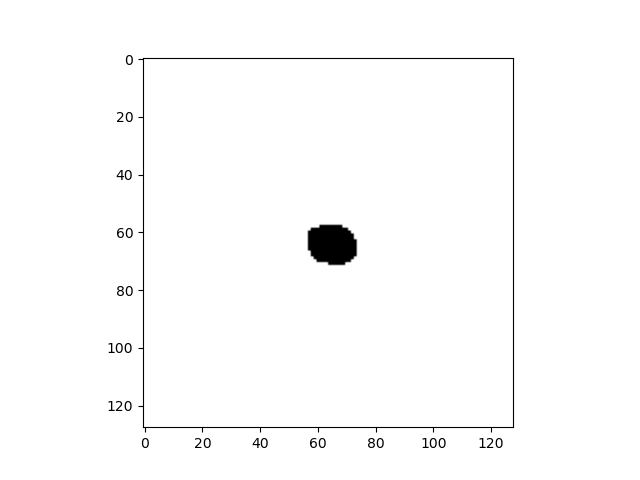}\hfill
    \includegraphics[width=0.2\textwidth]{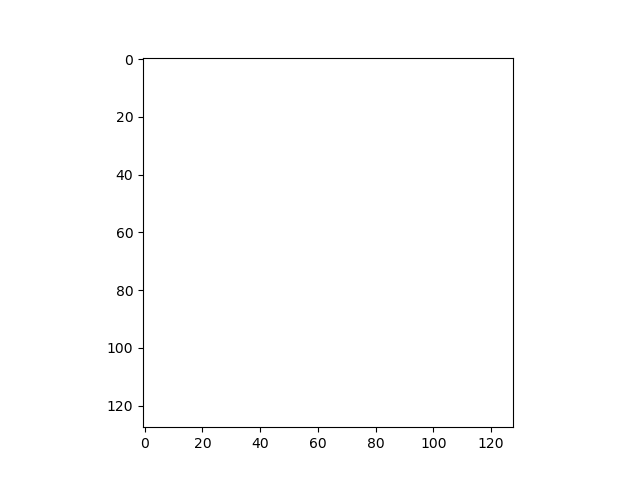}\hfill
    \includegraphics[width=0.2\textwidth]{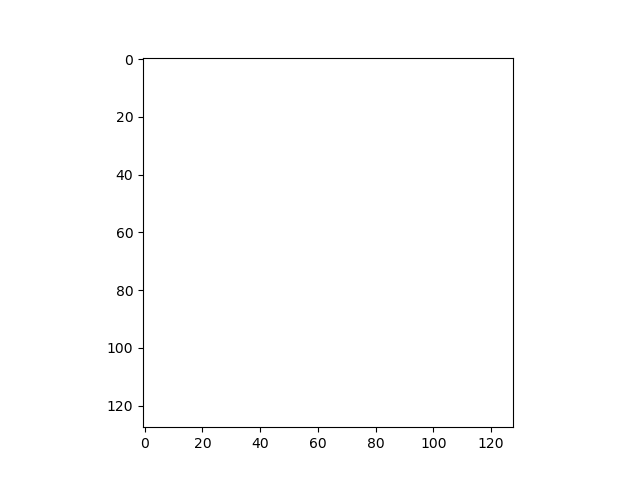}\hfill
    \includegraphics[width=0.2\textwidth]{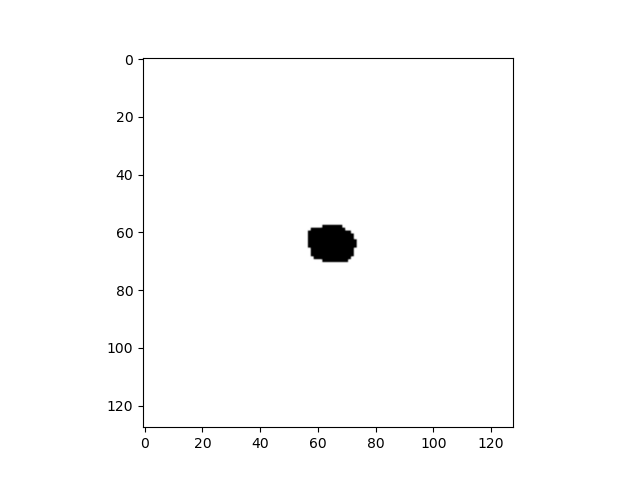}\hfill
    \includegraphics[width=0.2\textwidth]{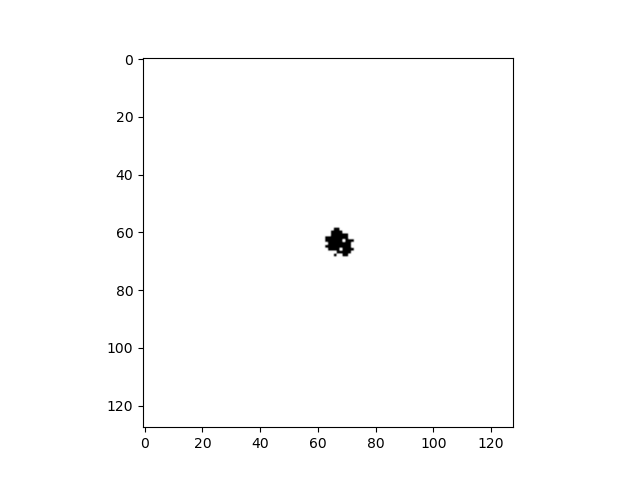}\hfill

    \hrule

    \includegraphics[width=0.2\textwidth]{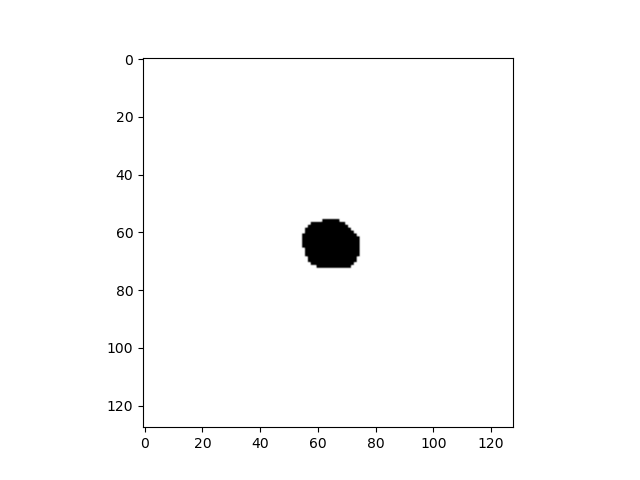}\hfill
    \includegraphics[width=0.2\textwidth]{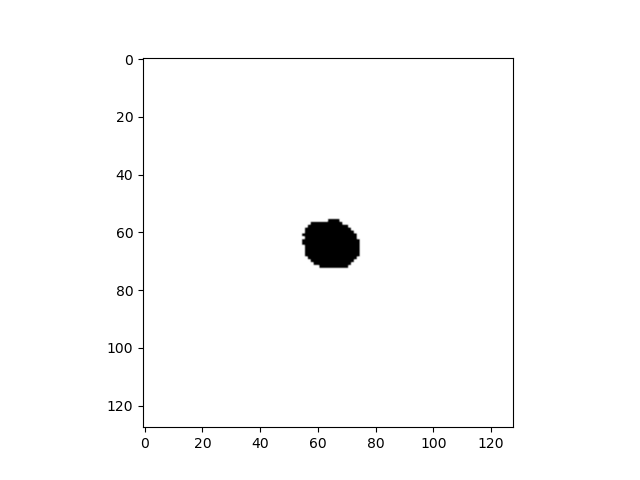}\hfill
    \includegraphics[width=0.2\textwidth]{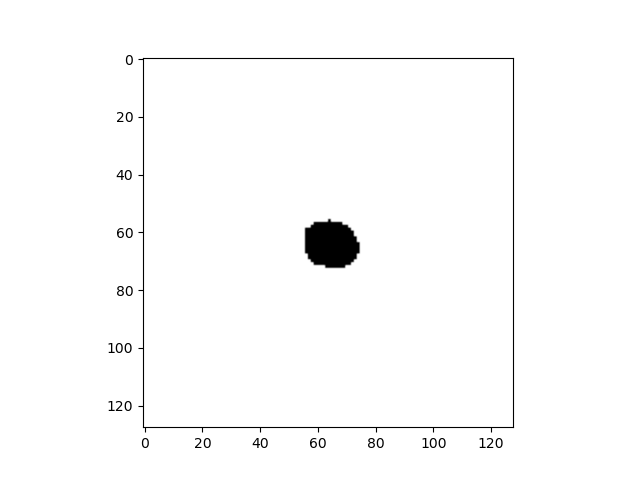}\hfill
    \includegraphics[width=0.2\textwidth]{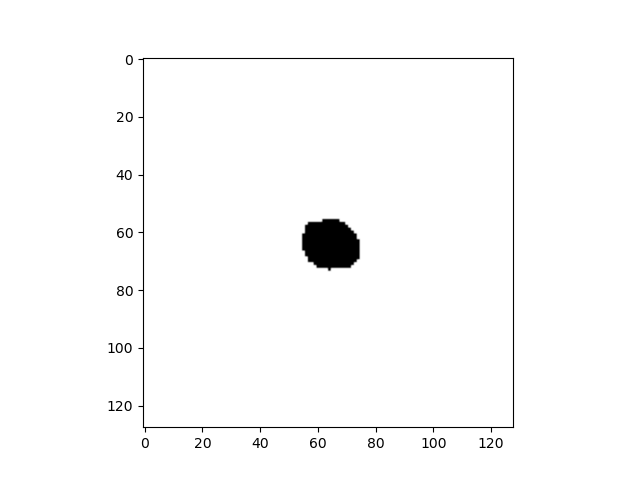}\hfill
    \includegraphics[width=0.2\textwidth]{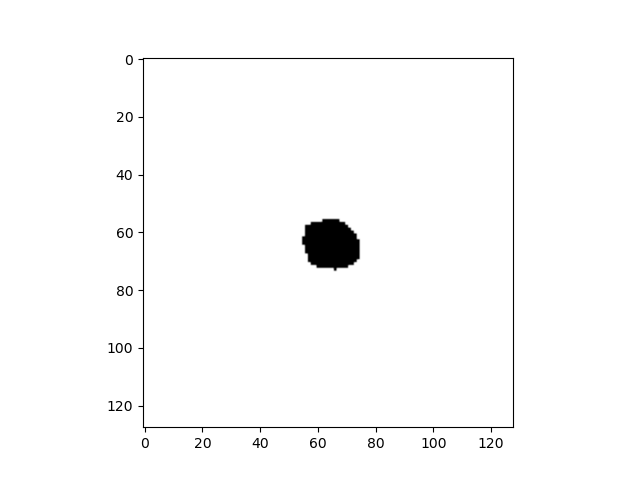}\hfill

    \caption{\textbf{Topleft}: Input image from LIDC data. 
    \textbf{Topright}: Ground truth segmentation.\\ 
    \textbf{Second row}: Segmentation samples from the original Probabilistic U-Net. \\
    \textbf{Third row}: Segmentation samples from Kendall Shape Probabilistic U-Net.}
    \label{fig_kendall_LIDC}
\end{figure}


We conducted the experiments on the LIDC-IDRI dataset \cite{Armato2015lidc}. Figure \ref{fig_kendall_LIDC} shows the resulting sample segmentations from our model and Probabilistic U-Net. 
 In our preliminary experimentation and qualitative assessments, we observed more robust segmentations with better spatial coherency of segmented regions using our model. 
 We aggregated full experimental results, which are in agreement with those in Figure \ref{fig_kendall_LIDC}, in Appendix \ref{appendix_experiments}. 
 In the next stage of our study, we intend to carry out a more detailed quantitative comparison on the LIDC-IDRI and other datasets.

\section{Conclusion}

In this work, we proposed a Kendall Shape Probabilistic U-Net, which is capable of producing probabilistic segmentations maintaining the underlying geometry of the regions. To this end, we modified the original probabilistic U-Net \cite{kohl2018probabilistic} by enforcing the latent space to be a Kendall shape space, a mathematical framework for modelling the shape of the object. This was implemented using the rotation equivariant steerable CNNs. We showed that our approach indeed generates more robust segmentation samples, implying that this model is more faithful to the implicit geometry of the regions in the input image.

We would like to note that our approach can still be extended in multiple directions. One way is to extend the distribution family for the latent space to distributions that can handle more flexibility, such as Offset Normal Shape distributions mentioned in Remark \ref{rmk_offset_normal}. Another direction would be to adopt different frameworks for the shape space. There are multiple ways to characterize the space of shapes, such as a diffeomorphism based  approach or a Persistent Homology Transform based approach. We refer to \cite{arya2024sheaf, bauer2014overview, younes2010shapes} for more details on the mathematical shape modelling. Since each framework has its own perspective and use, using a different geometric shape space as a latent space may have an advantage depending on the task and the goals.

\section*{Acknowledgements}
This work was supported by National Institute of Standards and Technology (NIST) and the Mathematical Sciences Graduate Internship (MSGI) program of the National Science Foundation (NSF).

\newpage

\bibliographystyle{plainnat}
\bibliography{pmlr-sample}

\newpage

\appendix

\section{Experimental Results}\label{appendix_experiments}

In this section, we provide more results on LIDC dataset, which was briefly described in Section \ref{sec_experi}. As mentioned in Section \ref{sec_experi}, we used $k = 4$, $m = 2$ for our Kendall shape space's hyperparameters. To implement $SO(2)$-steerable equivariant CNNs, we used an equivariance among a cyclic group of order of 8. We used $\beta = 1$, $\gamma=1$ in Eqn.\! \eqref{eq_exact_loss}. For the rest of our Probabilistic U-Net setup, we maintained the same hyperparameters architectures with the original Probabilistic U-Net. We trained the model with 100 epochs.

\begin{figure}
    \centering
    \includegraphics[width=0.49\textwidth]{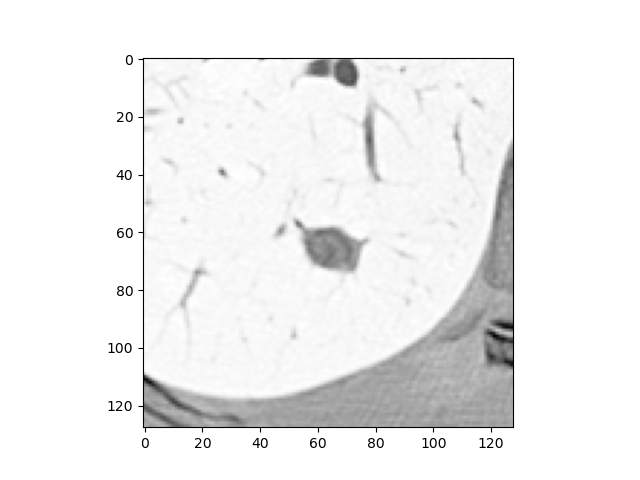}\hfill
    \includegraphics[width=0.49\textwidth]{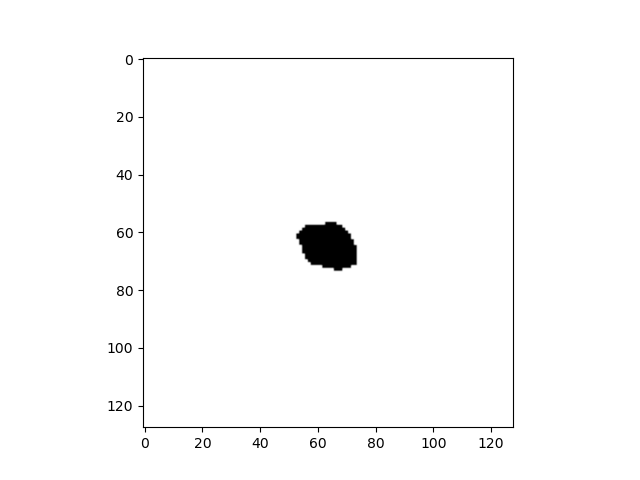}\hfill

    \hrule
    
    \includegraphics[width=0.2\textwidth]{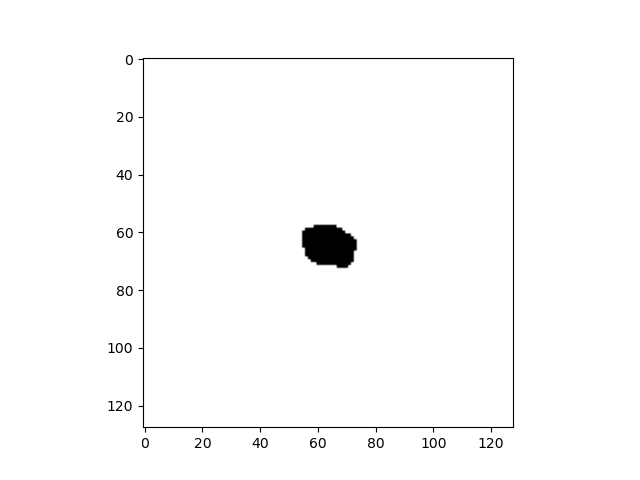}\hfill
    \includegraphics[width=0.2\textwidth]{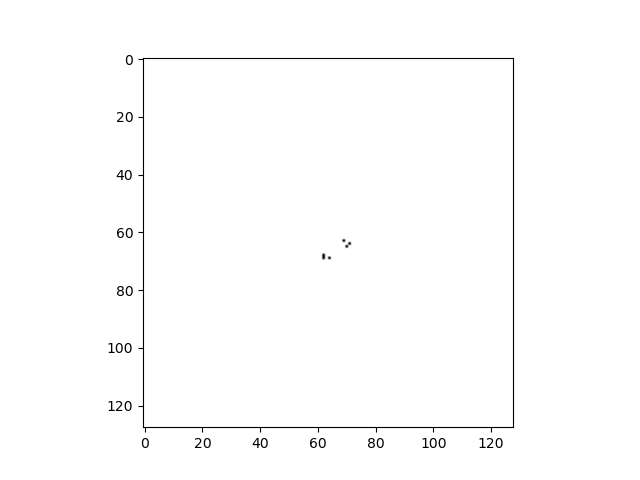}\hfill
    \includegraphics[width=0.2\textwidth]{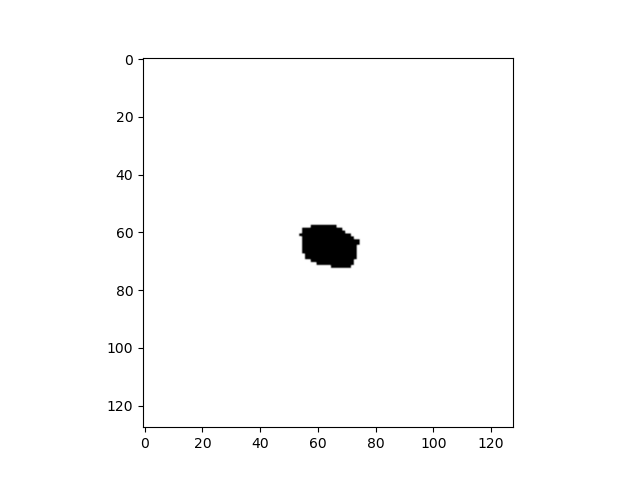}\hfill
    \includegraphics[width=0.2\textwidth]{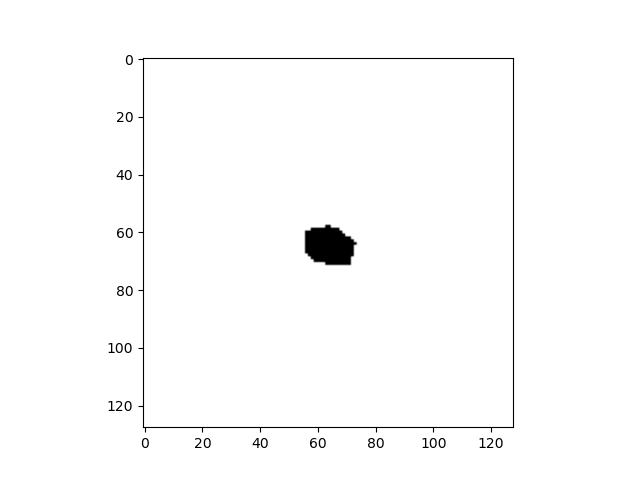}\hfill
    \includegraphics[width=0.2\textwidth]{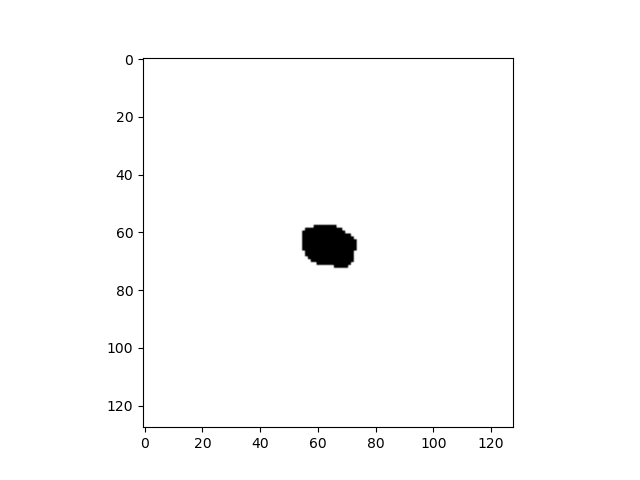}\hfill

    \includegraphics[width=0.2\textwidth]{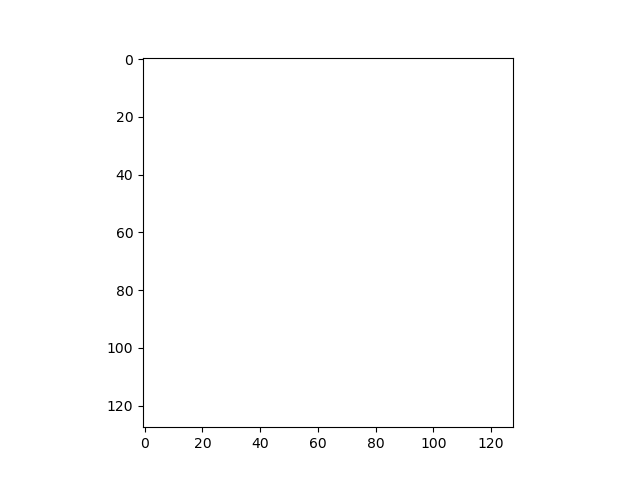}\hfill
    \includegraphics[width=0.2\textwidth]{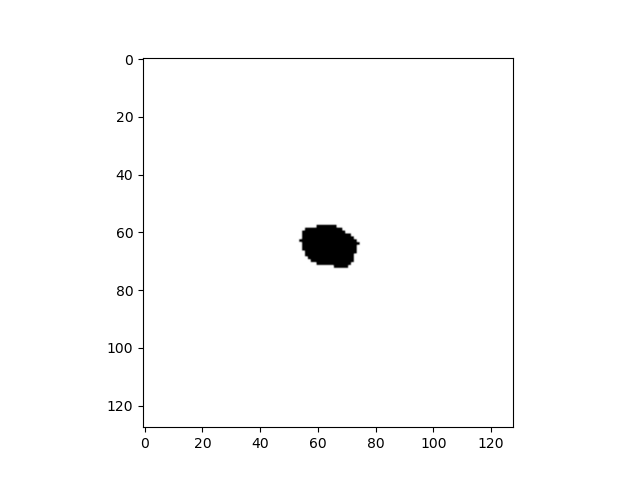}\hfill
    \includegraphics[width=0.2\textwidth]{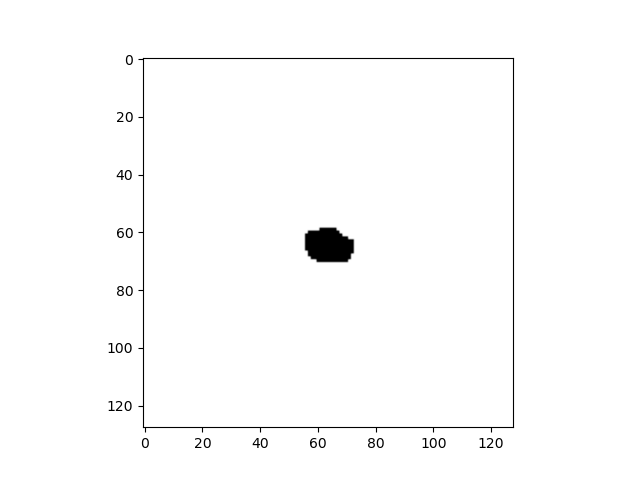}\hfill
    \includegraphics[width=0.2\textwidth]{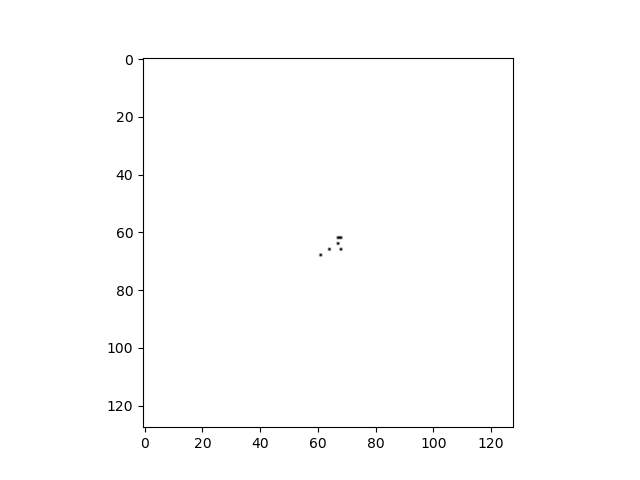}\hfill
    \includegraphics[width=0.2\textwidth]{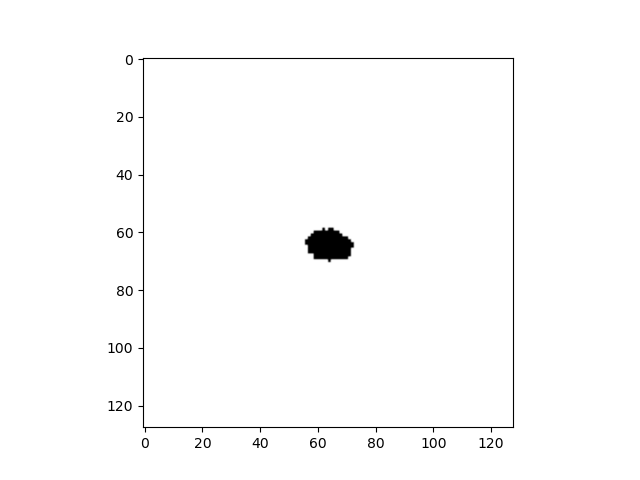}\hfill

    \includegraphics[width=0.2\textwidth]{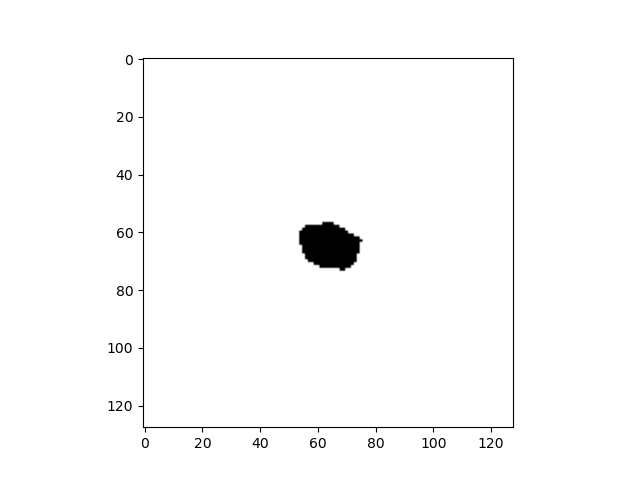}\hfill
    \includegraphics[width=0.2\textwidth]{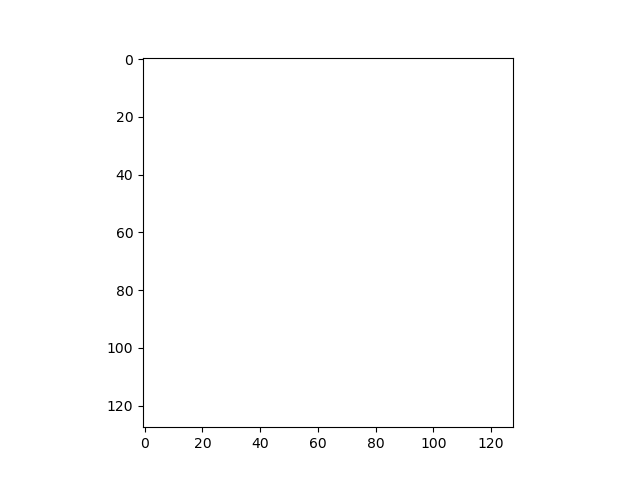}\hfill
    \includegraphics[width=0.2\textwidth]{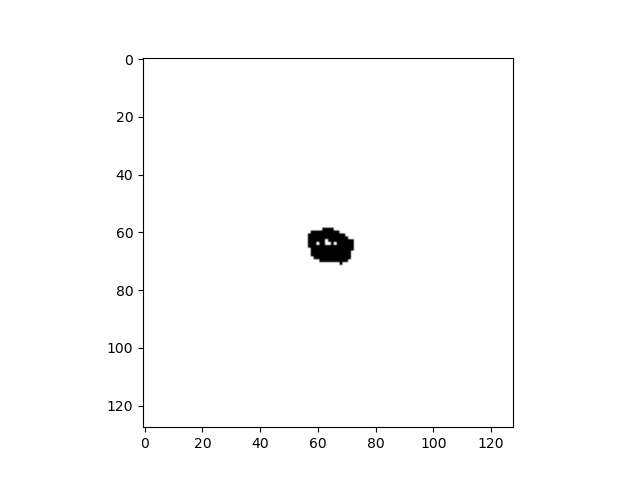}\hfill
    \includegraphics[width=0.2\textwidth]{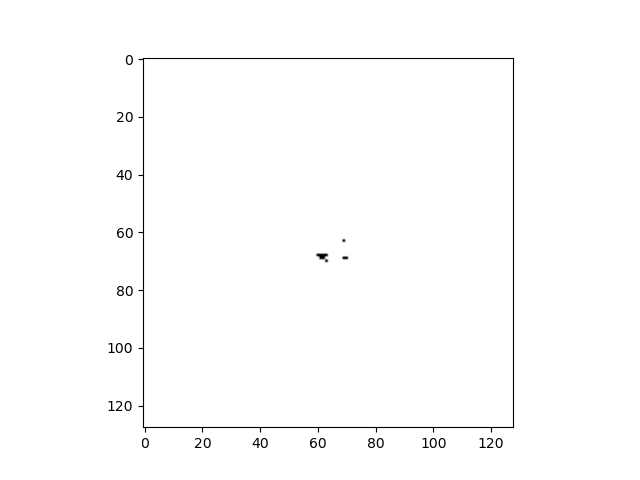}\hfill
    \includegraphics[width=0.2\textwidth]{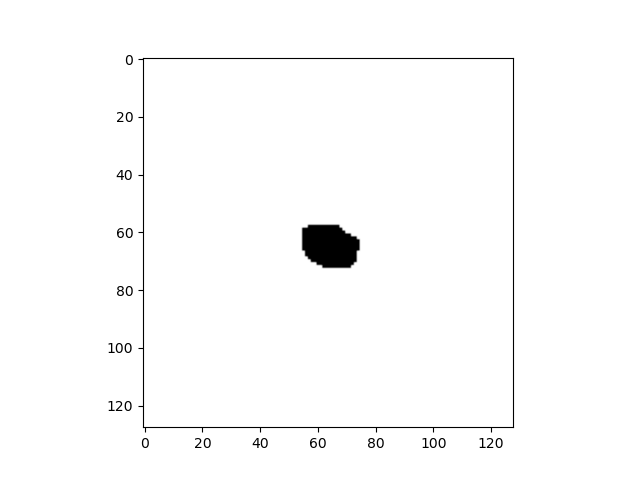}\hfill

    \hrule

    \includegraphics[width=0.2\textwidth]{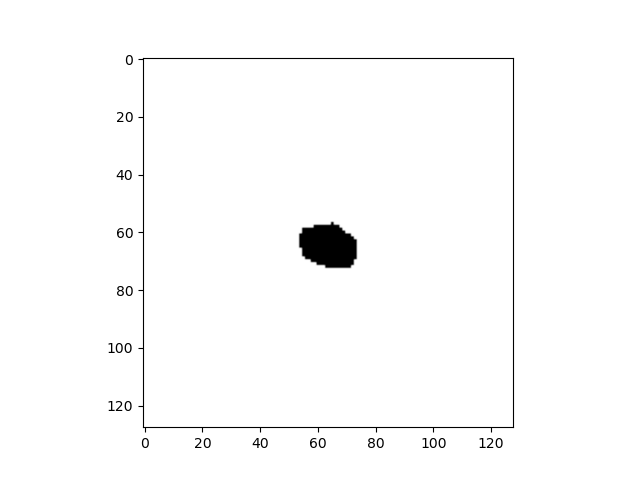}\hfill
    \includegraphics[width=0.2\textwidth]{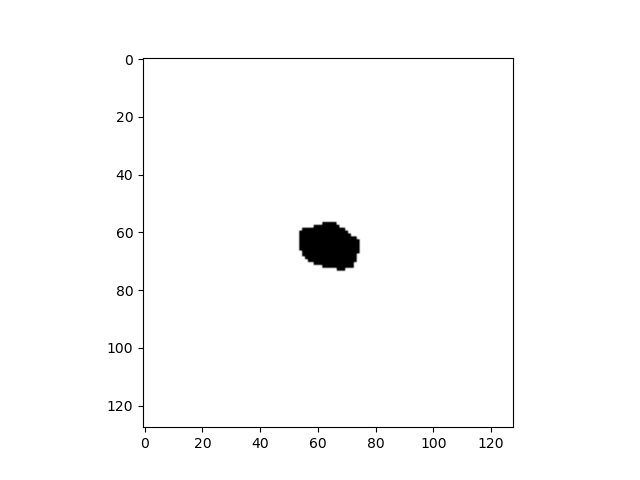}\hfill
    \includegraphics[width=0.2\textwidth]{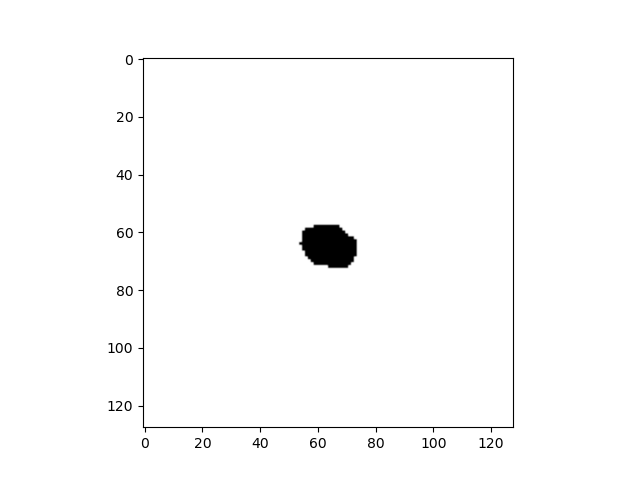}\hfill
    \includegraphics[width=0.2\textwidth]{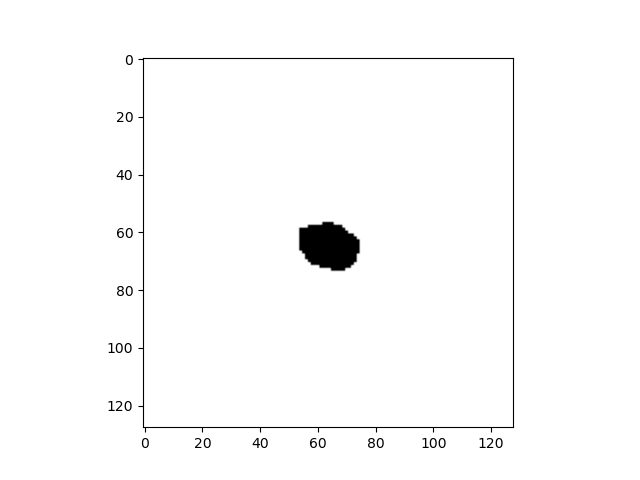}\hfill
    \includegraphics[width=0.2\textwidth]{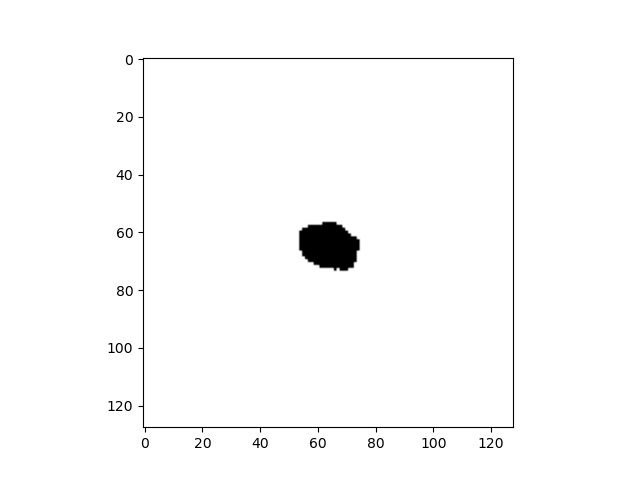}\hfill

    \includegraphics[width=0.2\textwidth]{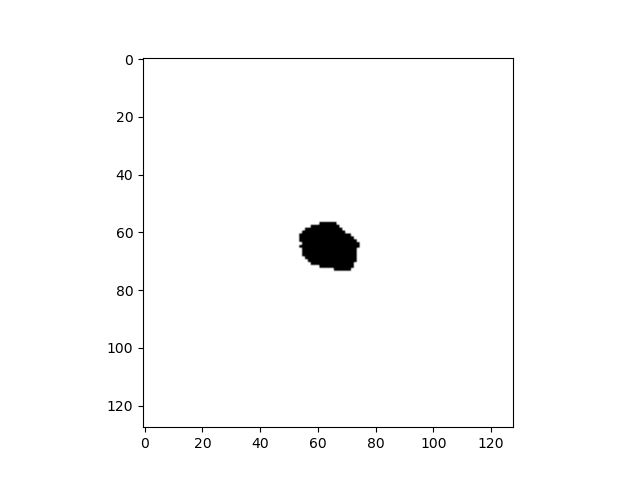}\hfill
    \includegraphics[width=0.2\textwidth]{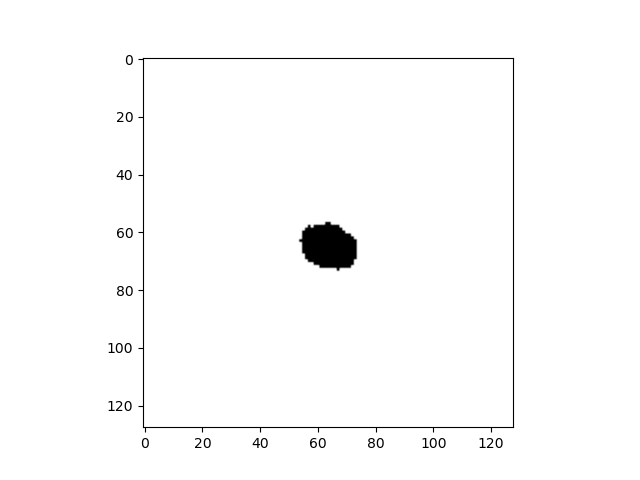}\hfill
    \includegraphics[width=0.2\textwidth]{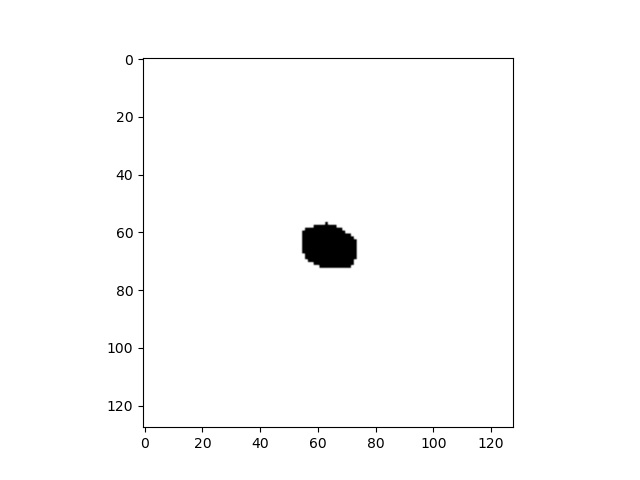}\hfill
    \includegraphics[width=0.2\textwidth]{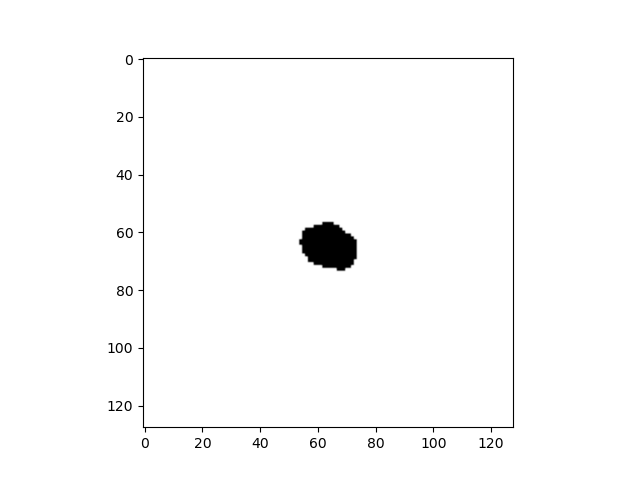}\hfill
    \includegraphics[width=0.2\textwidth]{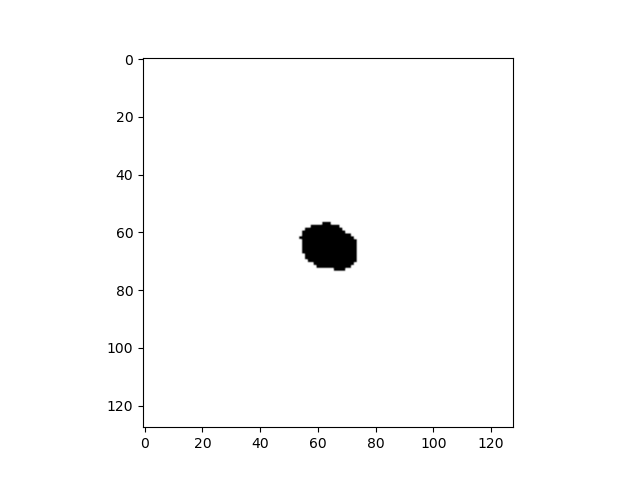}\hfill

    \includegraphics[width=0.2\textwidth]{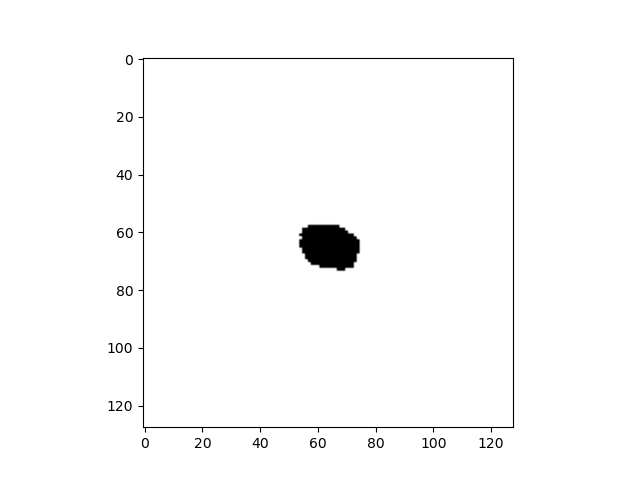}\hfill
    \includegraphics[width=0.2\textwidth]{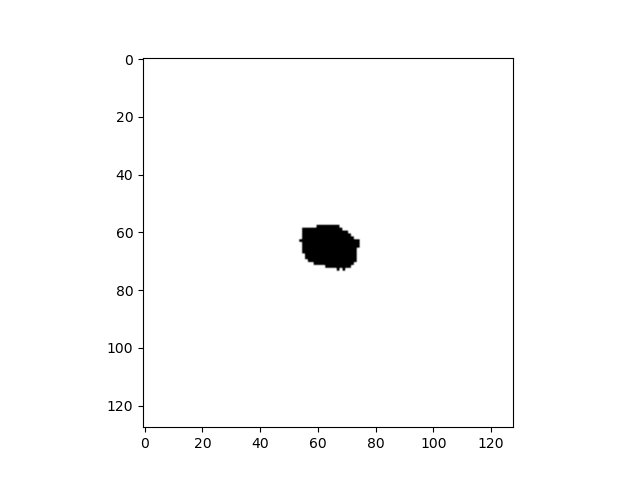}\hfill
    \includegraphics[width=0.2\textwidth]{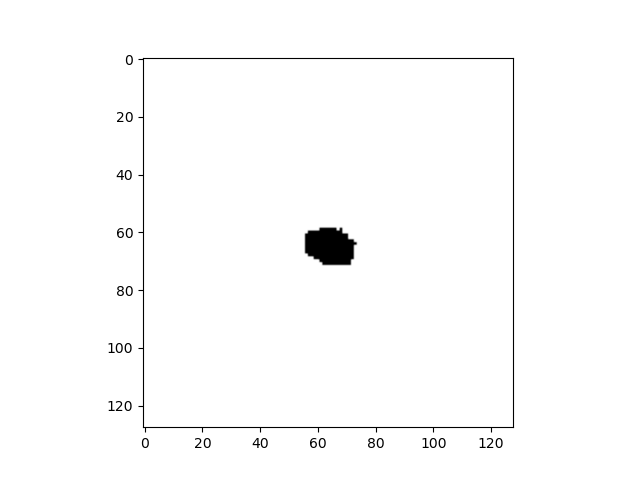}\hfill
    \includegraphics[width=0.2\textwidth]{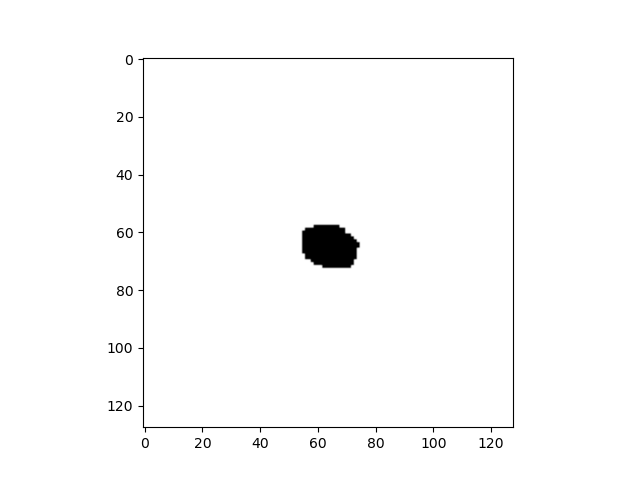}\hfill
    \includegraphics[width=0.2\textwidth]{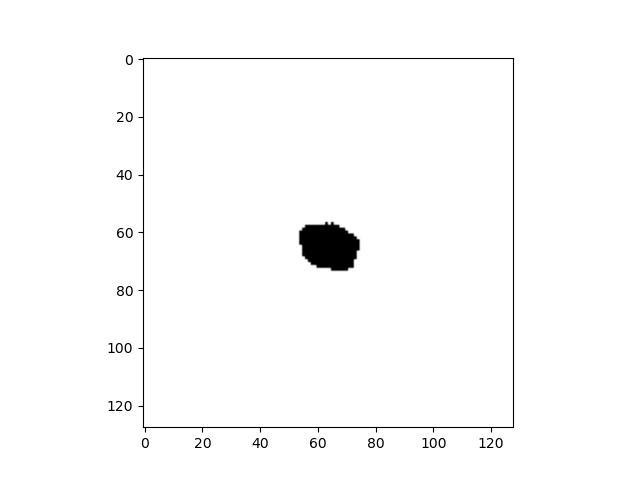}\hfill

    \caption{\textbf{Topleft}: Input image 0 from LIDC data. \textbf{Topright}: Ground truth segmentation. \textbf{2-4 rows}: Segmentation samples from original Probabilistic U-Net. \textbf{5-7 rows}: Segmentation samples from Kendall Shape Probabilistic U-Net. Each row shares the same seed.}
    \label{fig_img0}
\end{figure}

\begin{figure}
    \centering
    \includegraphics[width=0.49\textwidth]{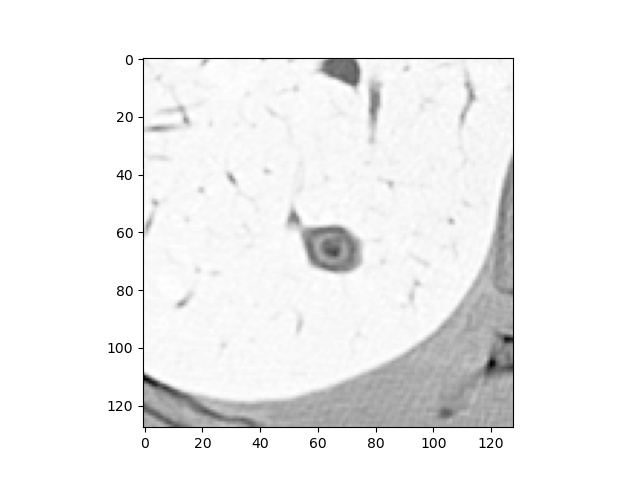}\hfill
    \includegraphics[width=0.49\textwidth]{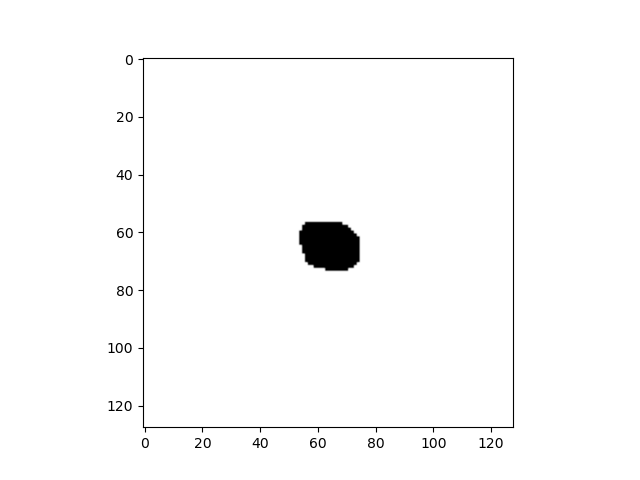}\hfill

    \hrule
    
    \includegraphics[width=0.2\textwidth]{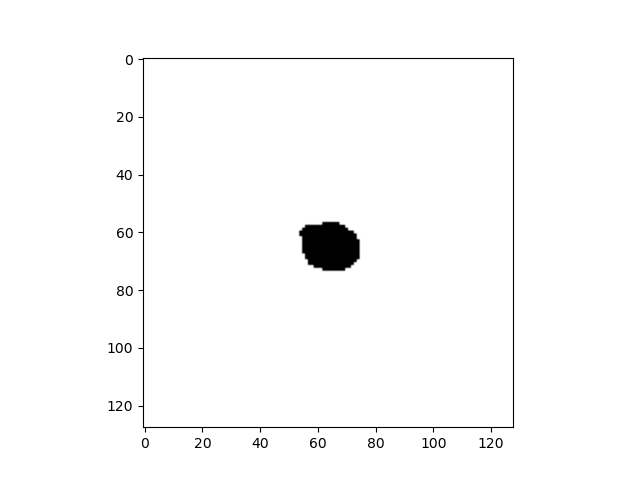}\hfill
    \includegraphics[width=0.2\textwidth]{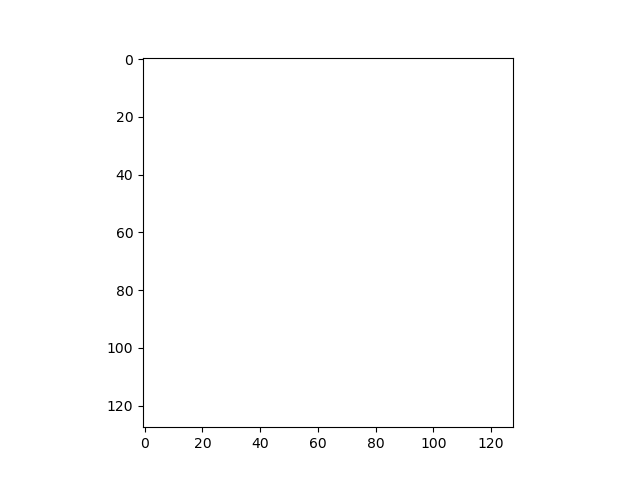}\hfill
    \includegraphics[width=0.2\textwidth]{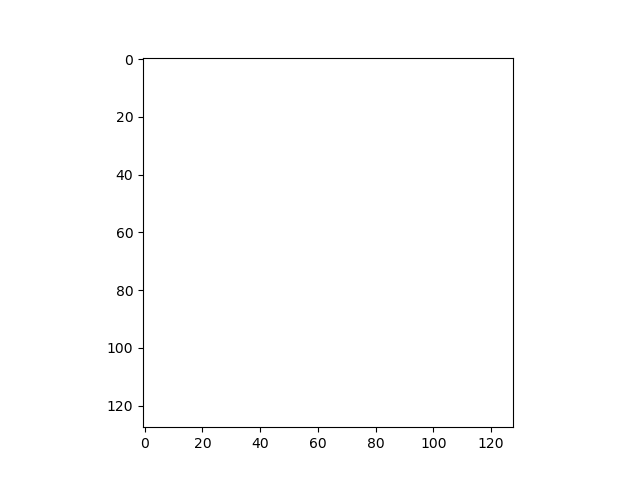}\hfill
    \includegraphics[width=0.2\textwidth]{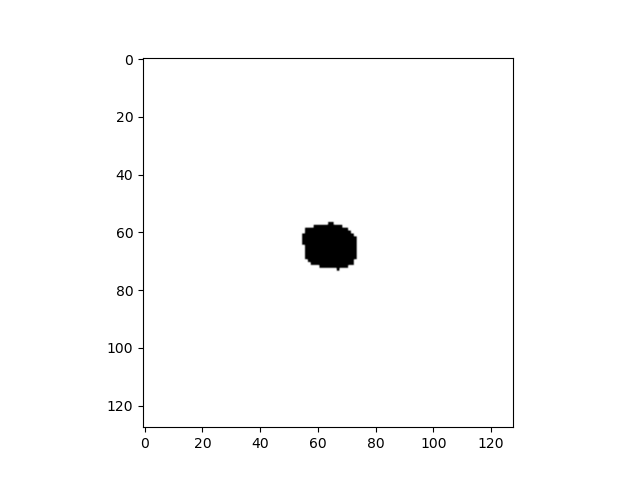}\hfill
    \includegraphics[width=0.2\textwidth]{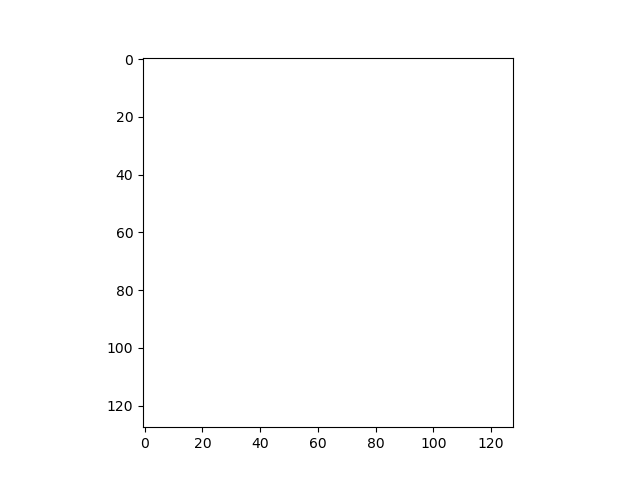}\hfill

    \includegraphics[width=0.2\textwidth]{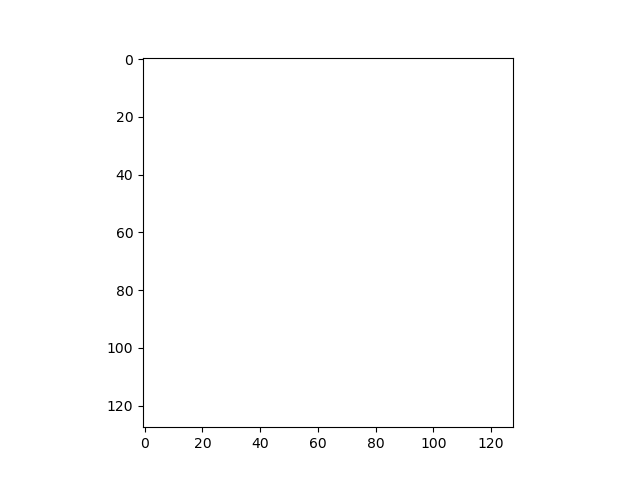}\hfill
    \includegraphics[width=0.2\textwidth]{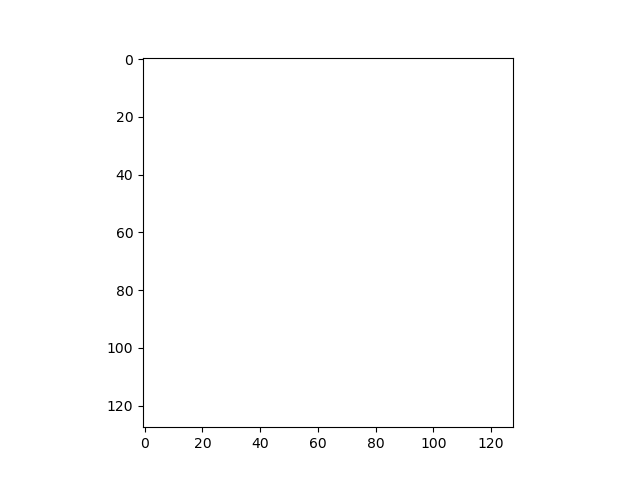}\hfill
    \includegraphics[width=0.2\textwidth]{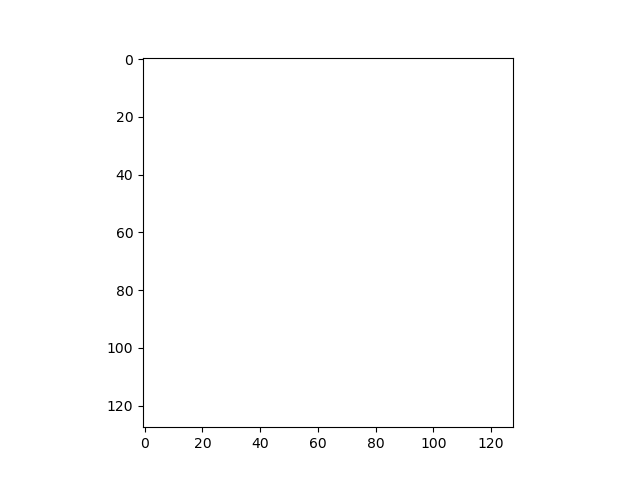}\hfill
    \includegraphics[width=0.2\textwidth]{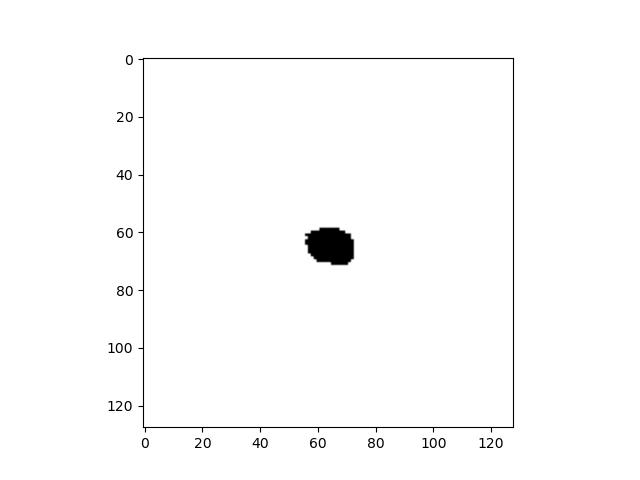}\hfill
    \includegraphics[width=0.2\textwidth]{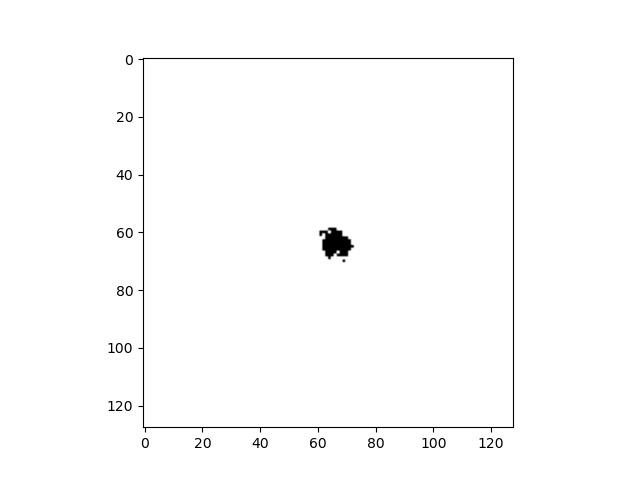}\hfill

    \includegraphics[width=0.2\textwidth]{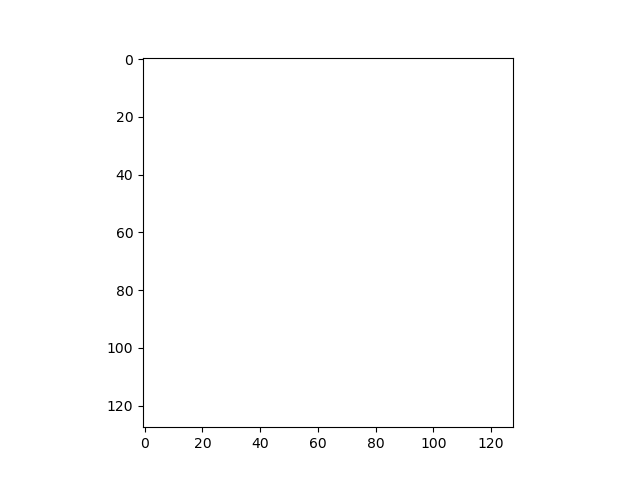}\hfill
    \includegraphics[width=0.2\textwidth]{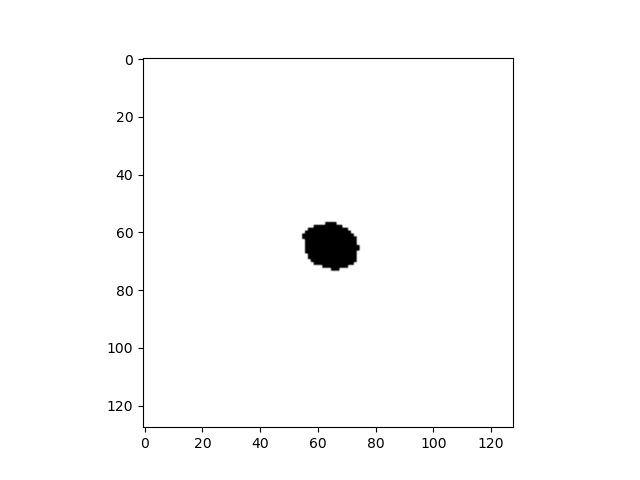}\hfill
    \includegraphics[width=0.2\textwidth]{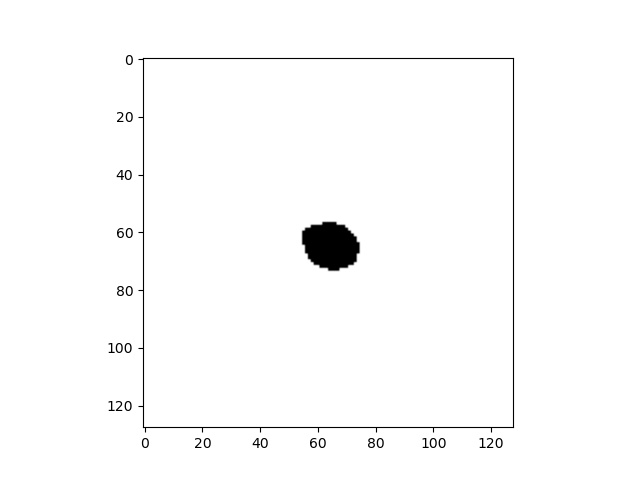}\hfill
    \includegraphics[width=0.2\textwidth]{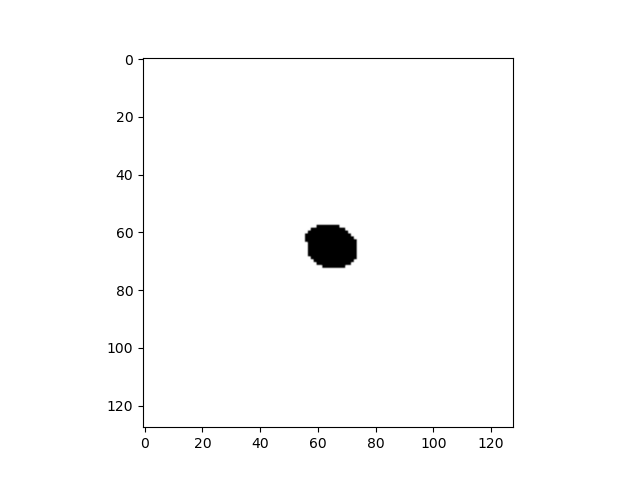}\hfill
    \includegraphics[width=0.2\textwidth]{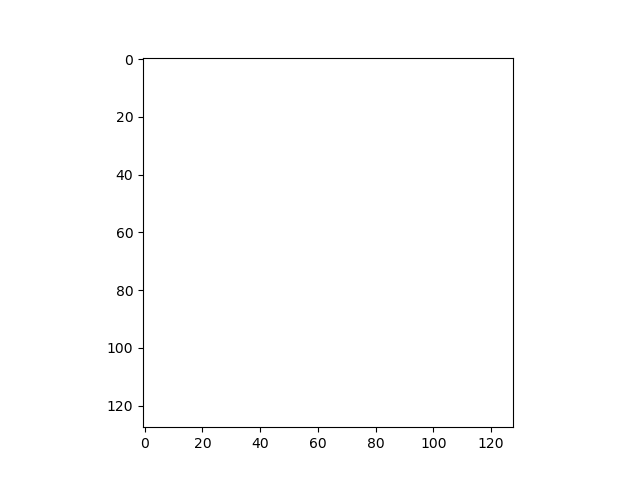}\hfill

    \hrule

    \includegraphics[width=0.2\textwidth]{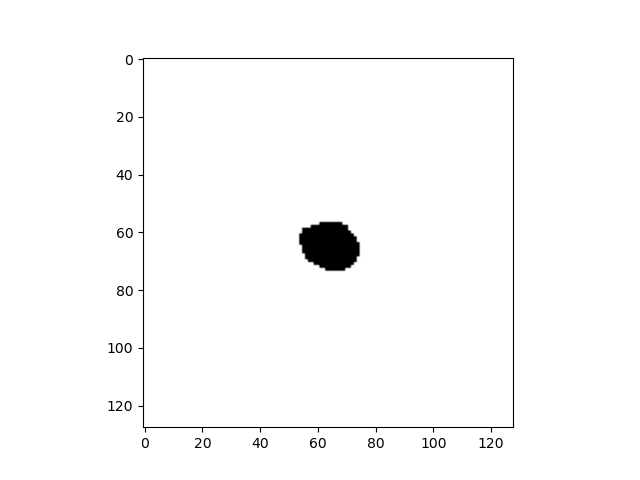}\hfill
    \includegraphics[width=0.2\textwidth]{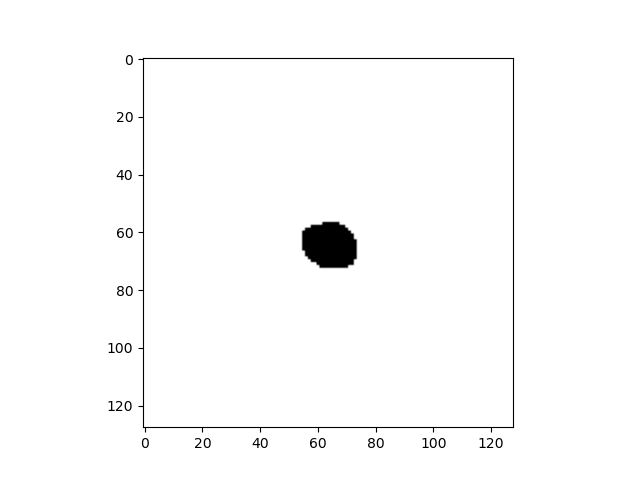}\hfill
    \includegraphics[width=0.2\textwidth]{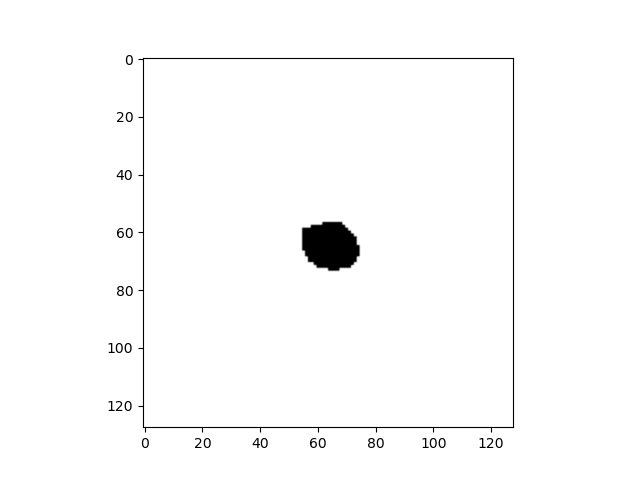}\hfill
    \includegraphics[width=0.2\textwidth]{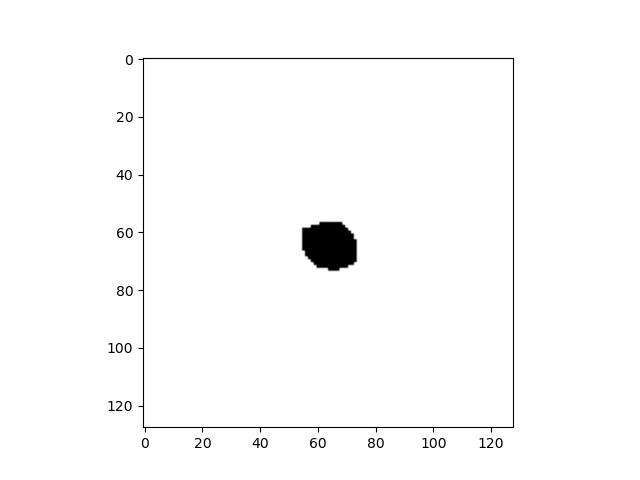}\hfill
    \includegraphics[width=0.2\textwidth]{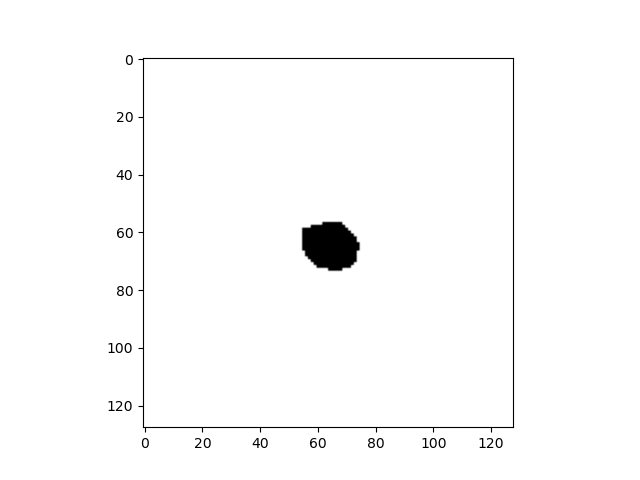}\hfill

    \includegraphics[width=0.2\textwidth]{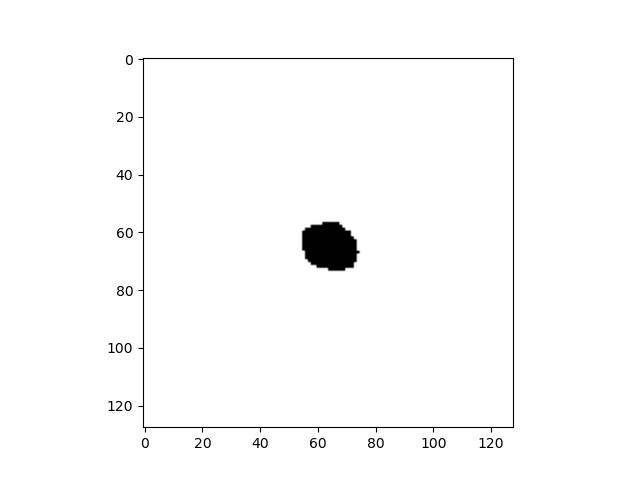}\hfill
    \includegraphics[width=0.2\textwidth]{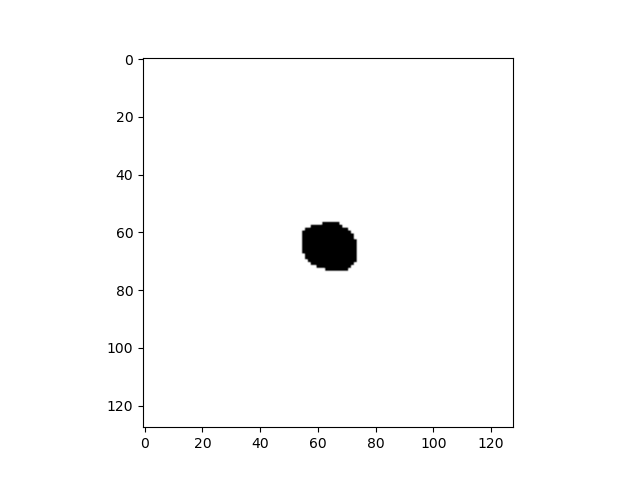}\hfill
    \includegraphics[width=0.2\textwidth]{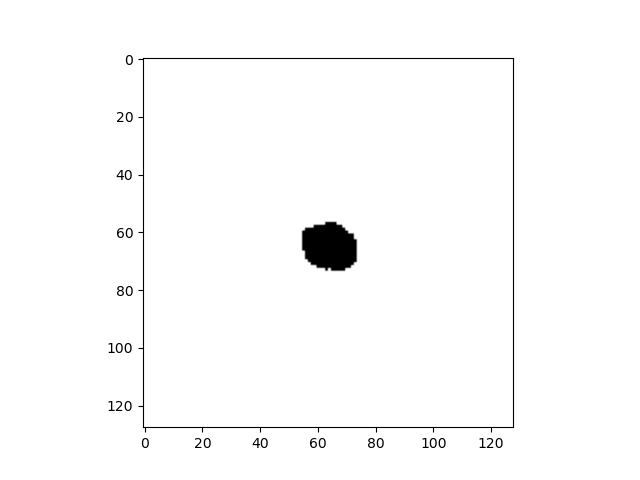}\hfill
    \includegraphics[width=0.2\textwidth]{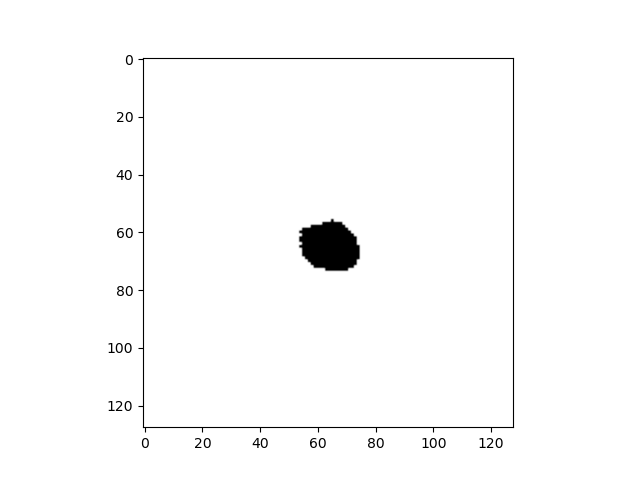}\hfill
    \includegraphics[width=0.2\textwidth]{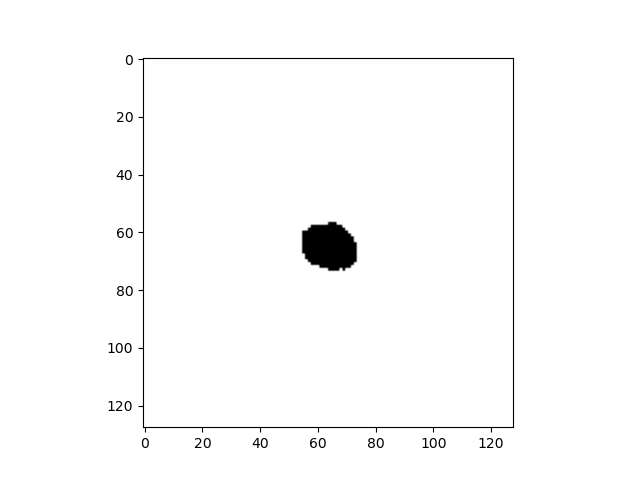}\hfill

    \includegraphics[width=0.2\textwidth]{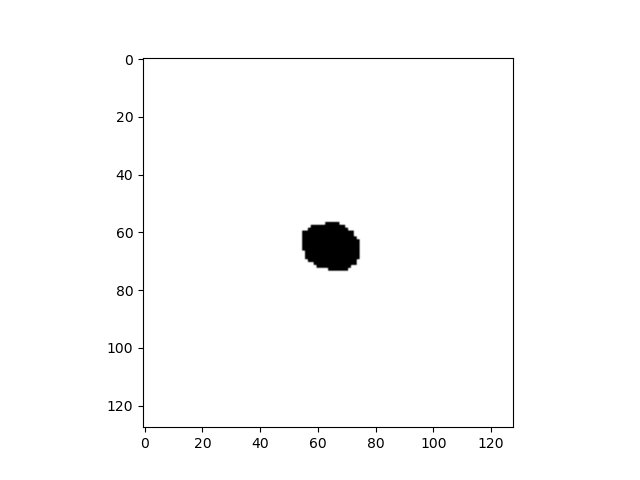}\hfill
    \includegraphics[width=0.2\textwidth]{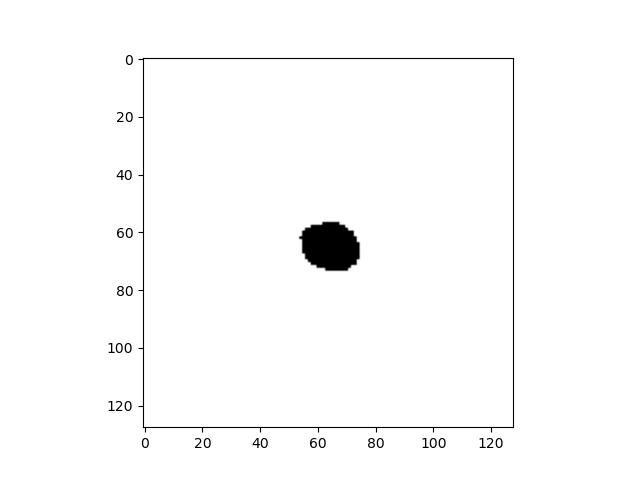}\hfill
    \includegraphics[width=0.2\textwidth]{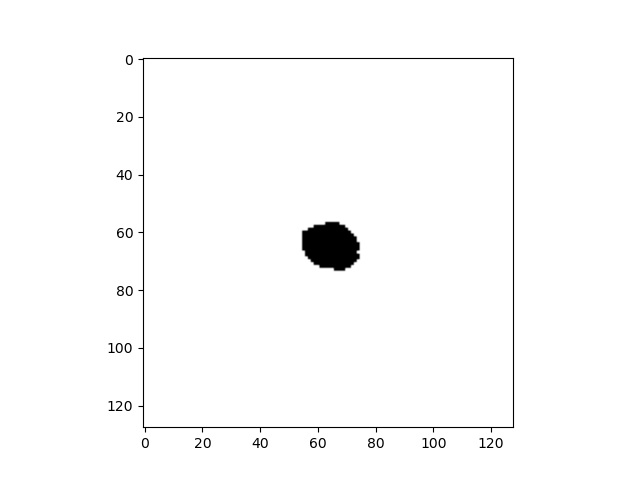}\hfill
    \includegraphics[width=0.2\textwidth]{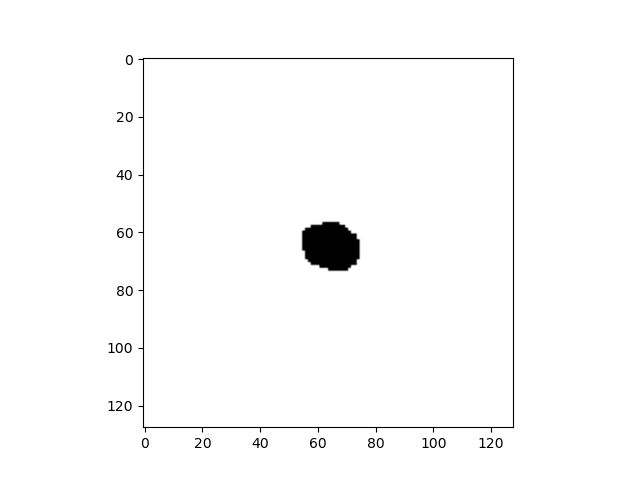}\hfill
    \includegraphics[width=0.2\textwidth]{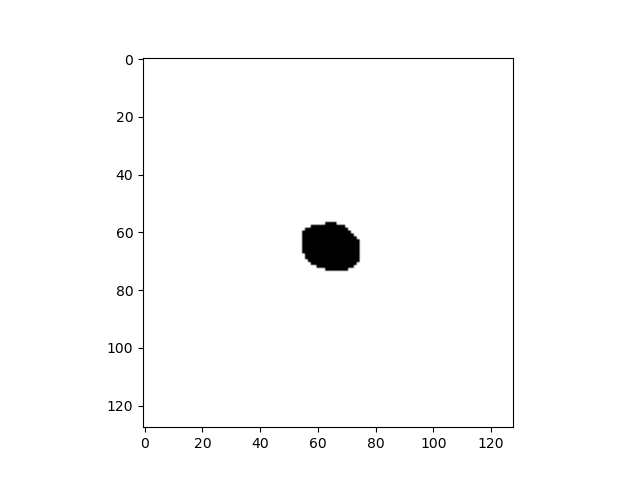}\hfill
    \caption{\textbf{Topleft}: Input image 1 from LIDC data. \textbf{Topright}: Ground truth segmentation. \textbf{2-4 rows}: Segmentation samples from original Probabilistic U-Net. \textbf{5-7 rows}: Segmentation samples from Kendall Shape Probabilistic U-Net. Each row shares the same seed.}
    \label{fig_img1}
\end{figure}

\begin{figure}
    \centering
    \includegraphics[width=0.49\textwidth]{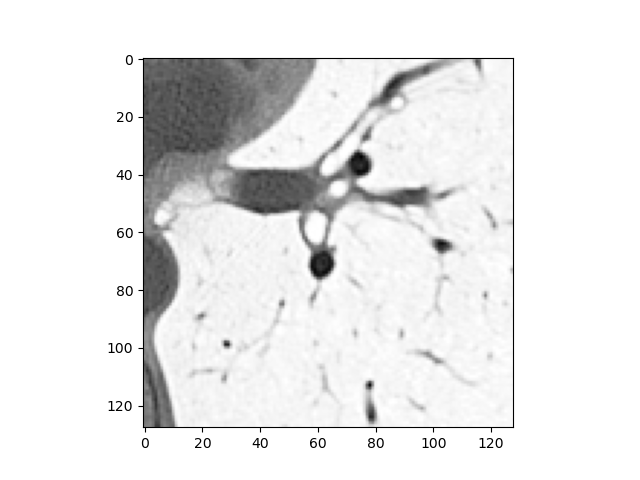}\hfill
    \includegraphics[width=0.49\textwidth]{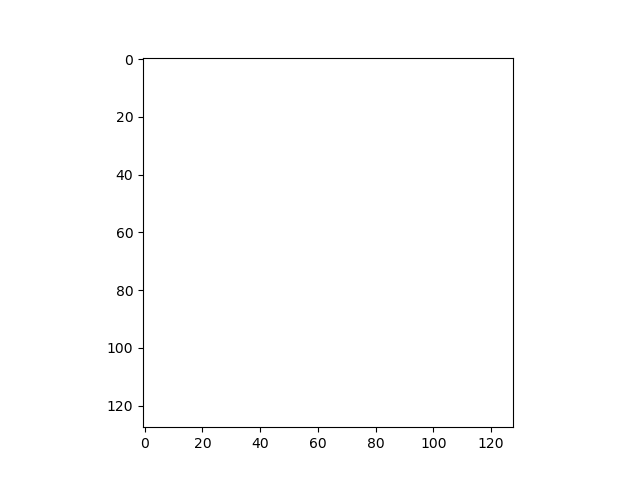}\hfill

    \hrule
    
    \includegraphics[width=0.2\textwidth]{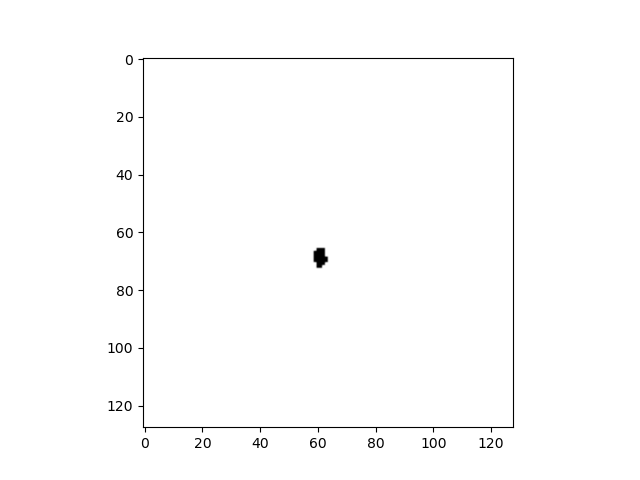}\hfill
    \includegraphics[width=0.2\textwidth]{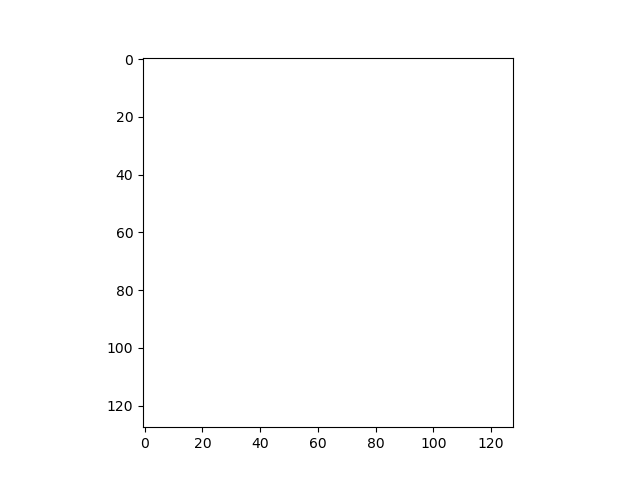}\hfill
    \includegraphics[width=0.2\textwidth]{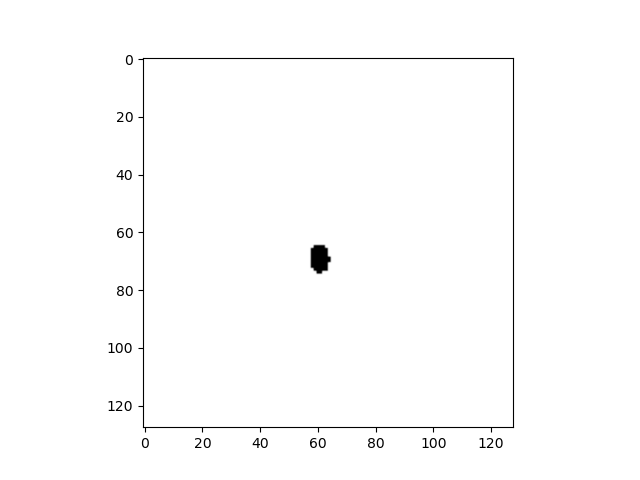}\hfill
    \includegraphics[width=0.2\textwidth]{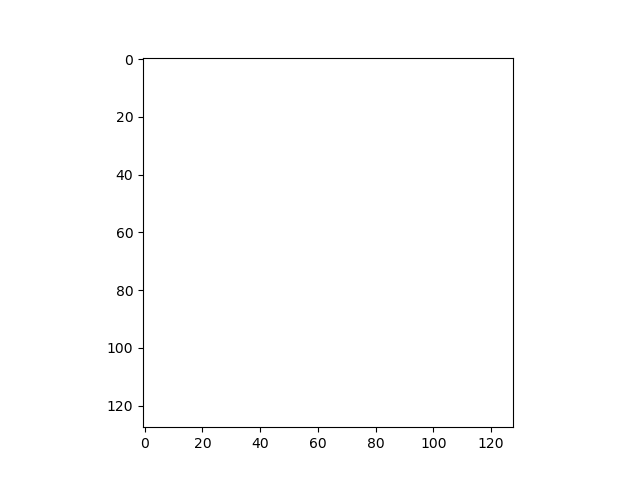}\hfill
    \includegraphics[width=0.2\textwidth]{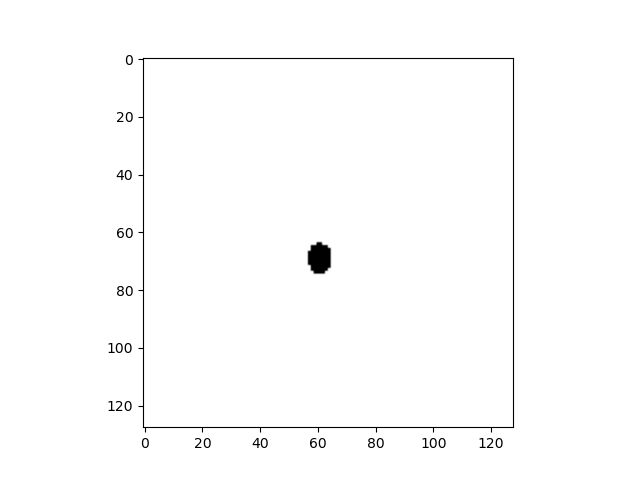}\hfill

    \includegraphics[width=0.2\textwidth]{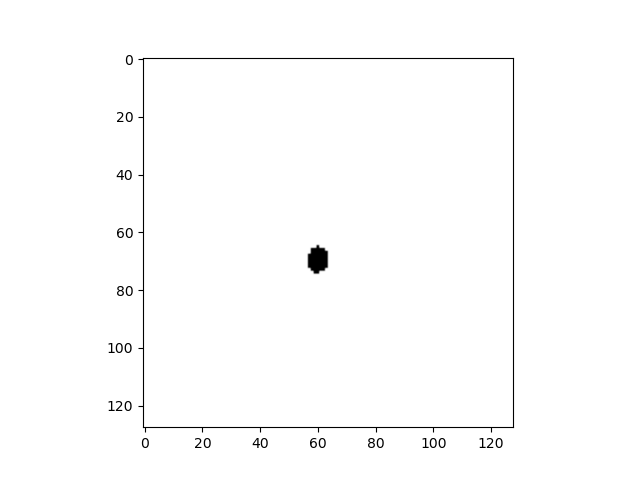}\hfill
    \includegraphics[width=0.2\textwidth]{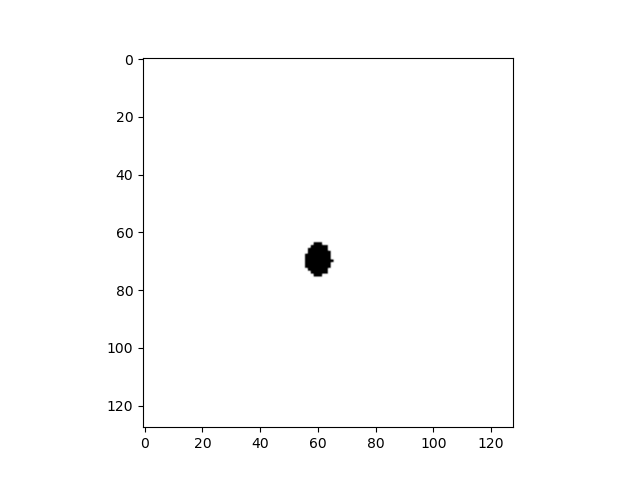}\hfill
    \includegraphics[width=0.2\textwidth]{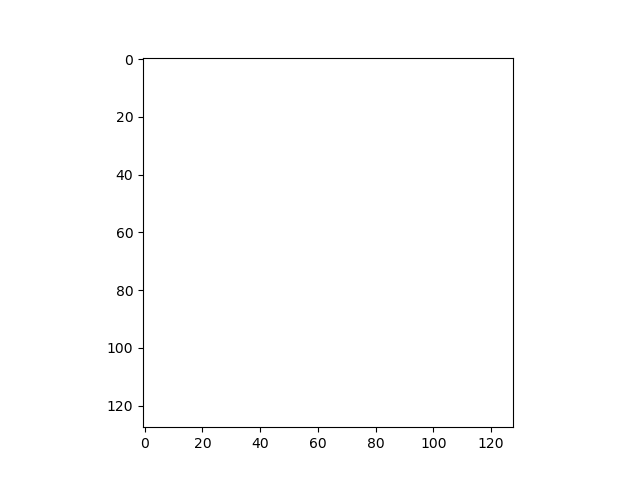}\hfill
    \includegraphics[width=0.2\textwidth]{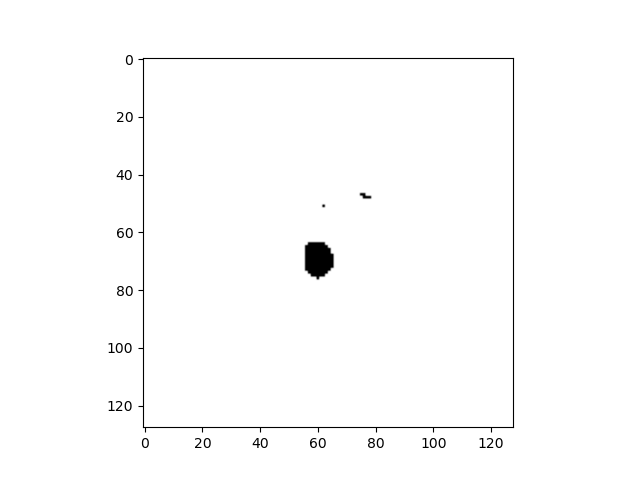}\hfill
    \includegraphics[width=0.2\textwidth]{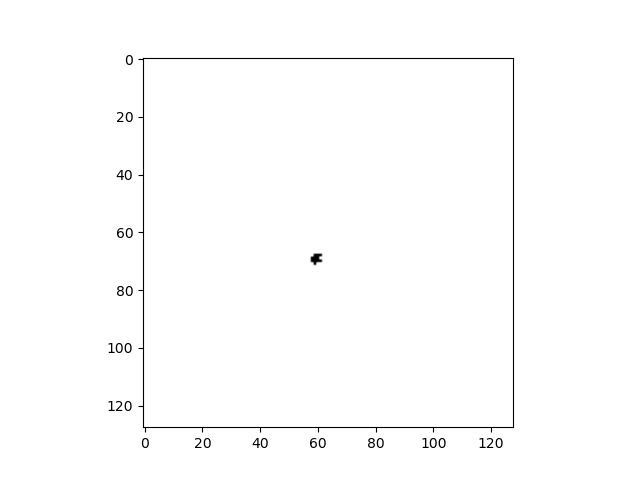}\hfill

    \includegraphics[width=0.2\textwidth]{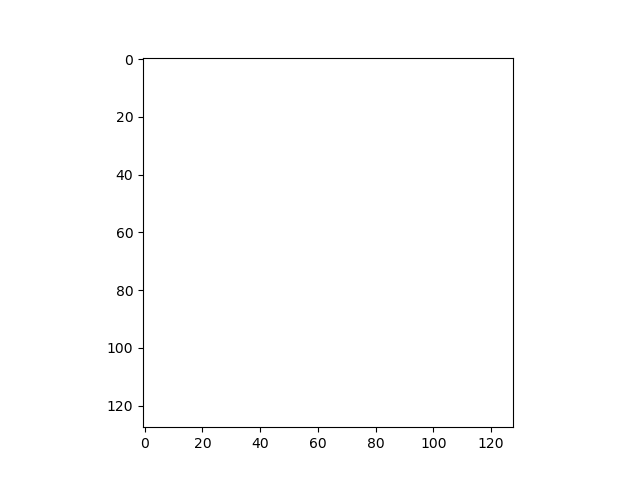}\hfill
    \includegraphics[width=0.2\textwidth]{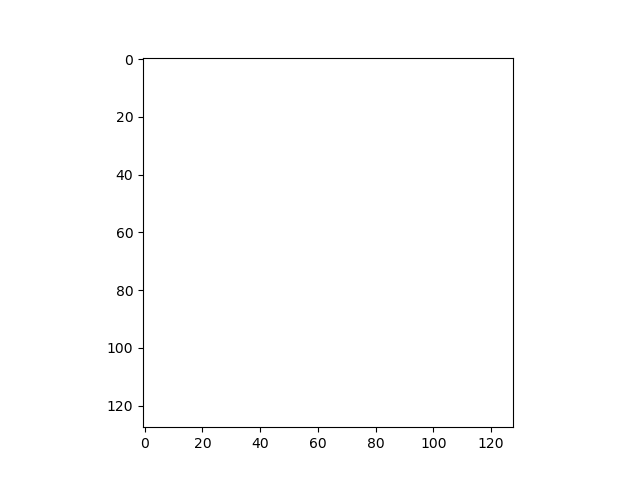}\hfill
    \includegraphics[width=0.2\textwidth]{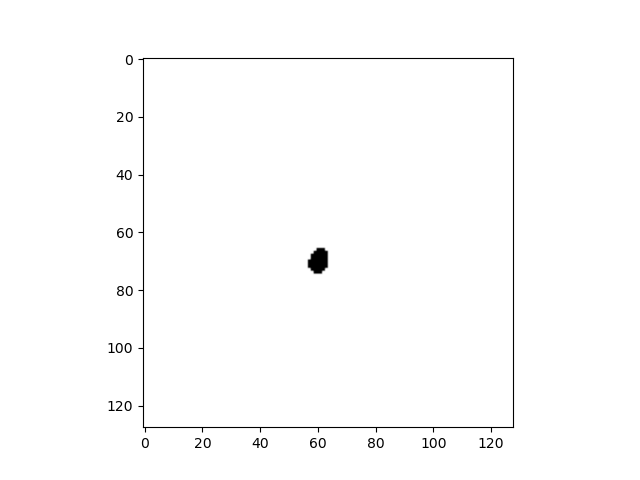}\hfill
    \includegraphics[width=0.2\textwidth]{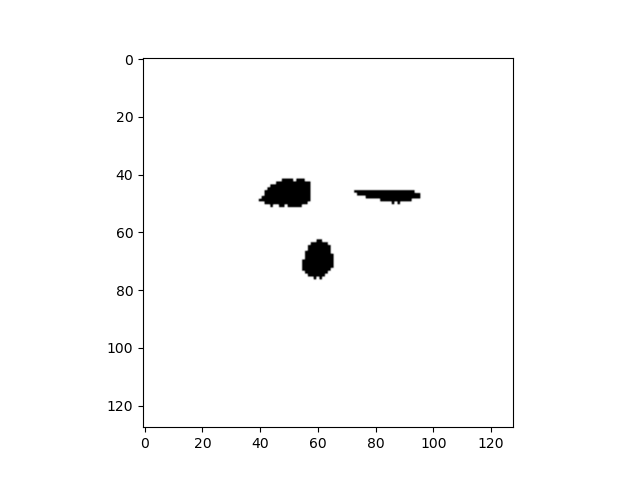}\hfill
    \includegraphics[width=0.2\textwidth]{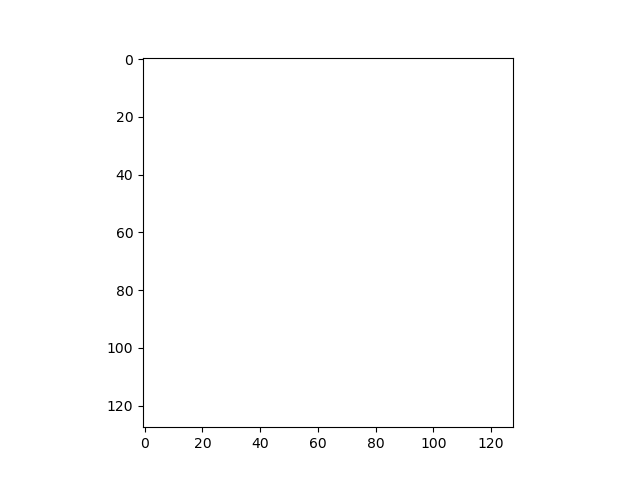}\hfill

    \hrule

    \includegraphics[width=0.2\textwidth]{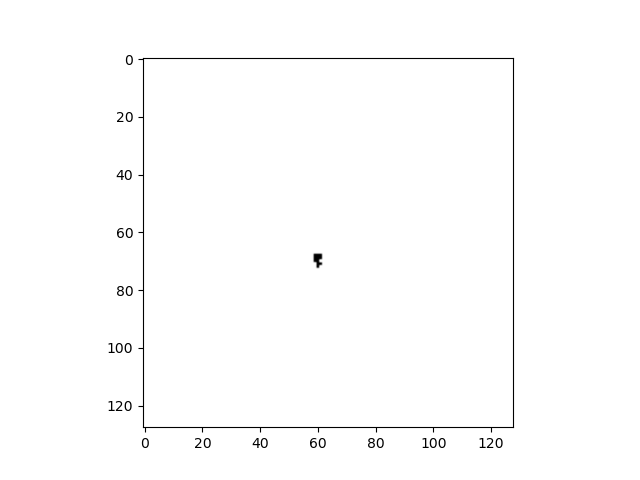}\hfill
    \includegraphics[width=0.2\textwidth]{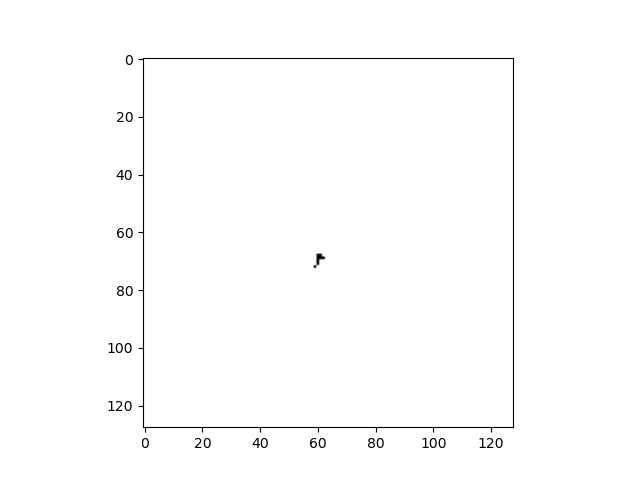}\hfill
    \includegraphics[width=0.2\textwidth]{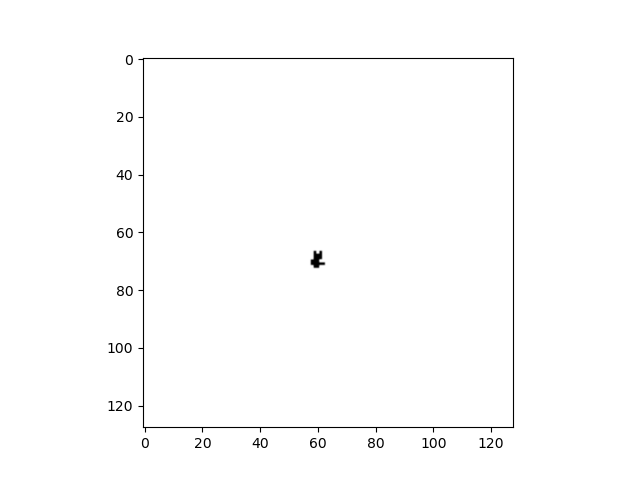}\hfill
    \includegraphics[width=0.2\textwidth]{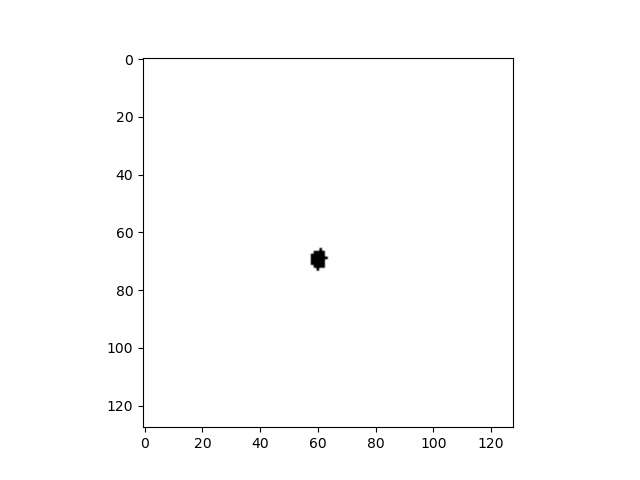}\hfill
    \includegraphics[width=0.2\textwidth]{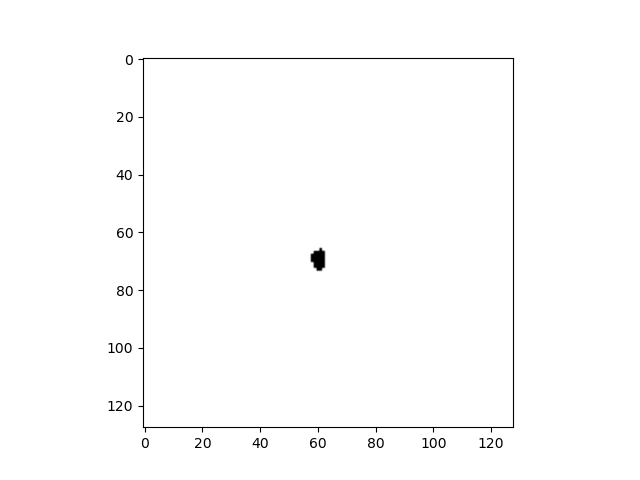}\hfill

    \includegraphics[width=0.2\textwidth]{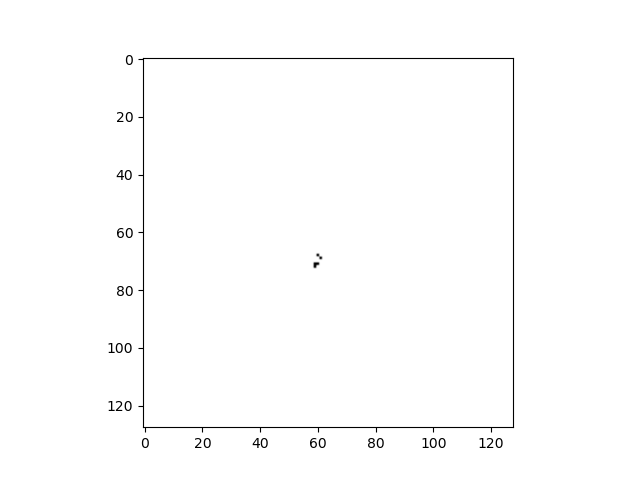}\hfill
    \includegraphics[width=0.2\textwidth]{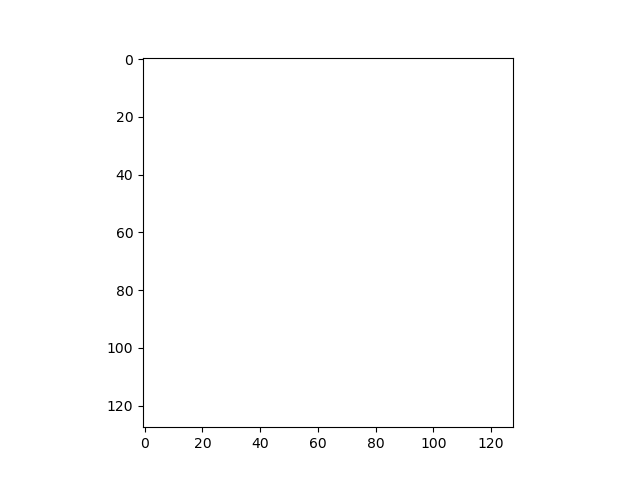}\hfill
    \includegraphics[width=0.2\textwidth]{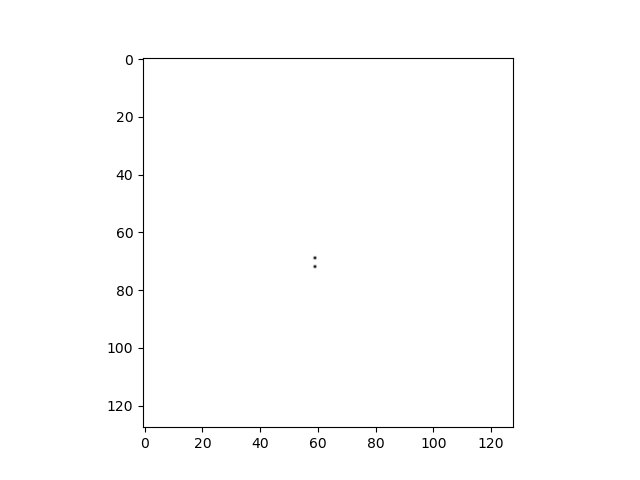}\hfill
    \includegraphics[width=0.2\textwidth]{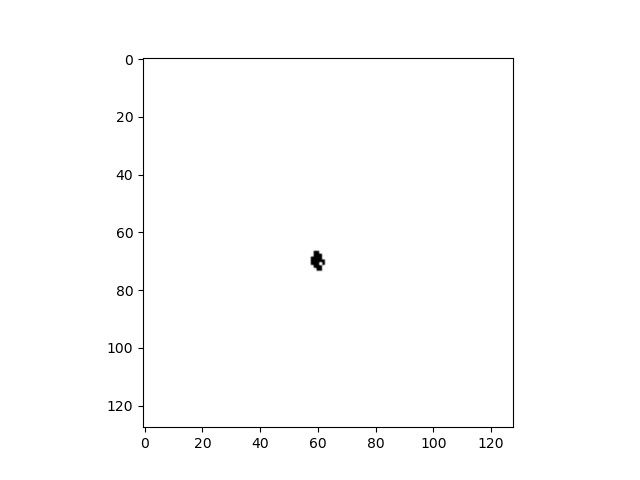}\hfill
    \includegraphics[width=0.2\textwidth]{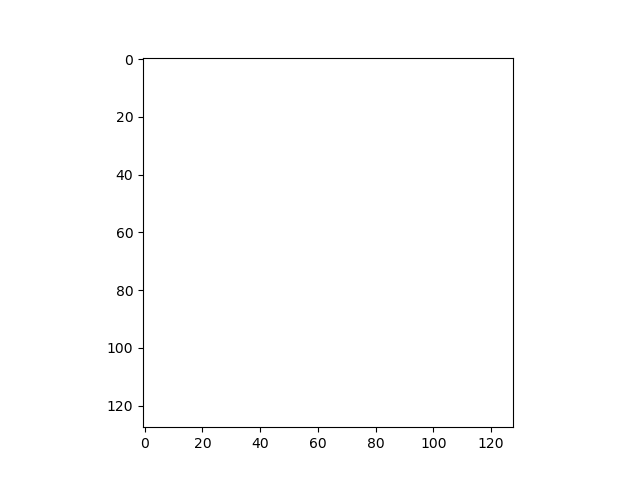}\hfill

    \includegraphics[width=0.2\textwidth]{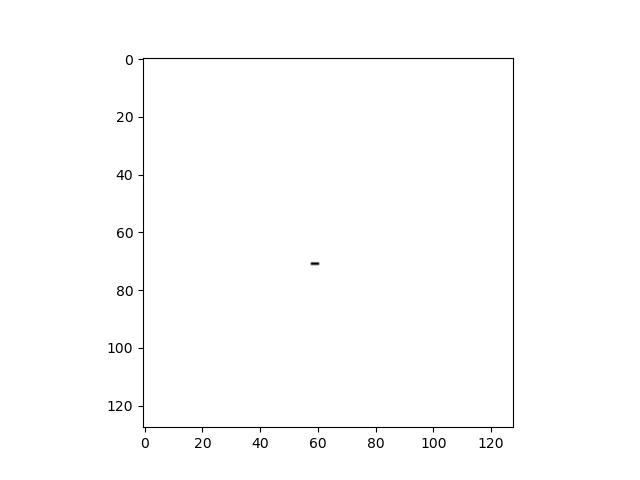}\hfill
    \includegraphics[width=0.2\textwidth]{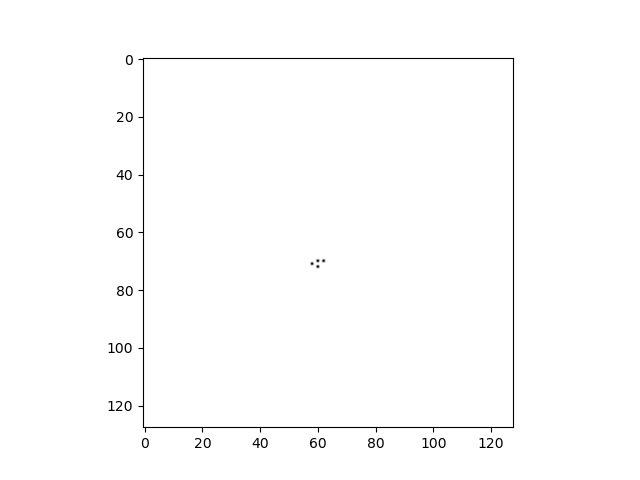}\hfill
    \includegraphics[width=0.2\textwidth]{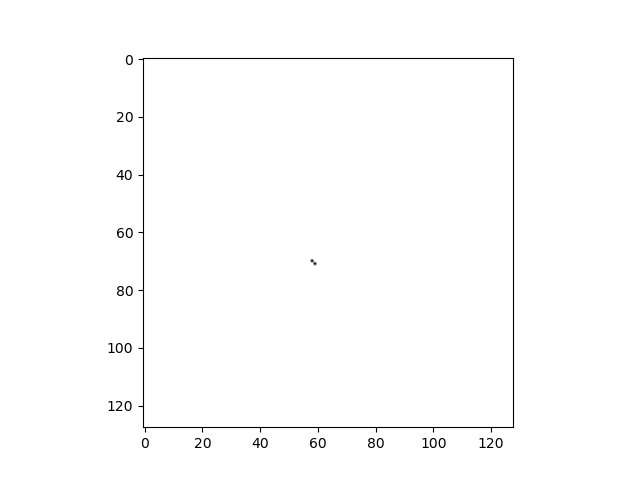}\hfill
    \includegraphics[width=0.2\textwidth]{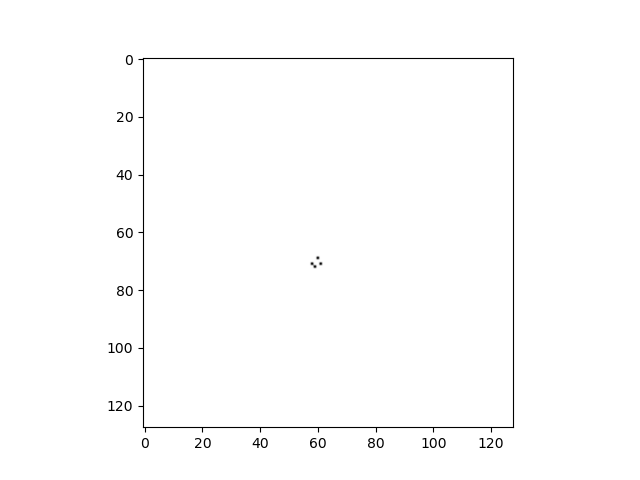}\hfill
    \includegraphics[width=0.2\textwidth]{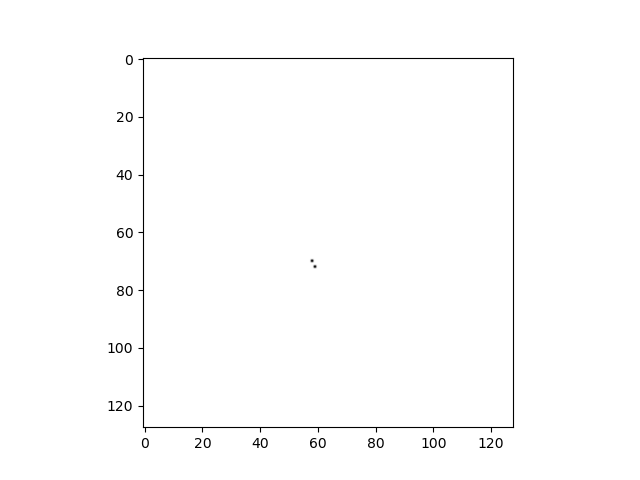}\hfill
    \caption{\textbf{Topleft}: Input image 2 from LIDC data. \textbf{Topright}: Ground truth segmentation. \textbf{2-4 rows}: Segmentation samples from original Probabilistic U-Net. \textbf{5-7 rows}: Segmentation samples from Kendall Shape Probabilistic U-Net. Each row shares the same seed.}
    \label{fig_img2}
\end{figure}

\begin{figure}
    \centering
    \includegraphics[width=0.49\textwidth]{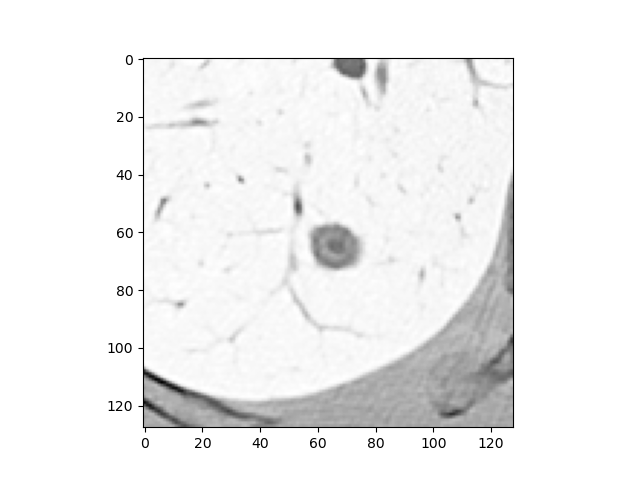}\hfill
    \includegraphics[width=0.49\textwidth]{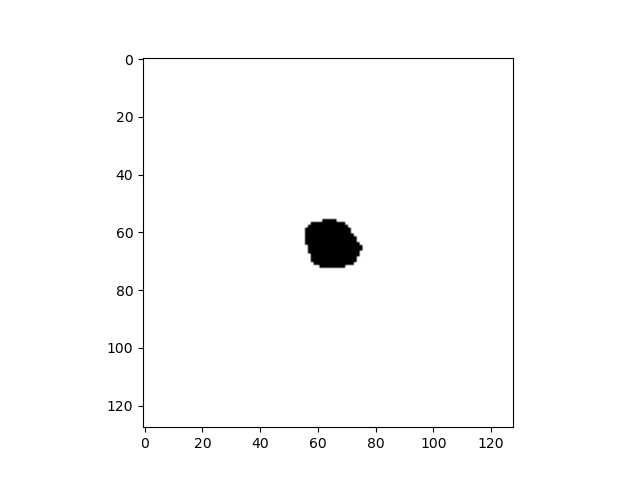}\hfill

    \hrule
    
    \includegraphics[width=0.2\textwidth]{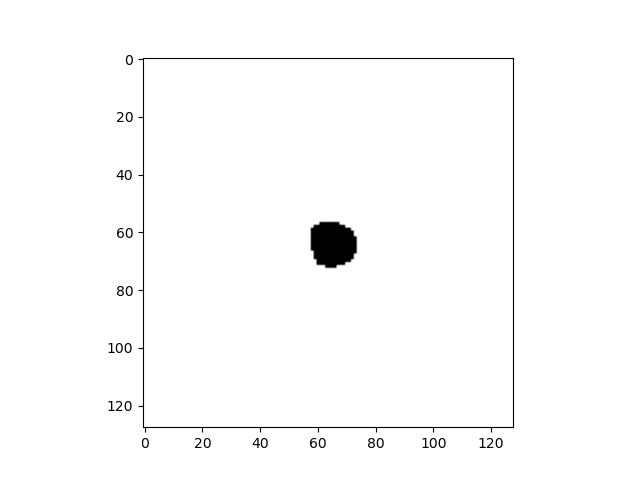}\hfill
    \includegraphics[width=0.2\textwidth]{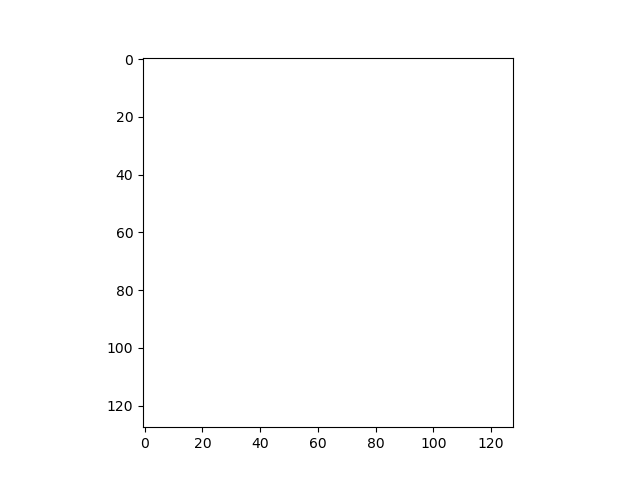}\hfill
    \includegraphics[width=0.2\textwidth]{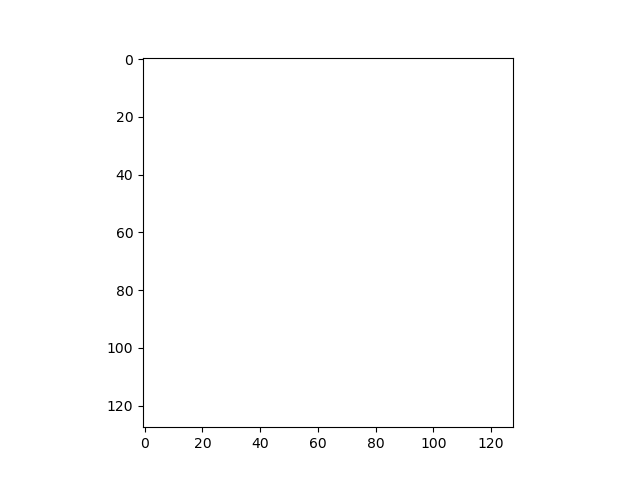}\hfill
    \includegraphics[width=0.2\textwidth]{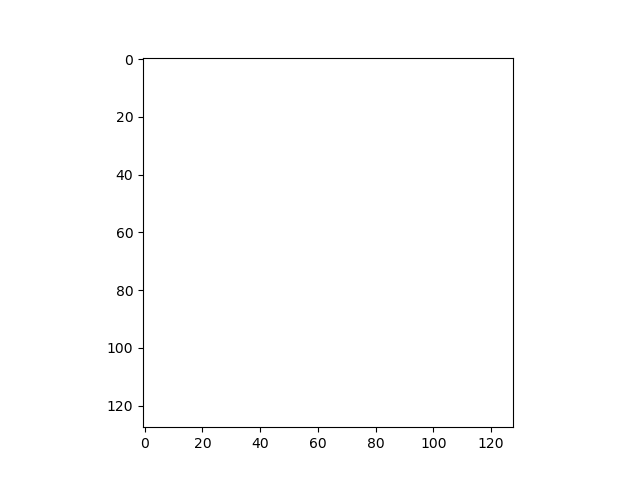}\hfill
    \includegraphics[width=0.2\textwidth]{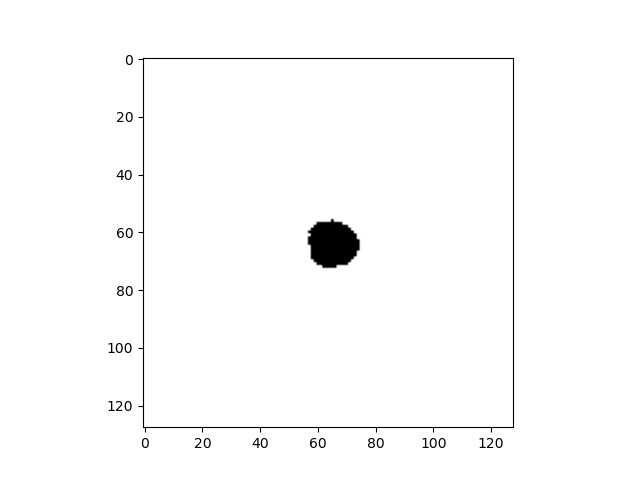}\hfill

    \includegraphics[width=0.2\textwidth]{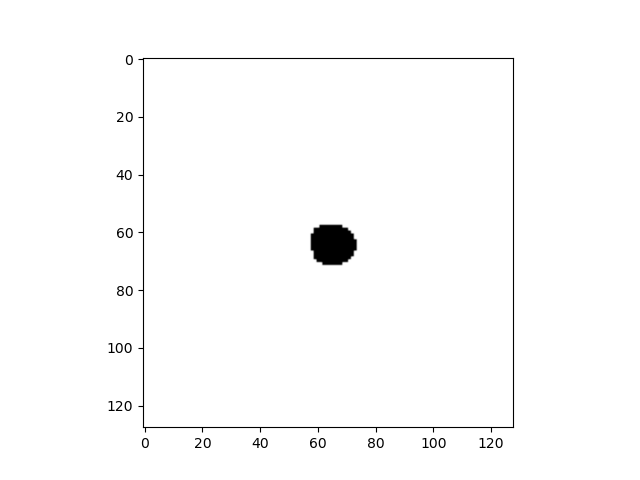}\hfill
    \includegraphics[width=0.2\textwidth]{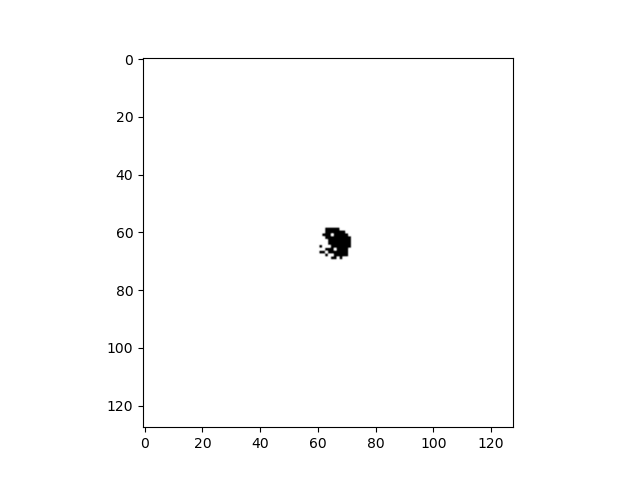}\hfill
    \includegraphics[width=0.2\textwidth]{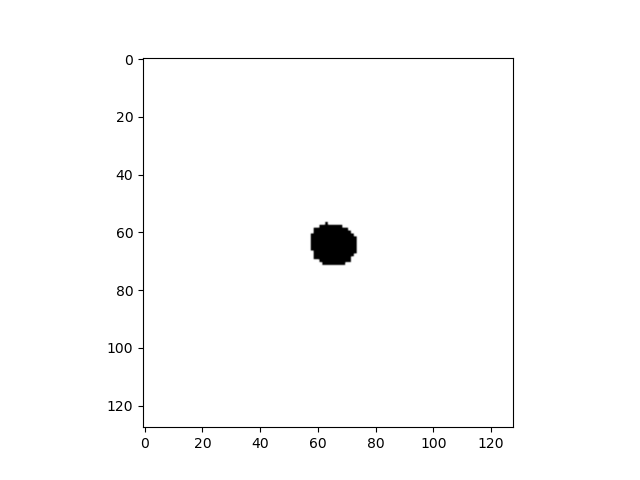}\hfill
    \includegraphics[width=0.2\textwidth]{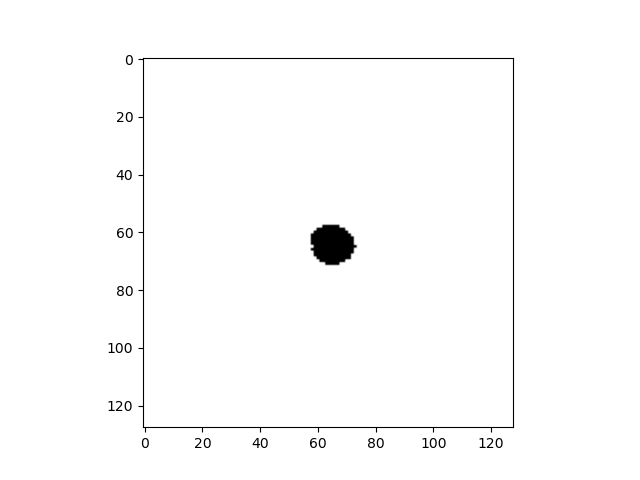}\hfill
    \includegraphics[width=0.2\textwidth]{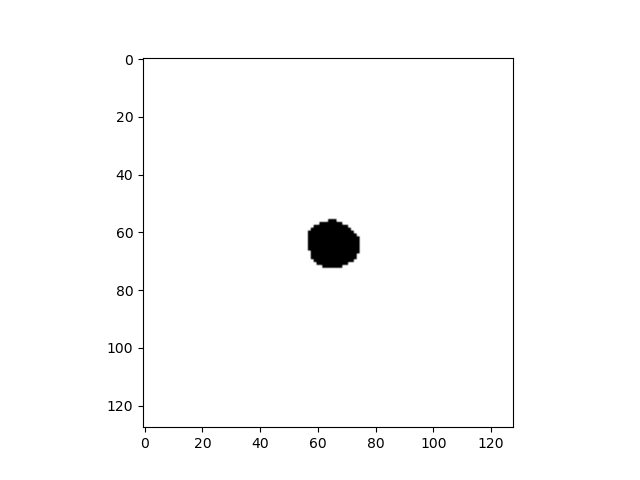}\hfill

    \includegraphics[width=0.2\textwidth]{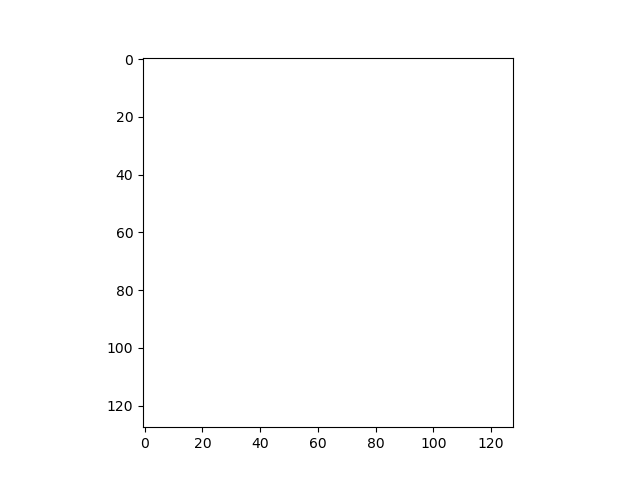}\hfill
    \includegraphics[width=0.2\textwidth]{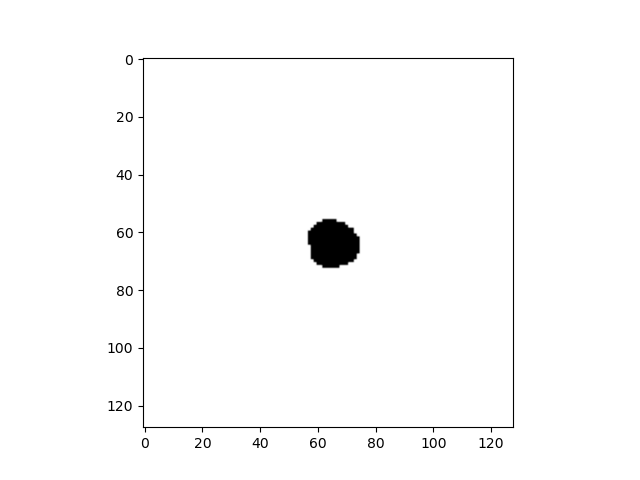}\hfill
    \includegraphics[width=0.2\textwidth]{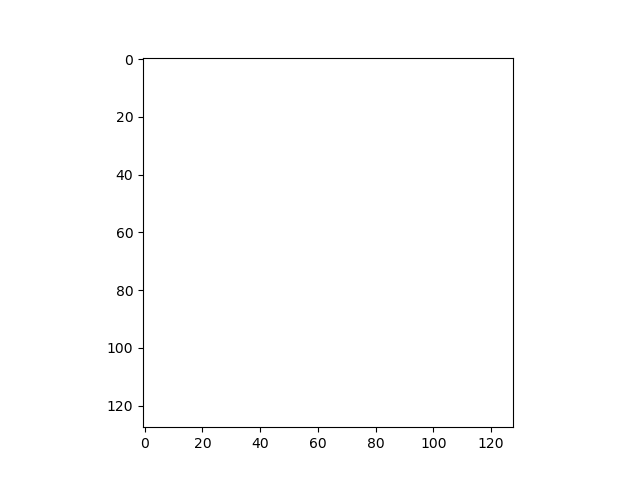}\hfill
    \includegraphics[width=0.2\textwidth]{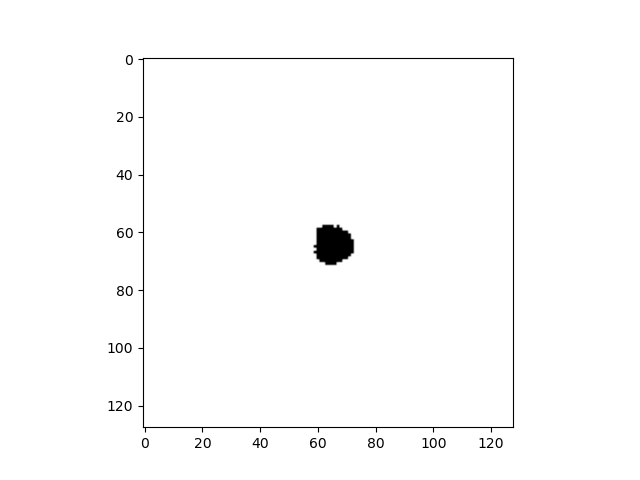}\hfill
    \includegraphics[width=0.2\textwidth]{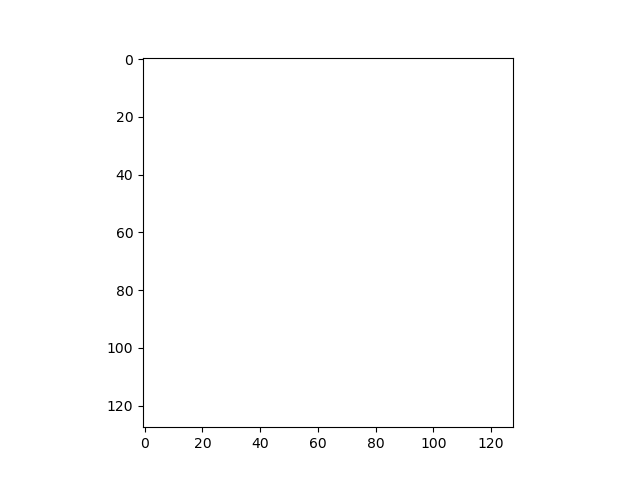}\hfill

    \hrule

    \includegraphics[width=0.2\textwidth]{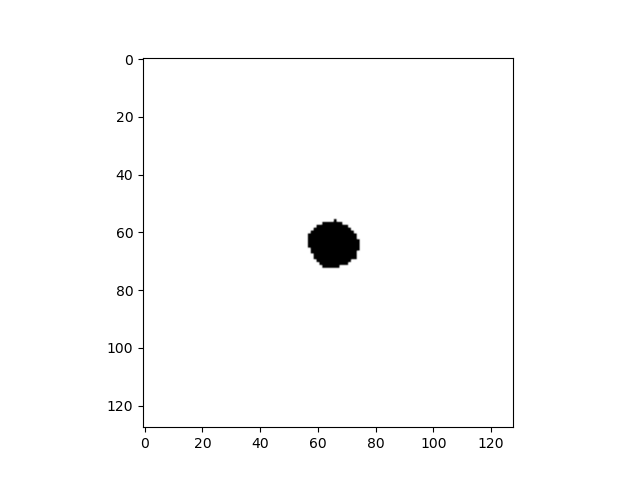}\hfill
    \includegraphics[width=0.2\textwidth]{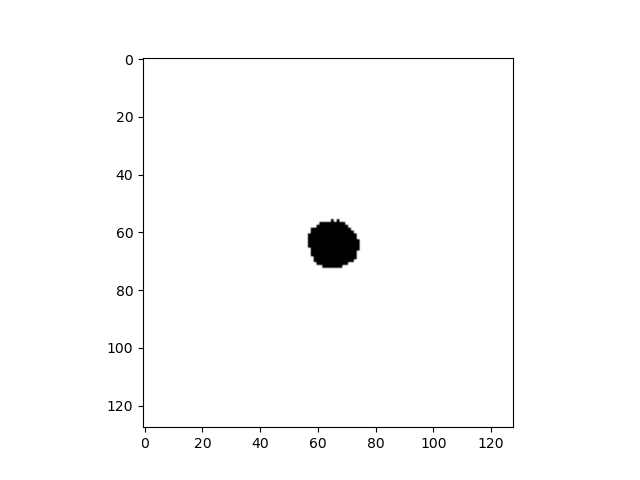}\hfill
    \includegraphics[width=0.2\textwidth]{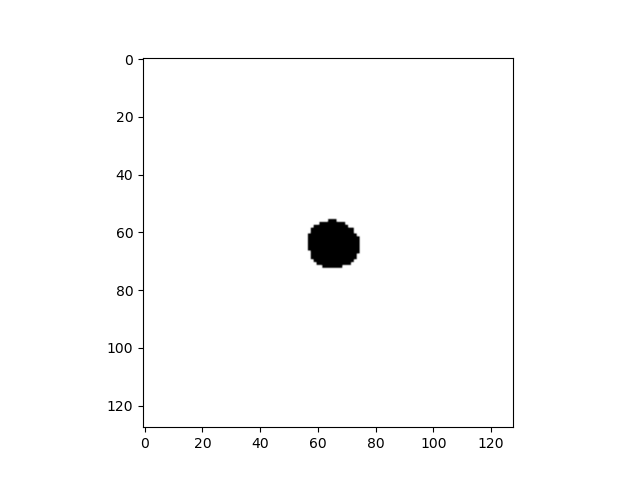}\hfill
    \includegraphics[width=0.2\textwidth]{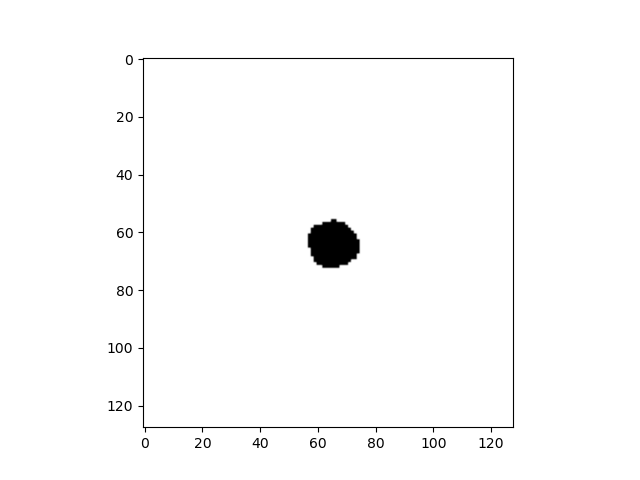}\hfill
    \includegraphics[width=0.2\textwidth]{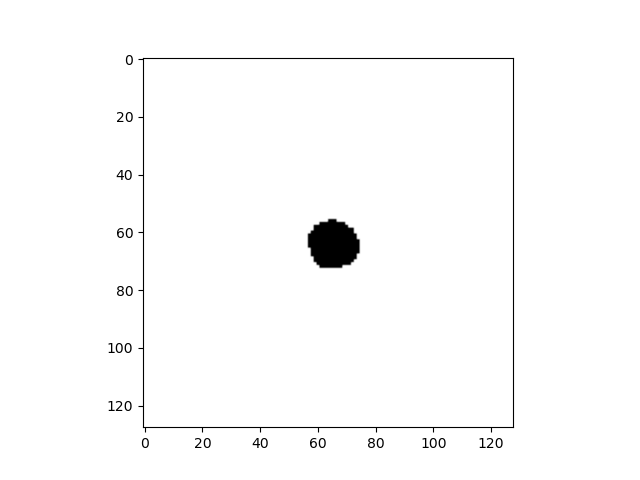}\hfill

    \includegraphics[width=0.2\textwidth]{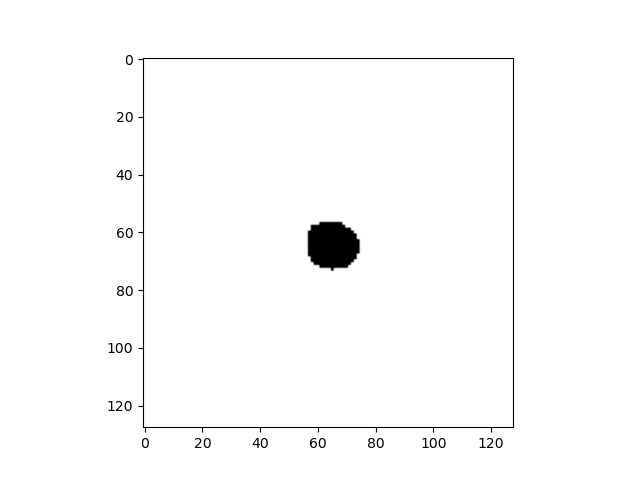}\hfill
    \includegraphics[width=0.2\textwidth]{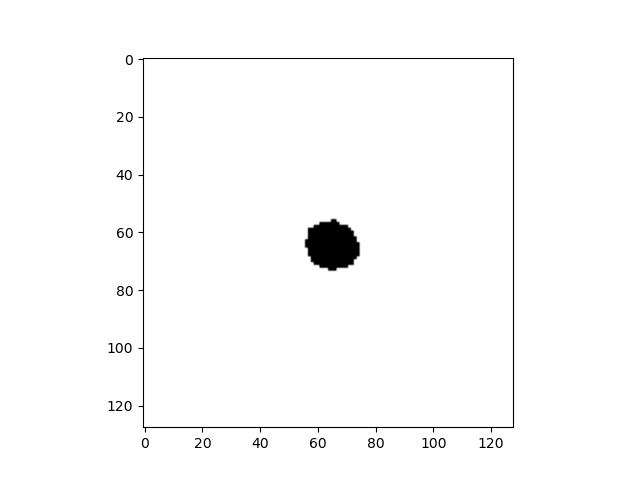}\hfill
    \includegraphics[width=0.2\textwidth]{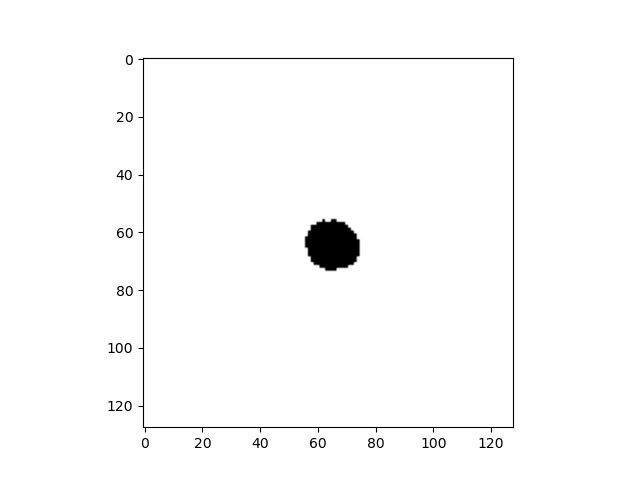}\hfill
    \includegraphics[width=0.2\textwidth]{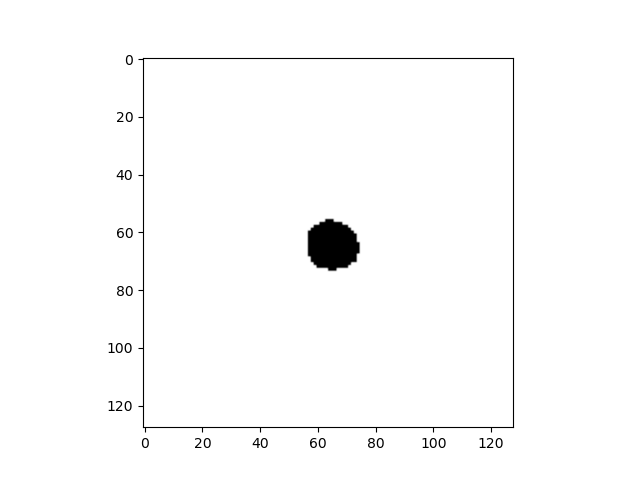}\hfill
    \includegraphics[width=0.2\textwidth]{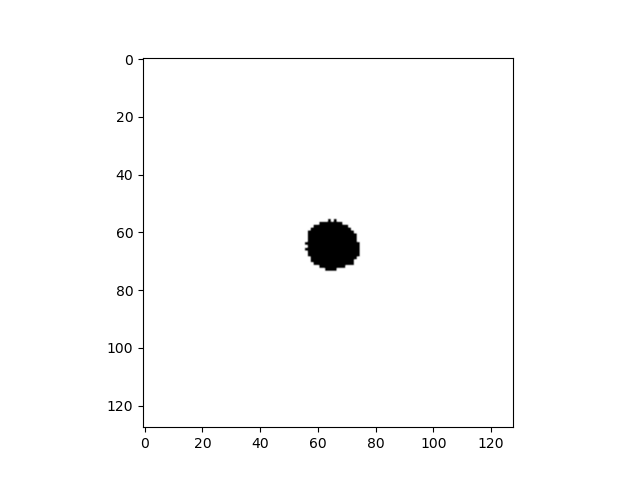}\hfill

    \includegraphics[width=0.2\textwidth]{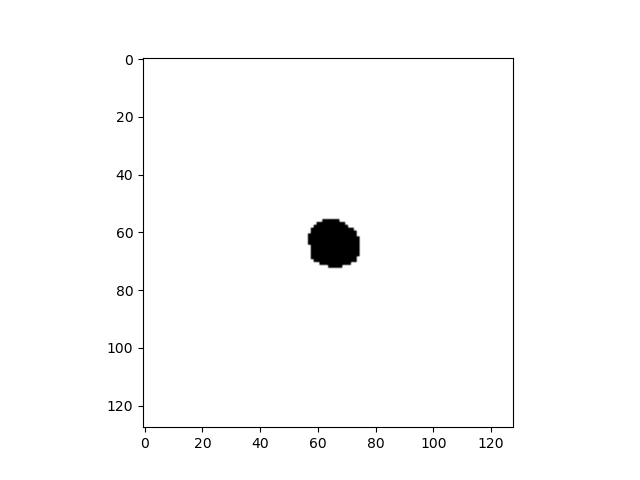}\hfill
    \includegraphics[width=0.2\textwidth]{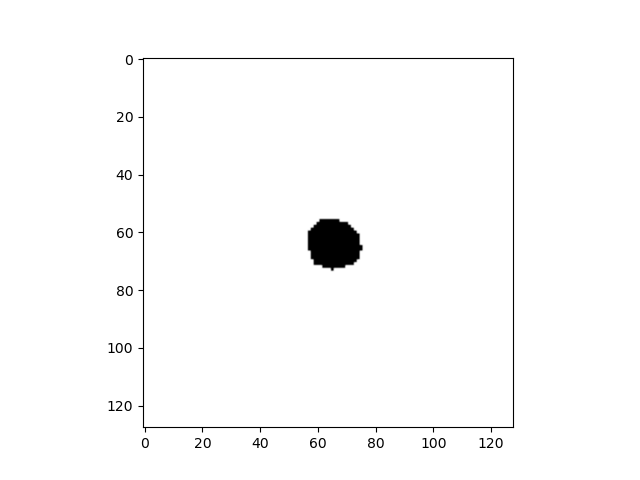}\hfill
    \includegraphics[width=0.2\textwidth]{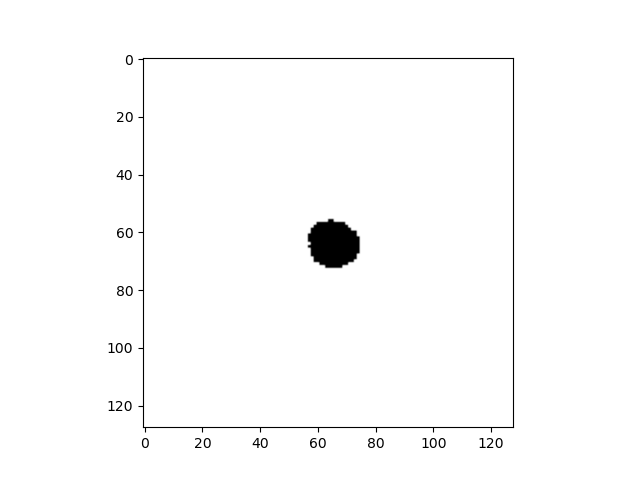}\hfill
    \includegraphics[width=0.2\textwidth]{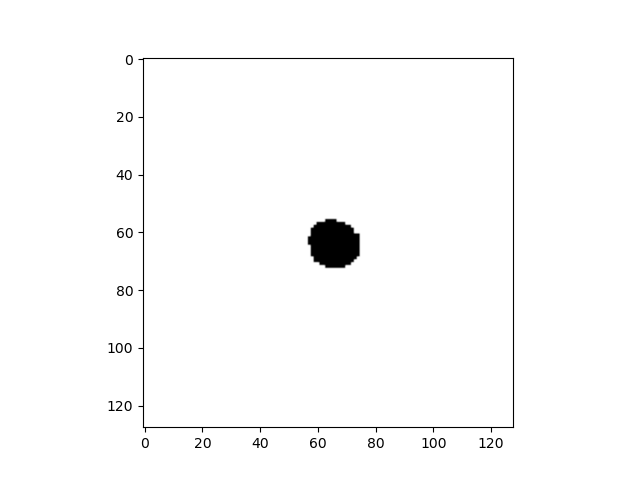}\hfill
    \includegraphics[width=0.2\textwidth]{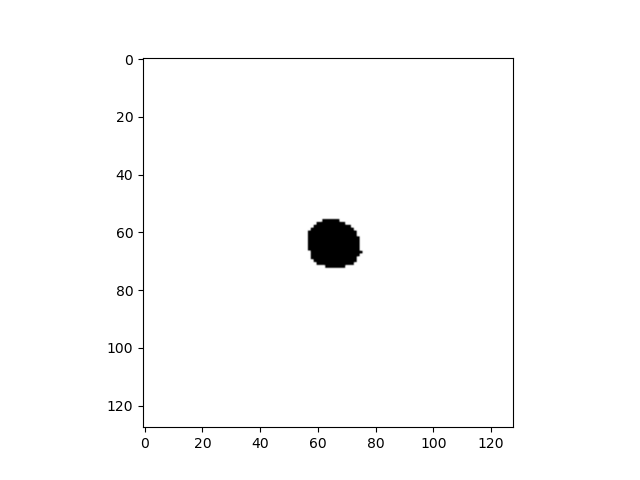}\hfill
    \caption{\textbf{Topleft}: Input image 3 from LIDC data. \textbf{Topright}: Ground truth segmentation. \textbf{2-4 rows}: Segmentation samples from original Probabilistic U-Net. \textbf{5-7 rows}: Segmentation samples from Kendall Shape Probabilistic U-Net. Each row shares the same seed.}
    \label{fig_img3}
\end{figure}

\begin{figure}
    \centering
    \includegraphics[width=0.49\textwidth]{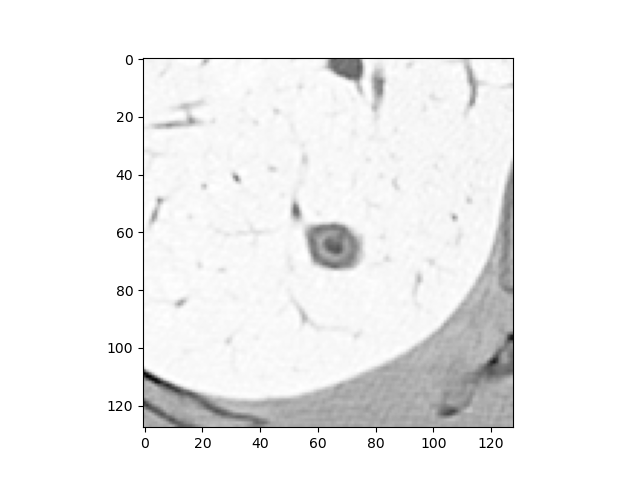}\hfill
    \includegraphics[width=0.49\textwidth]{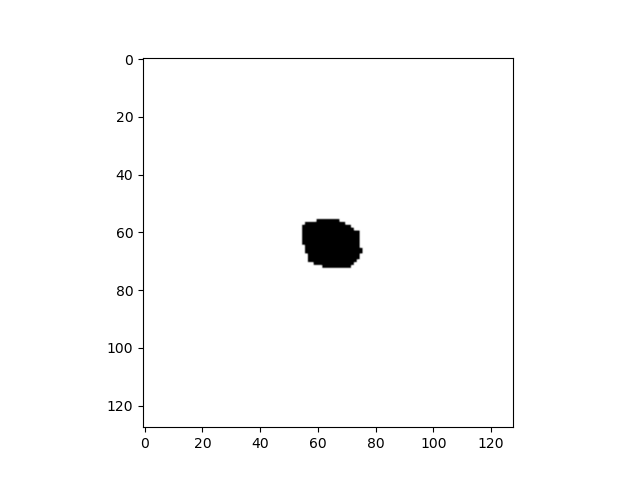}\hfill

    \hrule
    
    \includegraphics[width=0.2\textwidth]{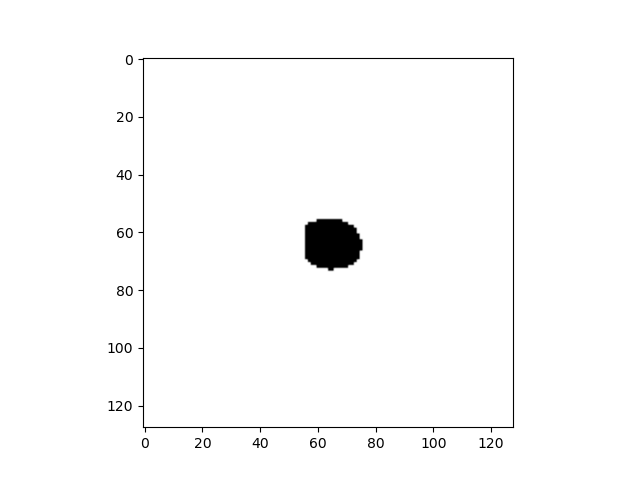}\hfill
    \includegraphics[width=0.2\textwidth]{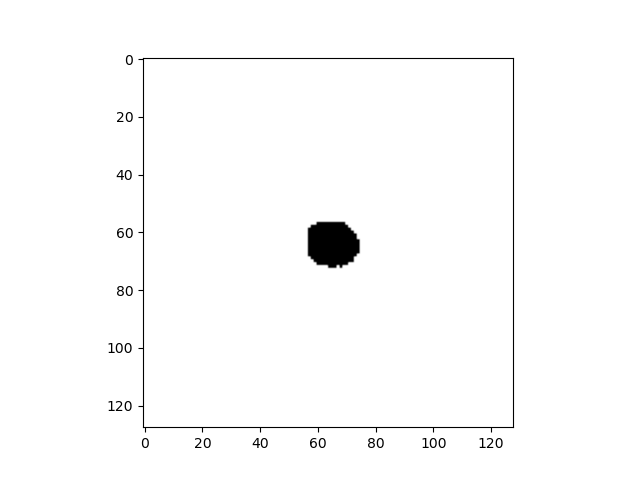}\hfill
    \includegraphics[width=0.2\textwidth]{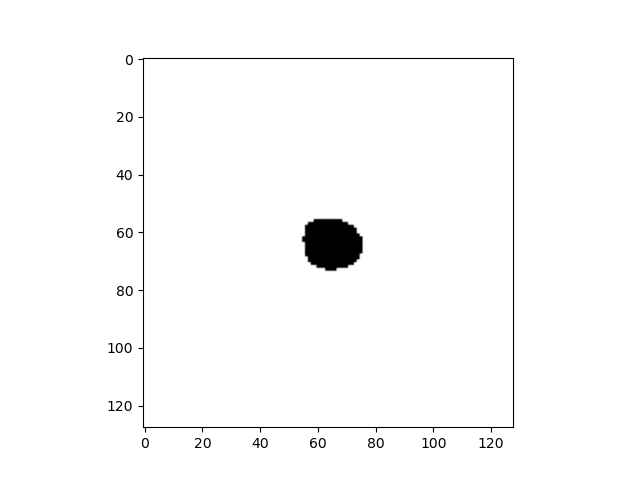}\hfill
    \includegraphics[width=0.2\textwidth]{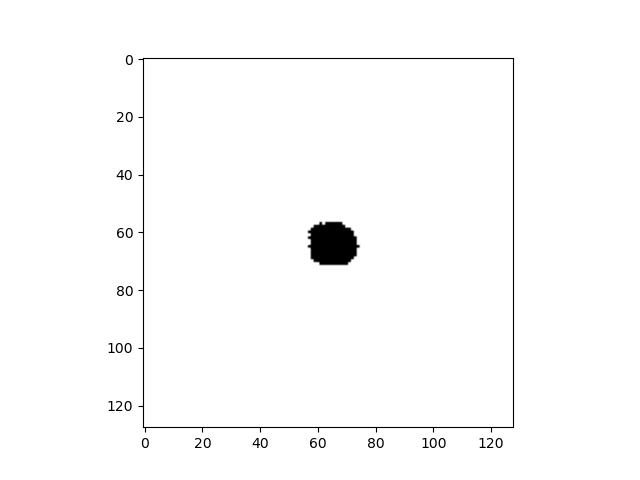}\hfill
    \includegraphics[width=0.2\textwidth]{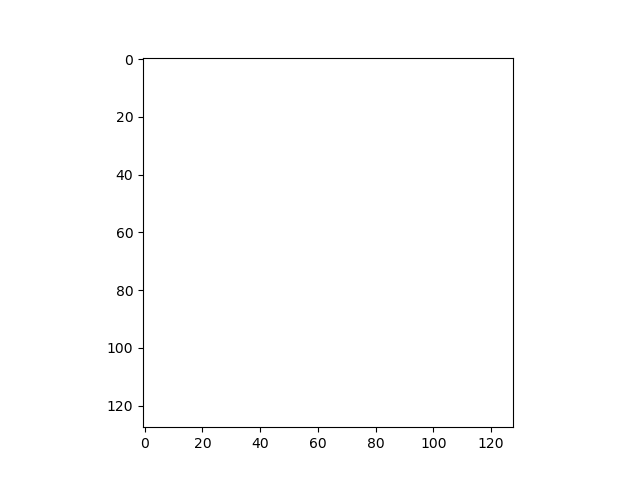}\hfill

    \includegraphics[width=0.2\textwidth]{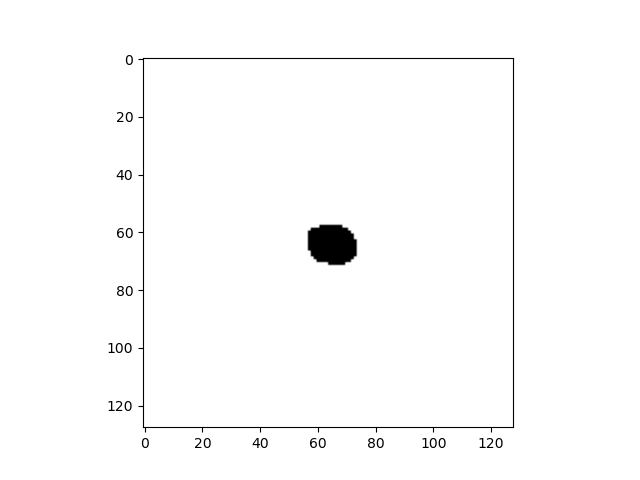}\hfill
    \includegraphics[width=0.2\textwidth]{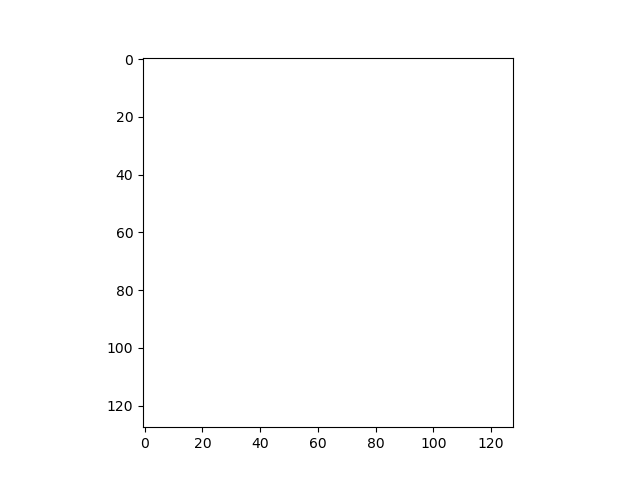}\hfill
    \includegraphics[width=0.2\textwidth]{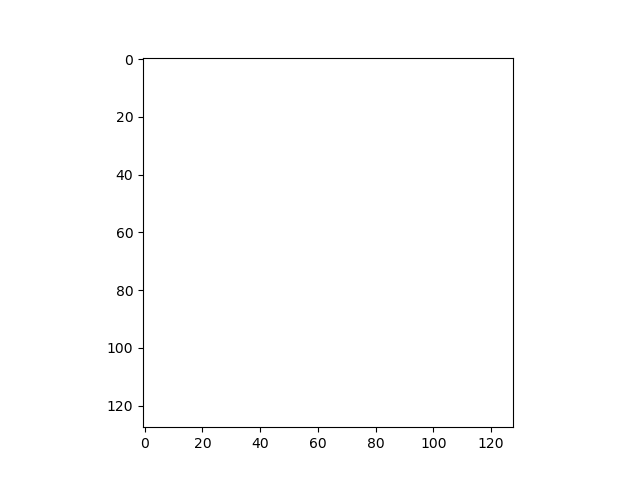}\hfill
    \includegraphics[width=0.2\textwidth]{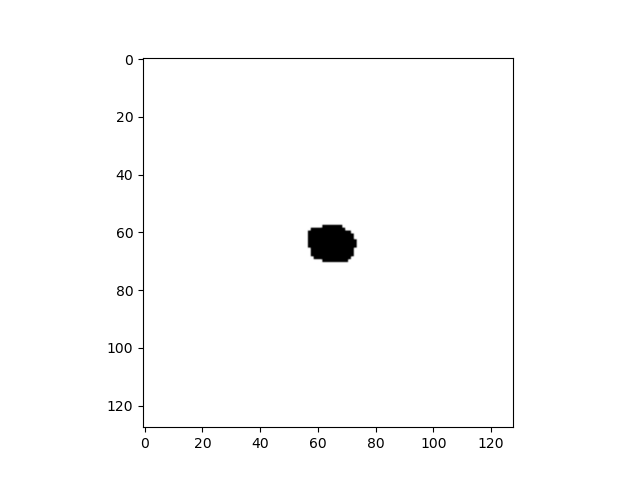}\hfill
    \includegraphics[width=0.2\textwidth]{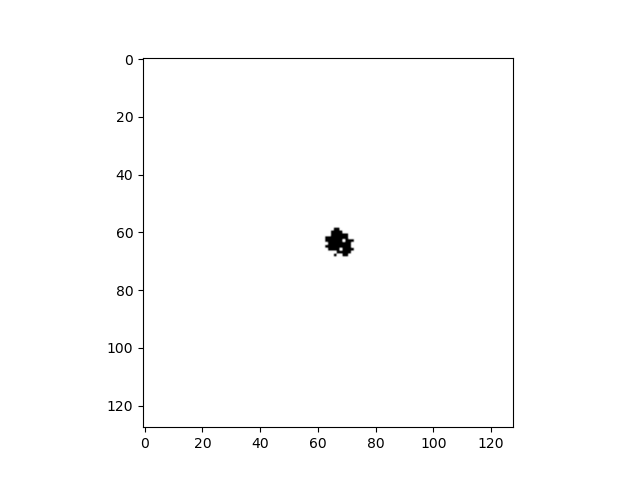}\hfill

    \includegraphics[width=0.2\textwidth]{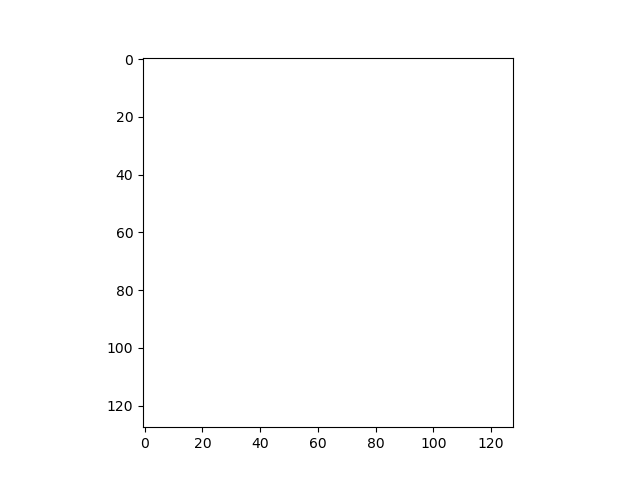}\hfill
    \includegraphics[width=0.2\textwidth]{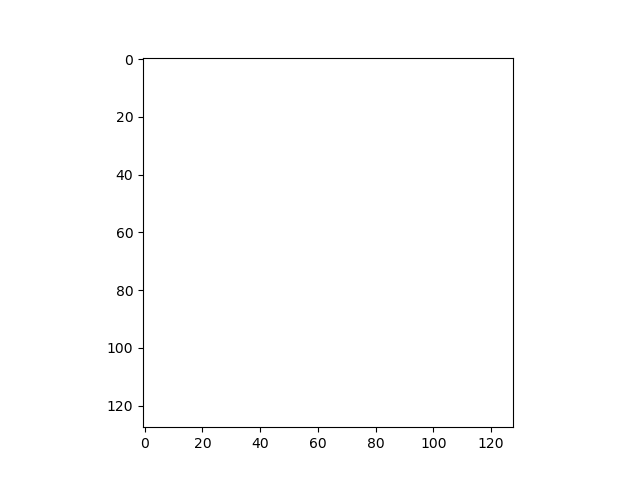}\hfill
    \includegraphics[width=0.2\textwidth]{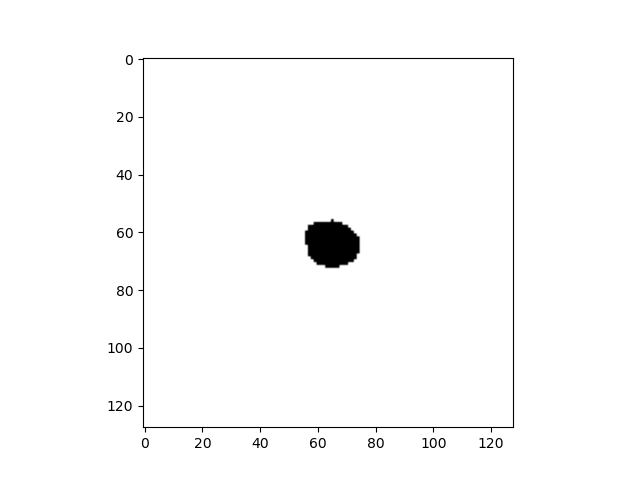}\hfill
    \includegraphics[width=0.2\textwidth]{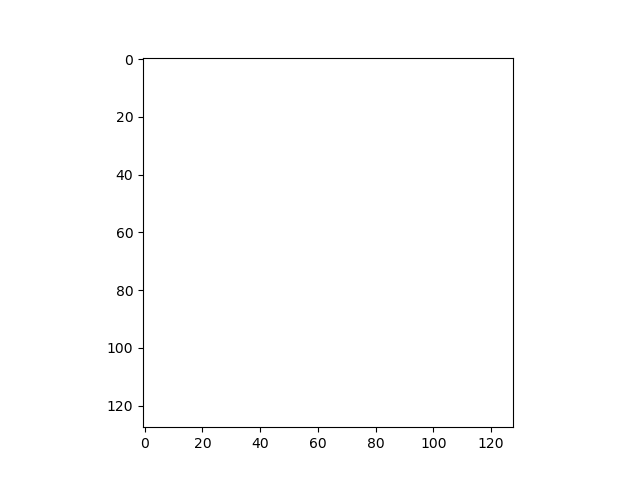}\hfill
    \includegraphics[width=0.2\textwidth]{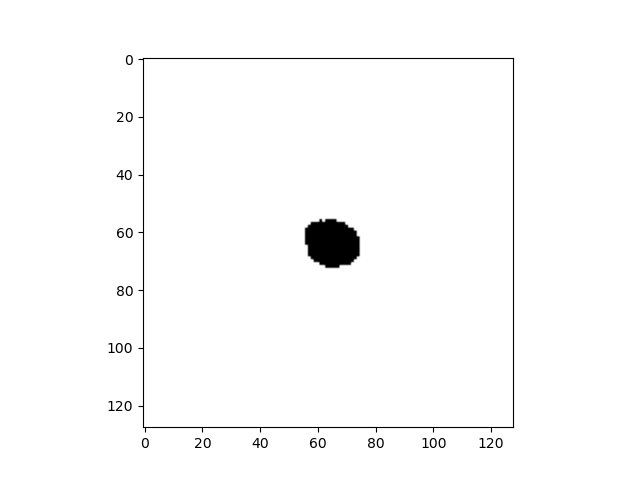}\hfill

    \hrule

    \includegraphics[width=0.2\textwidth]{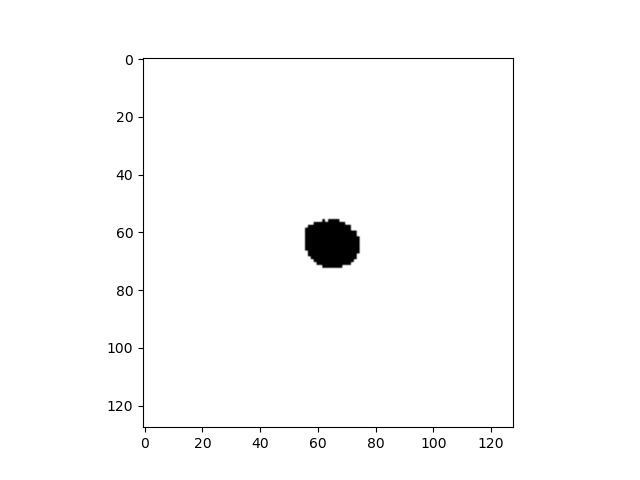}\hfill
    \includegraphics[width=0.2\textwidth]{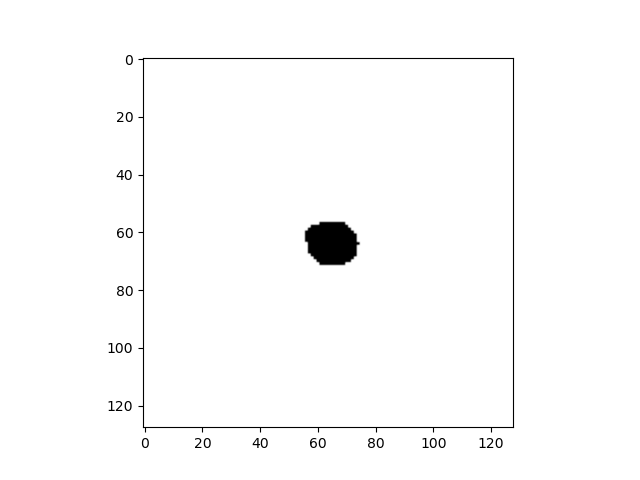}\hfill
    \includegraphics[width=0.2\textwidth]{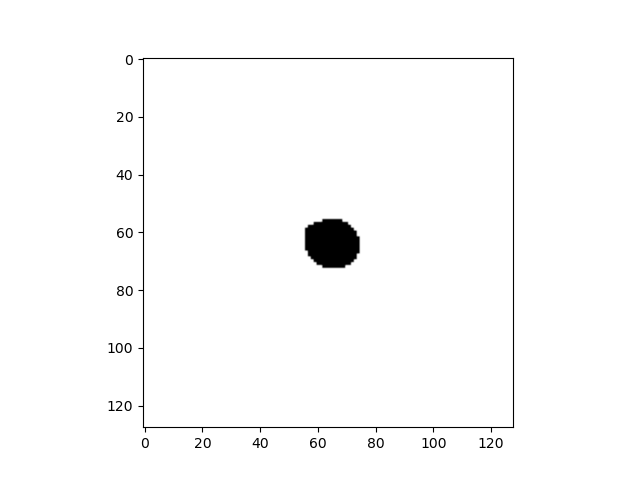}\hfill
    \includegraphics[width=0.2\textwidth]{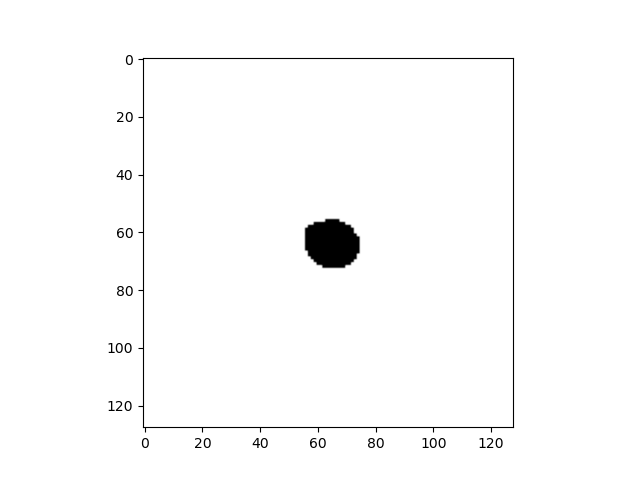}\hfill
    \includegraphics[width=0.2\textwidth]{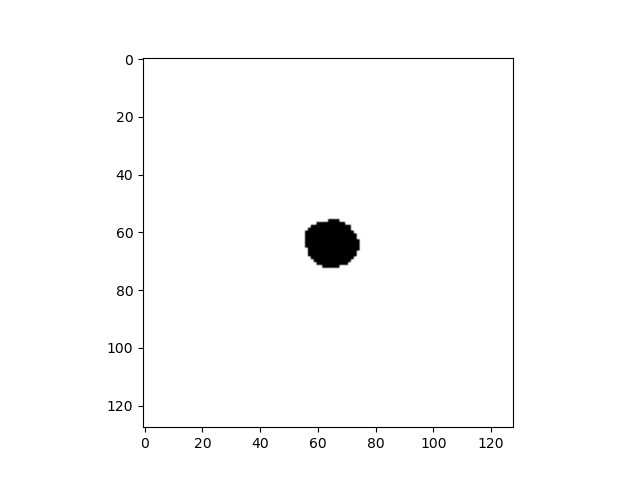}\hfill

    \includegraphics[width=0.2\textwidth]{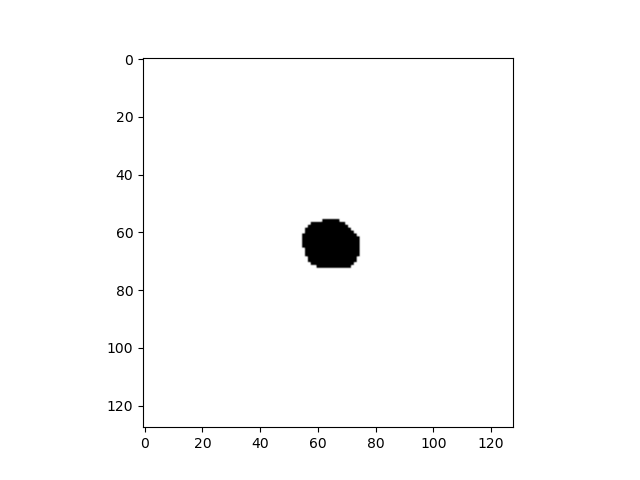}\hfill
    \includegraphics[width=0.2\textwidth]{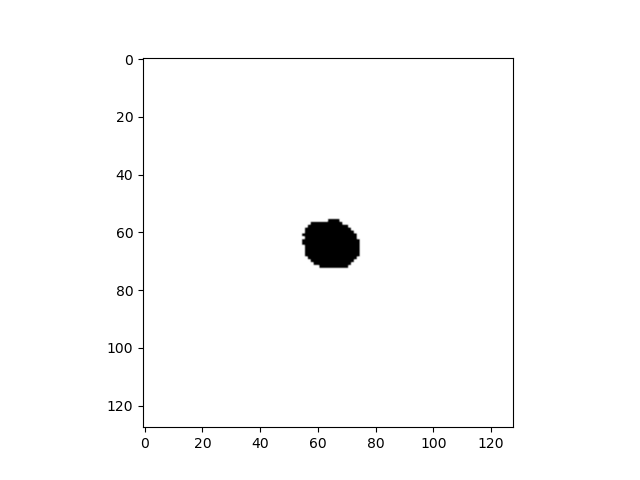}\hfill
    \includegraphics[width=0.2\textwidth]{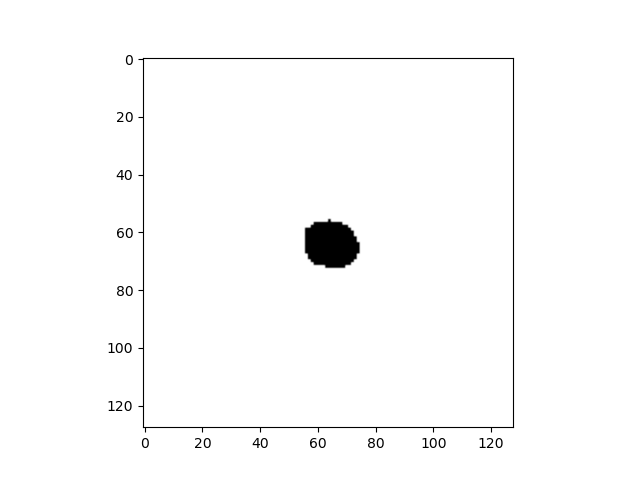}\hfill
    \includegraphics[width=0.2\textwidth]{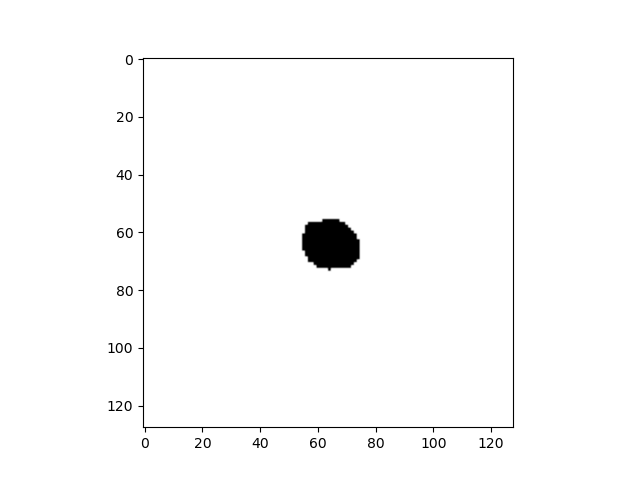}\hfill
    \includegraphics[width=0.2\textwidth]{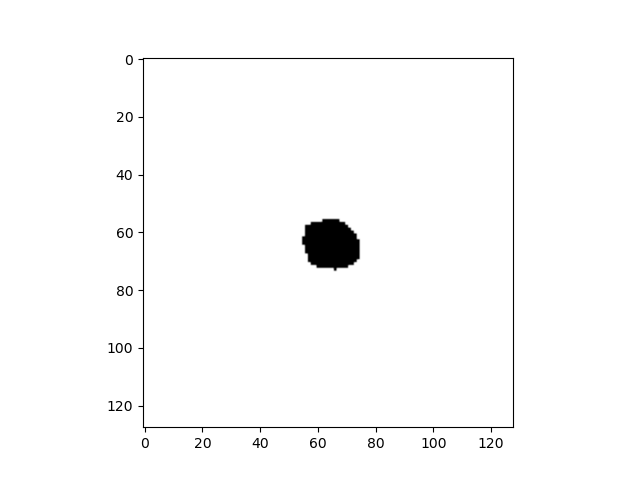}\hfill

    \includegraphics[width=0.2\textwidth]{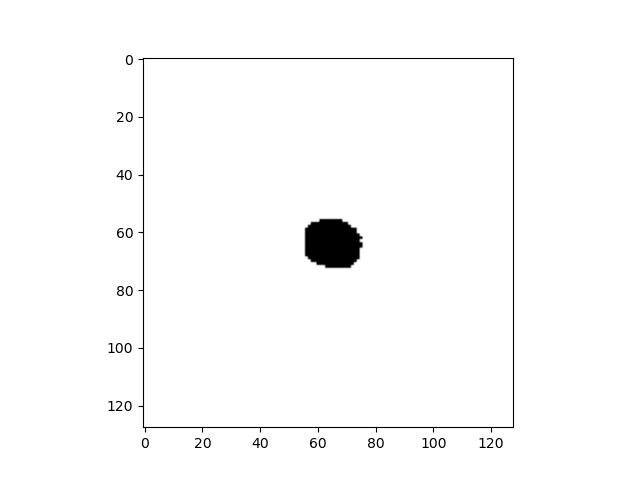}\hfill
    \includegraphics[width=0.2\textwidth]{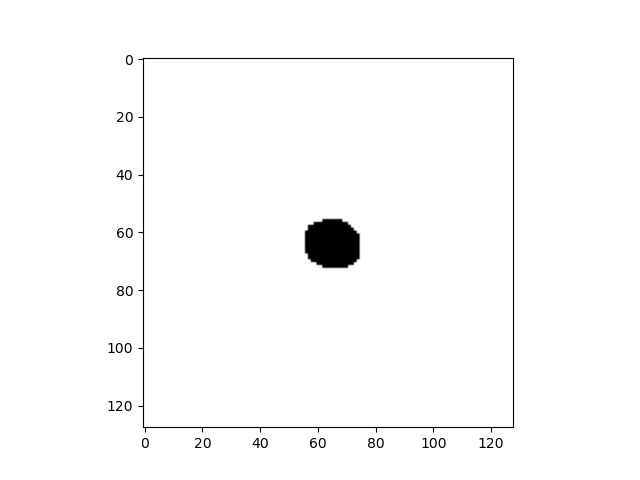}\hfill
    \includegraphics[width=0.2\textwidth]{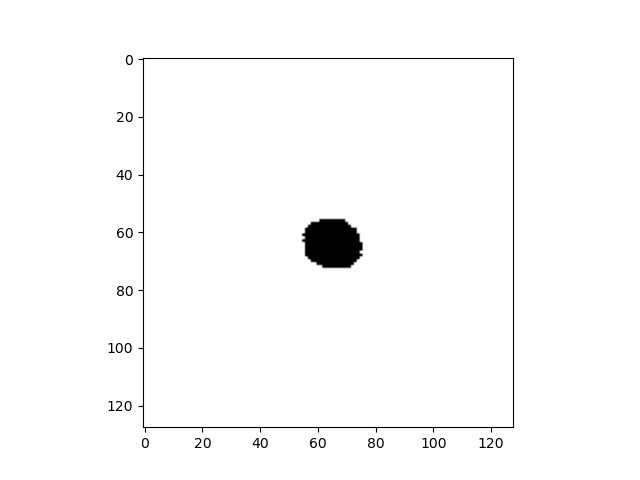}\hfill
    \includegraphics[width=0.2\textwidth]{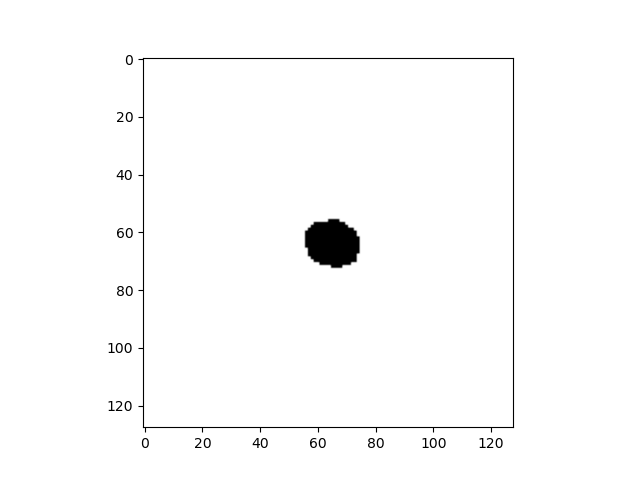}\hfill
    \includegraphics[width=0.2\textwidth]{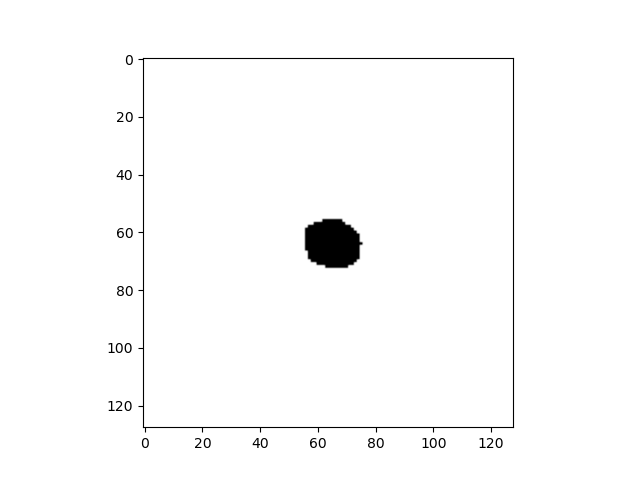}\hfill
    \caption{\textbf{Topleft}: Input image 4 from LIDC data. \textbf{Topright}: Ground truth segmentation. \textbf{2-4 rows}: Segmentation samples from original Probabilistic U-Net. \textbf{5-7 rows}: Segmentation samples from Kendall Shape Probabilistic U-Net. Each row shares the same seed.}
    \label{fig_img4}
\end{figure}

\begin{figure}
    \centering
    \includegraphics[width=0.49\textwidth]{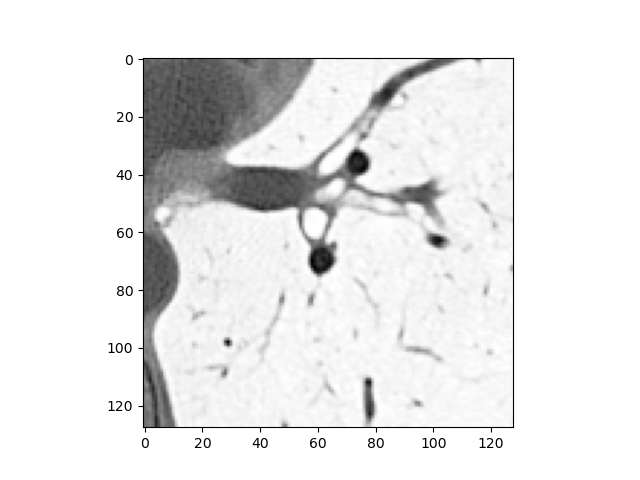}\hfill
    \includegraphics[width=0.49\textwidth]{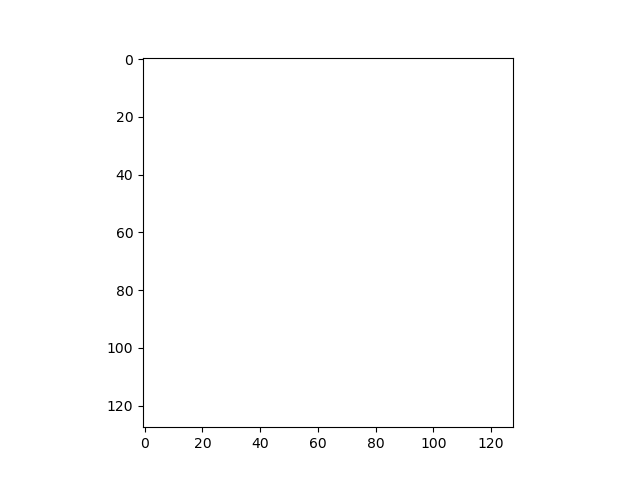}\hfill

    \hrule
    
    \includegraphics[width=0.2\textwidth]{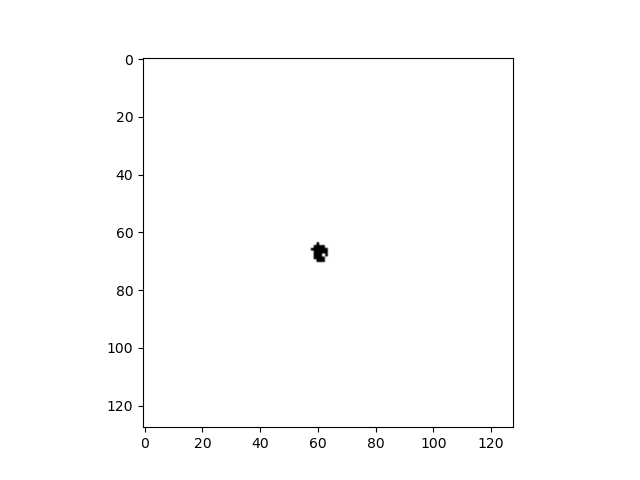}\hfill
    \includegraphics[width=0.2\textwidth]{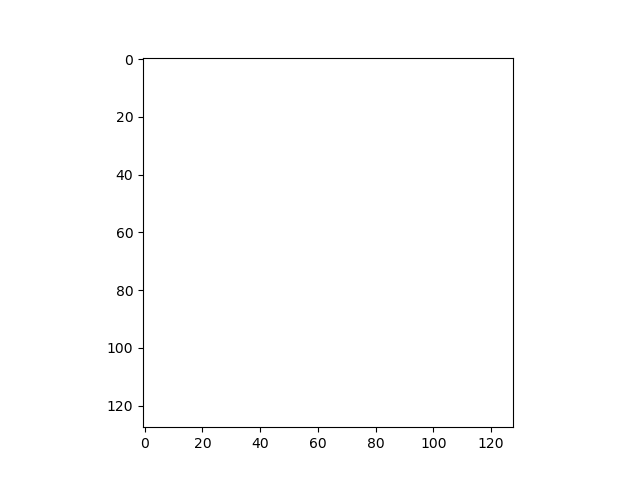}\hfill
    \includegraphics[width=0.2\textwidth]{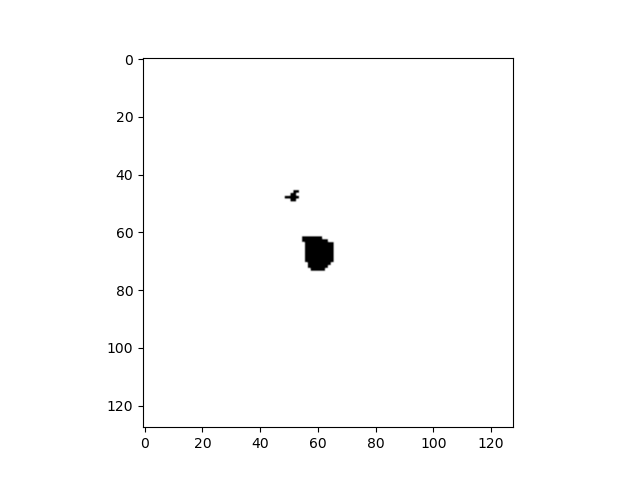}\hfill
    \includegraphics[width=0.2\textwidth]{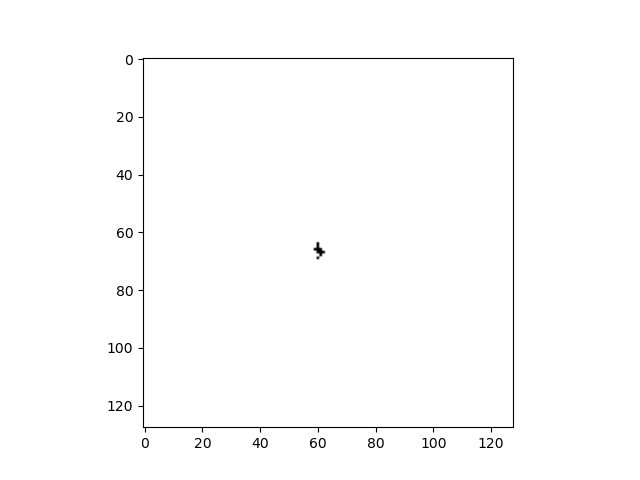}\hfill
    \includegraphics[width=0.2\textwidth]{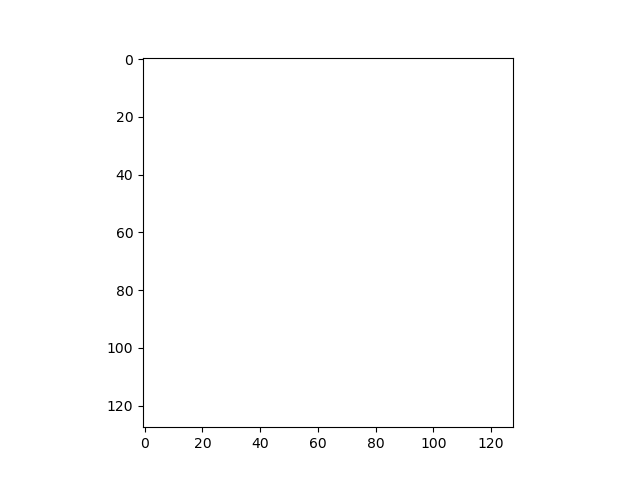}\hfill

    \includegraphics[width=0.2\textwidth]{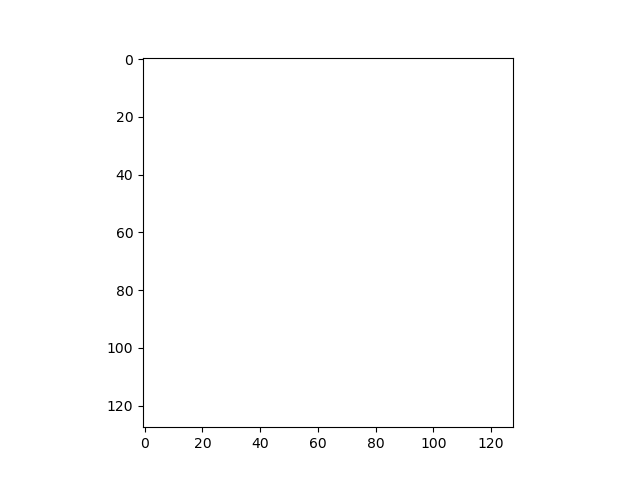}\hfill
    \includegraphics[width=0.2\textwidth]{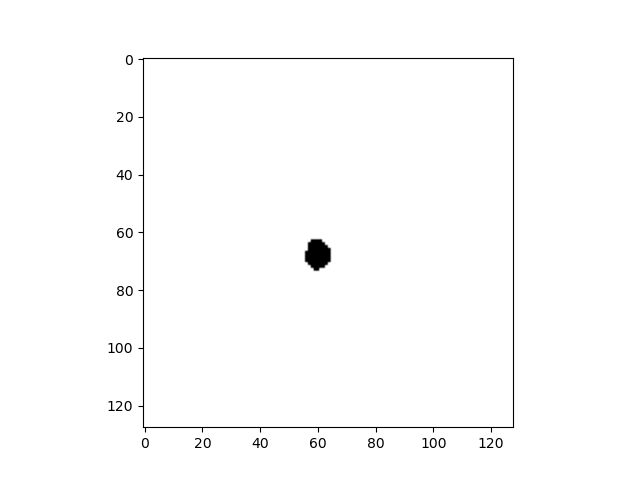}\hfill
    \includegraphics[width=0.2\textwidth]{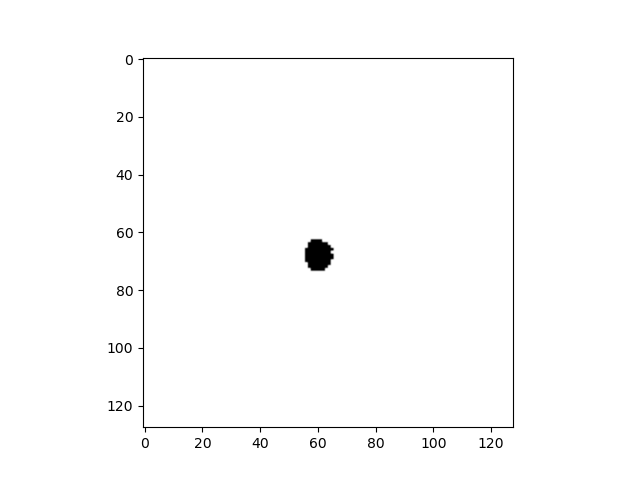}\hfill
    \includegraphics[width=0.2\textwidth]{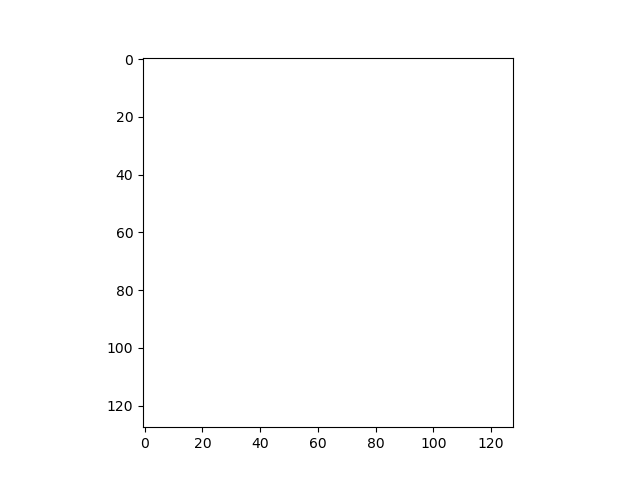}\hfill
    \includegraphics[width=0.2\textwidth]{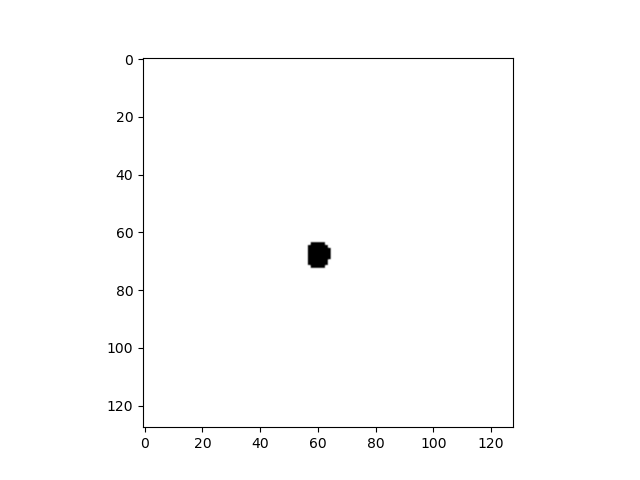}\hfill

    \includegraphics[width=0.2\textwidth]{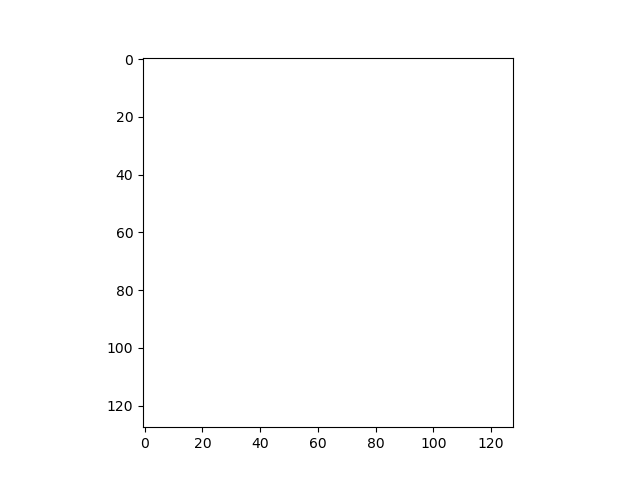}\hfill
    \includegraphics[width=0.2\textwidth]{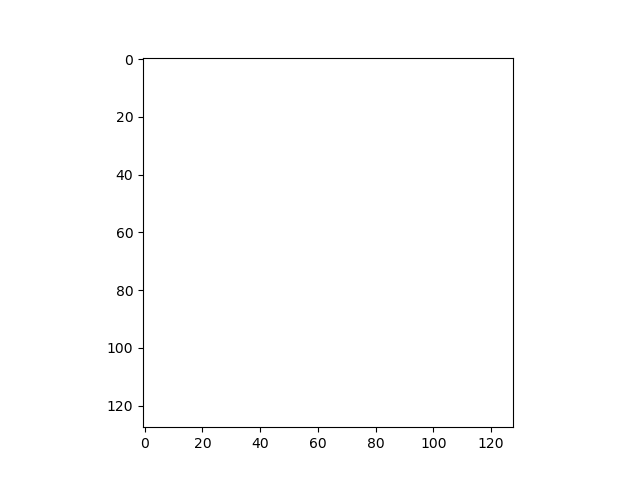}\hfill
    \includegraphics[width=0.2\textwidth]{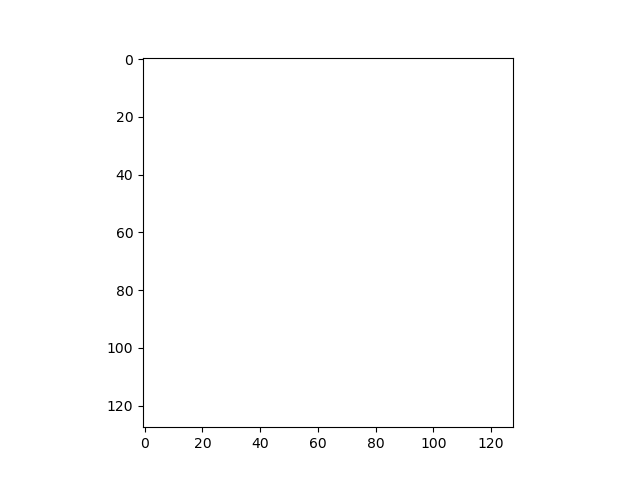}\hfill
    \includegraphics[width=0.2\textwidth]{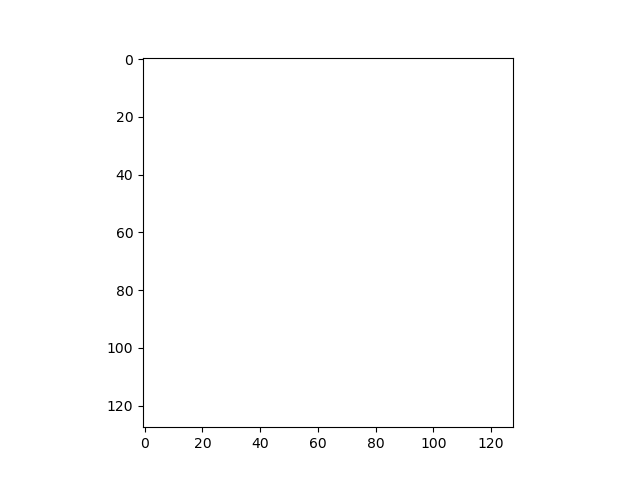}\hfill
    \includegraphics[width=0.2\textwidth]{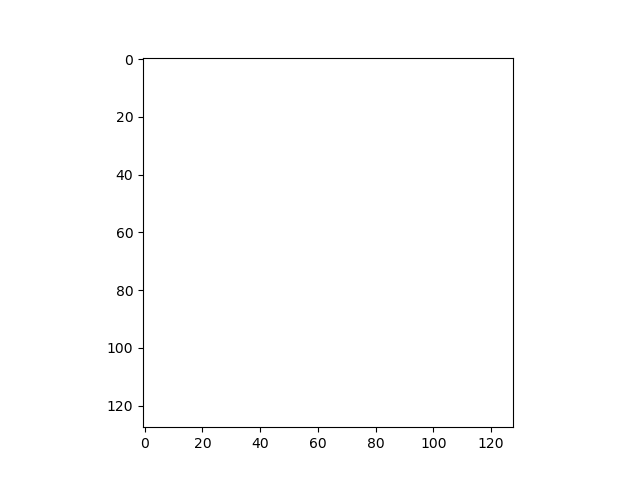}\hfill

    \hrule

    \includegraphics[width=0.2\textwidth]{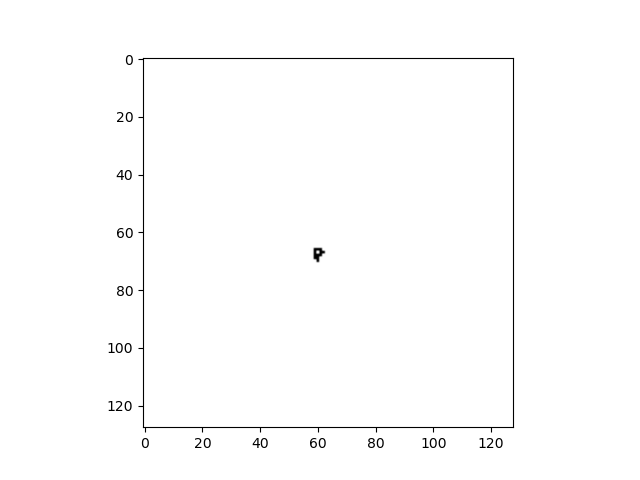}\hfill
    \includegraphics[width=0.2\textwidth]{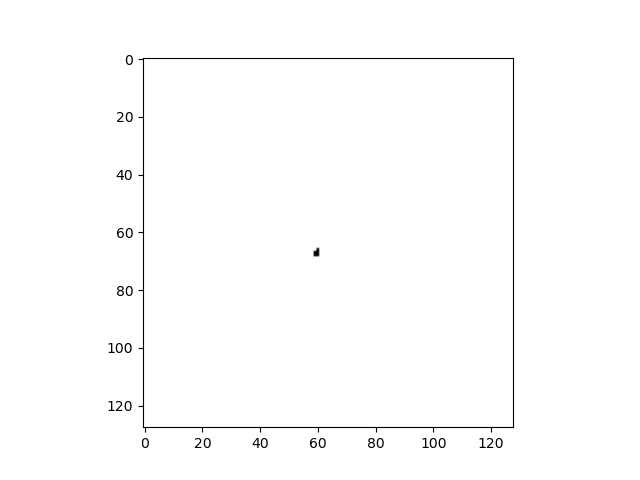}\hfill
    \includegraphics[width=0.2\textwidth]{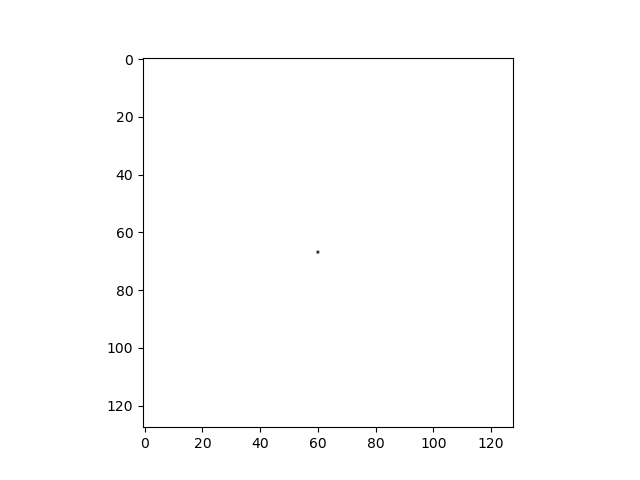}\hfill
    \includegraphics[width=0.2\textwidth]{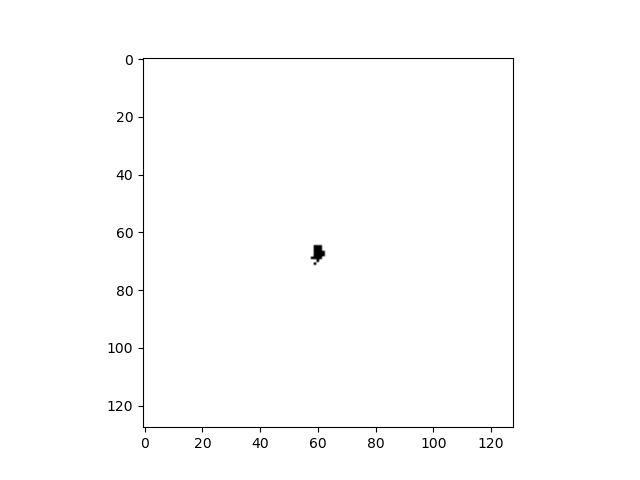}\hfill
    \includegraphics[width=0.2\textwidth]{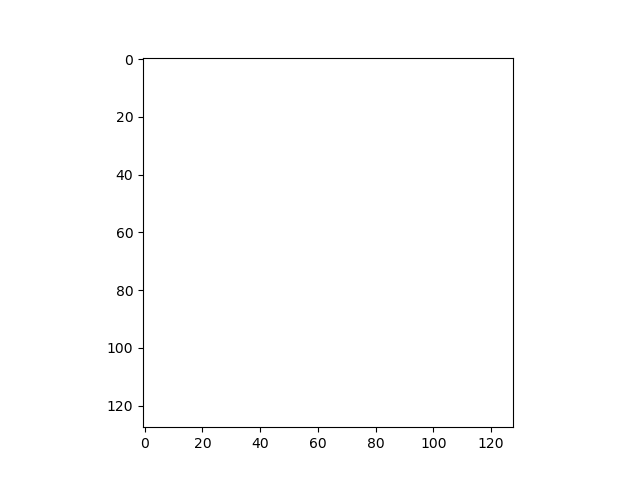}\hfill

    \includegraphics[width=0.2\textwidth]{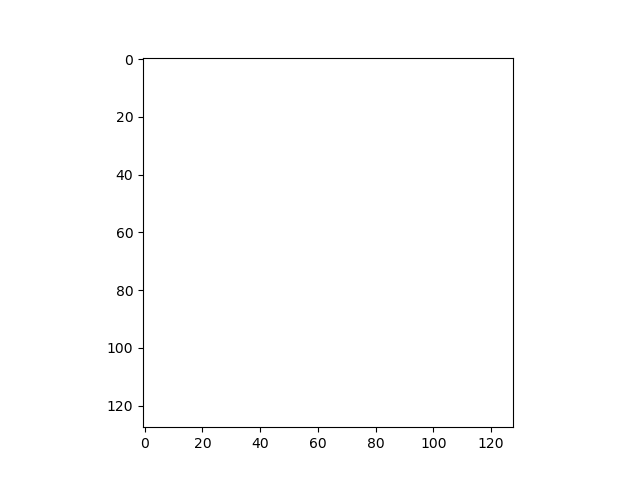}\hfill
    \includegraphics[width=0.2\textwidth]{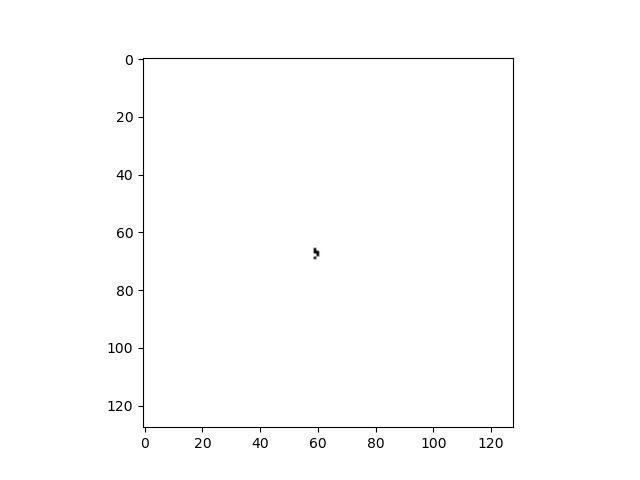}\hfill
    \includegraphics[width=0.2\textwidth]{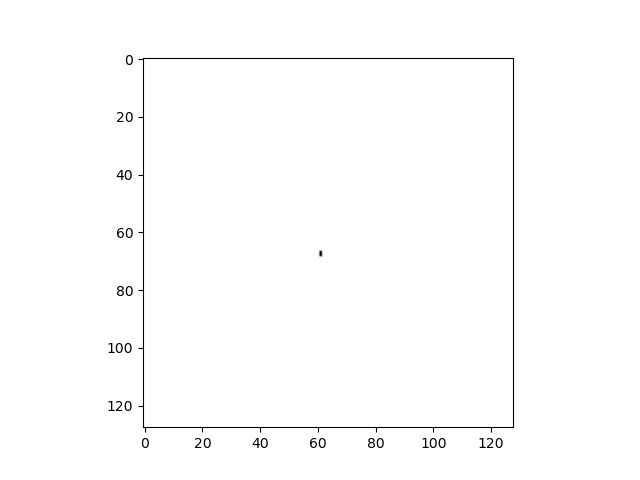}\hfill
    \includegraphics[width=0.2\textwidth]{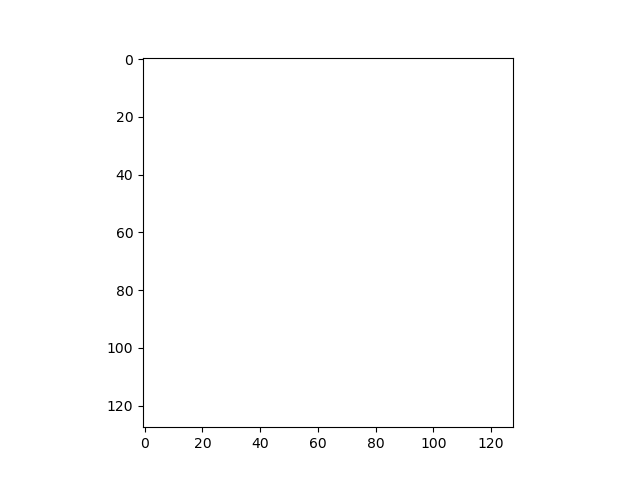}\hfill
    \includegraphics[width=0.2\textwidth]{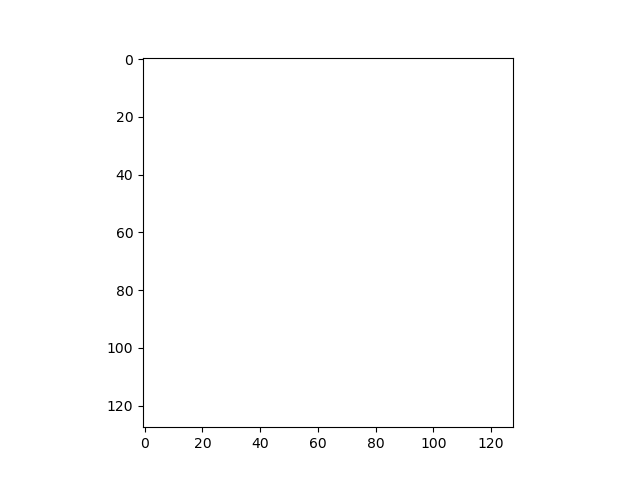}\hfill

    \includegraphics[width=0.2\textwidth]{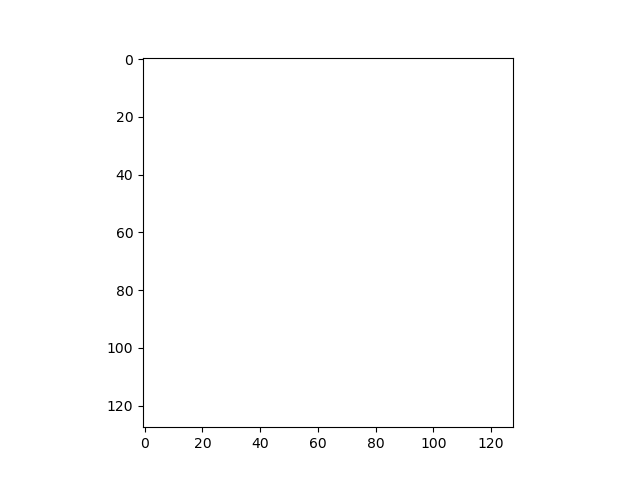}\hfill
    \includegraphics[width=0.2\textwidth]{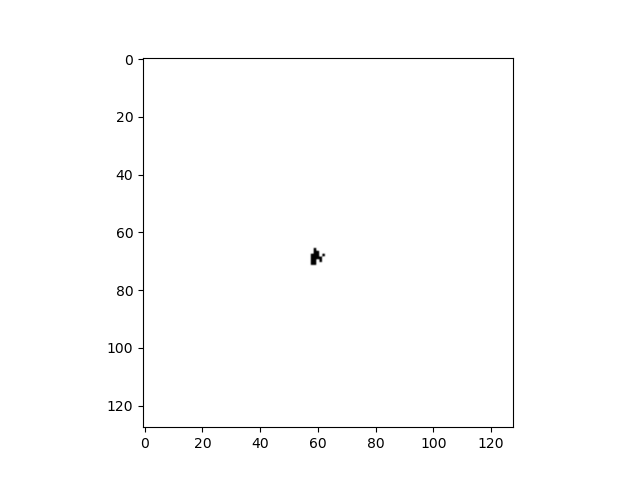}\hfill
    \includegraphics[width=0.2\textwidth]{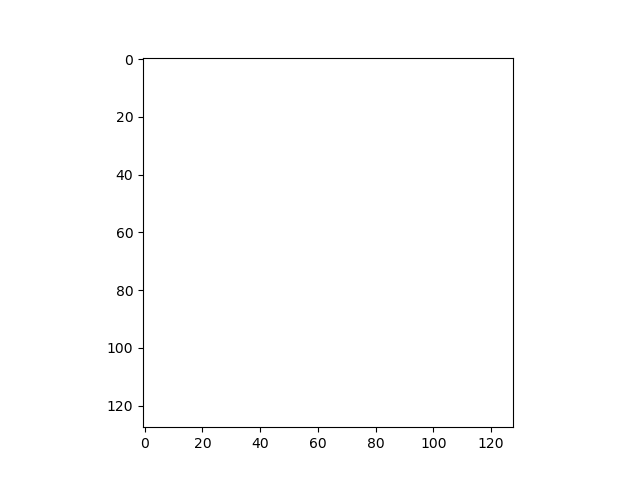}\hfill
    \includegraphics[width=0.2\textwidth]{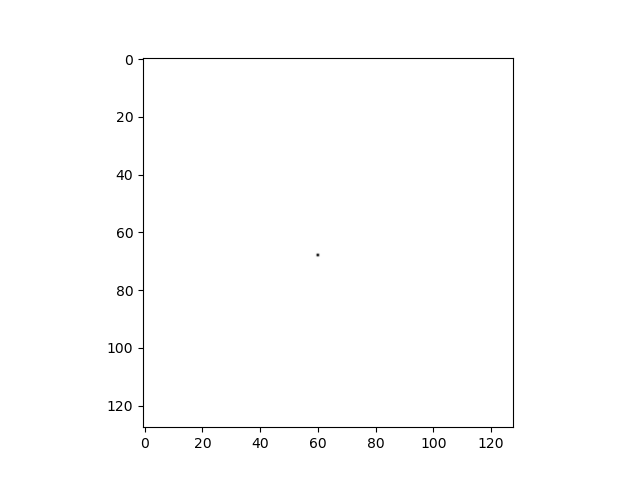}\hfill
    \includegraphics[width=0.2\textwidth]{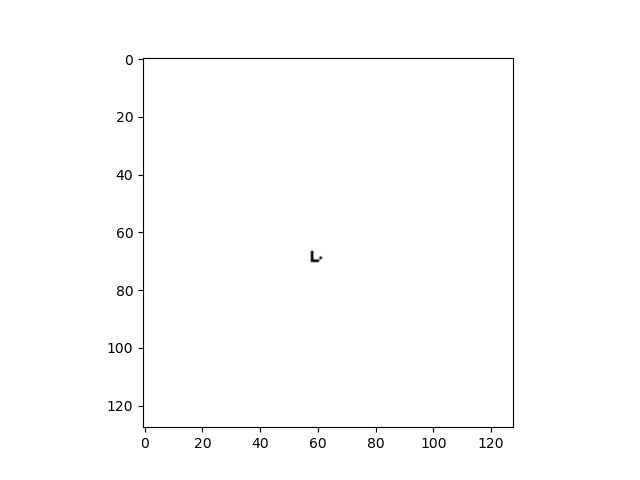}\hfill
    \caption{\textbf{Topleft}: Input image 5 from LIDC data. \textbf{Topright}: Ground truth segmentation. \textbf{2-4 rows}: Segmentation samples from original Probabilistic U-Net. \textbf{5-7 rows}: Segmentation samples from Kendall Shape Probabilistic U-Net. Each row shares the same seed.}
    \label{fig_img5}
\end{figure}

\begin{figure}
    \centering
    \includegraphics[width=0.49\textwidth]{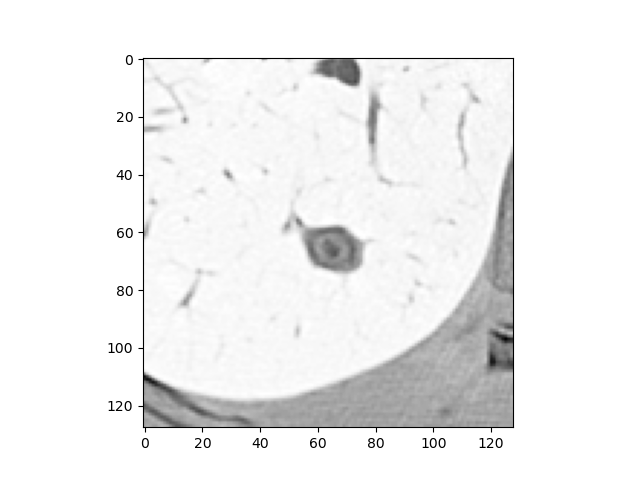}\hfill
    \includegraphics[width=0.49\textwidth]{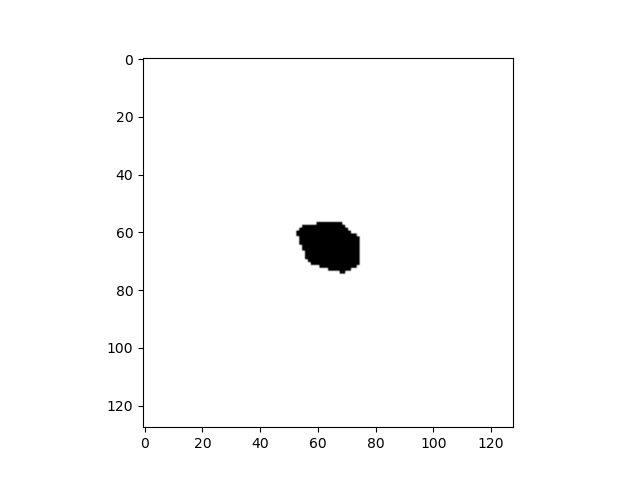}\hfill

    \hrule
    
    \includegraphics[width=0.2\textwidth]{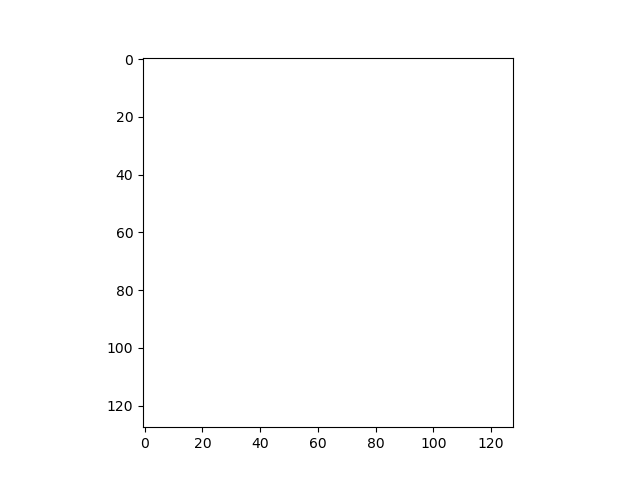}\hfill
    \includegraphics[width=0.2\textwidth]{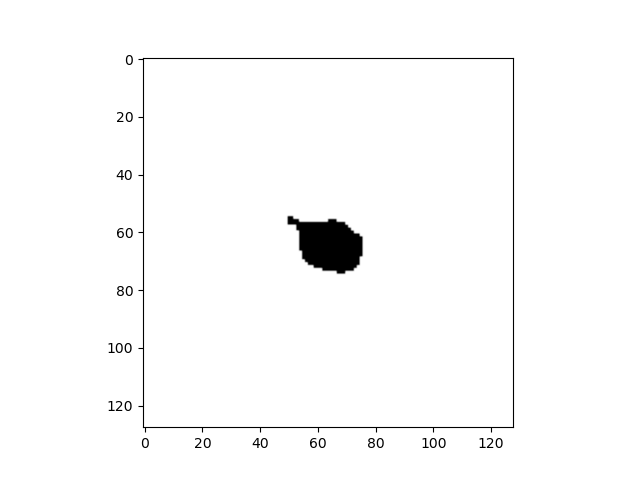}\hfill
    \includegraphics[width=0.2\textwidth]{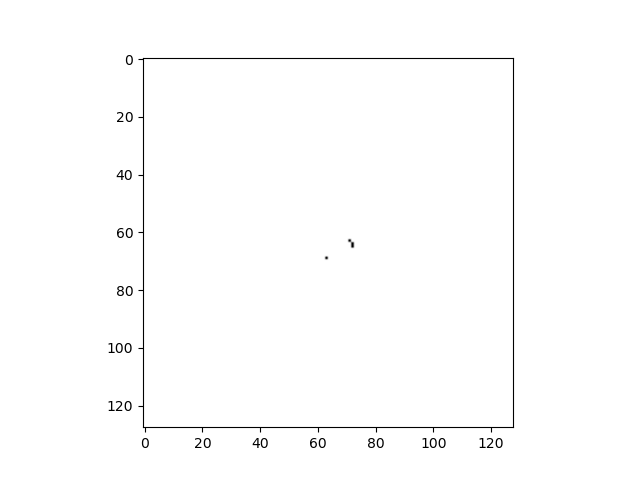}\hfill
    \includegraphics[width=0.2\textwidth]{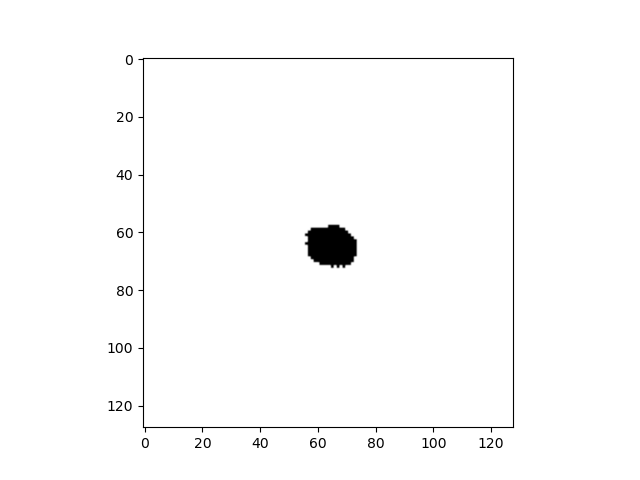}\hfill
    \includegraphics[width=0.2\textwidth]{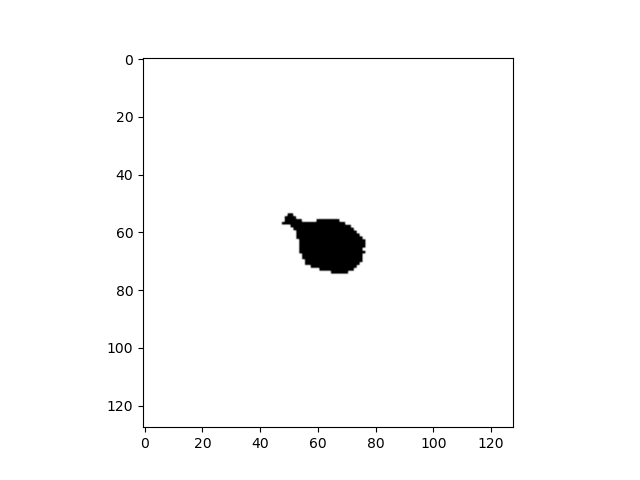}\hfill

    \includegraphics[width=0.2\textwidth]{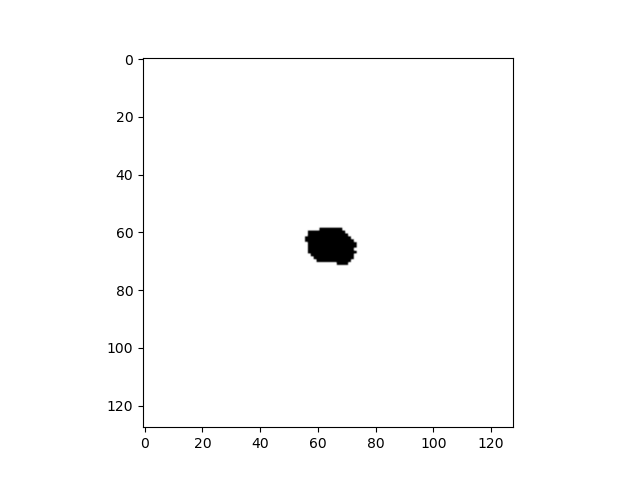}\hfill
    \includegraphics[width=0.2\textwidth]{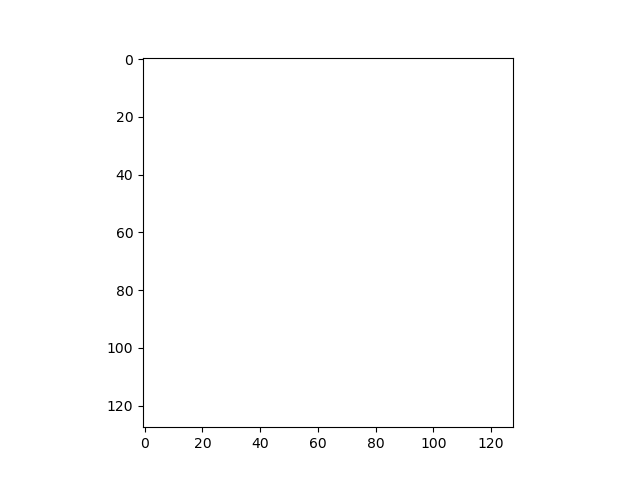}\hfill
    \includegraphics[width=0.2\textwidth]{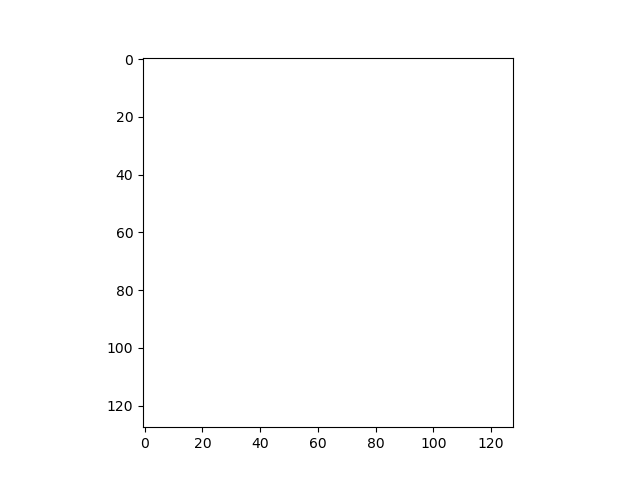}\hfill
    \includegraphics[width=0.2\textwidth]{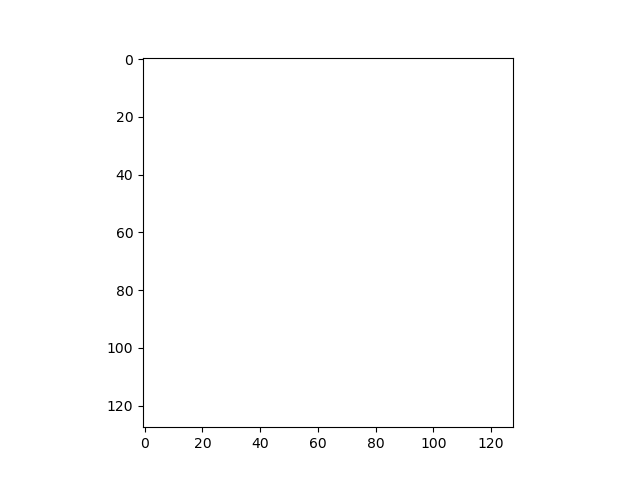}\hfill
    \includegraphics[width=0.2\textwidth]{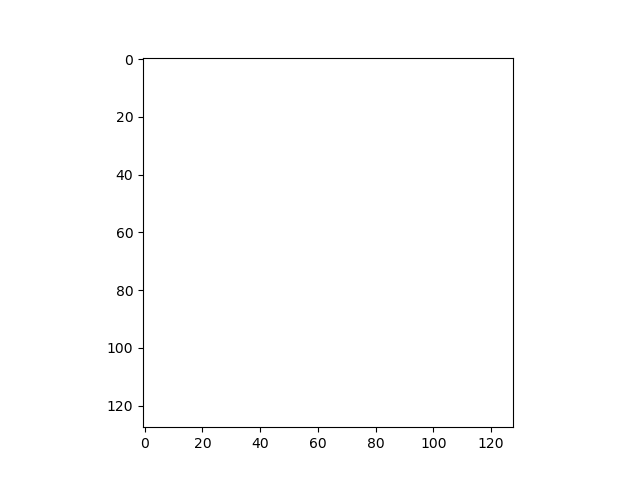}\hfill

    \includegraphics[width=0.2\textwidth]{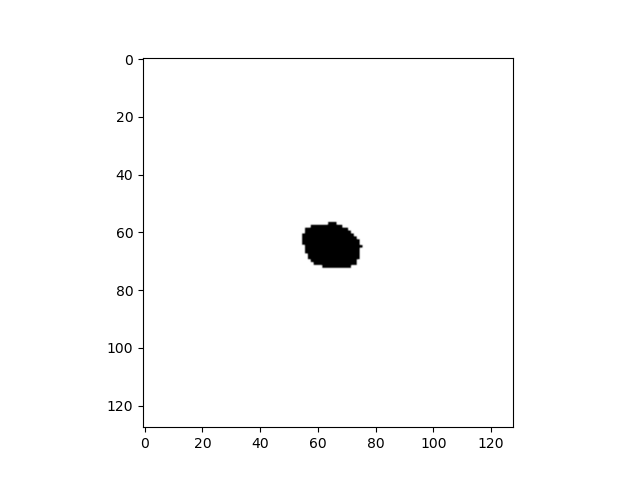}\hfill
    \includegraphics[width=0.2\textwidth]{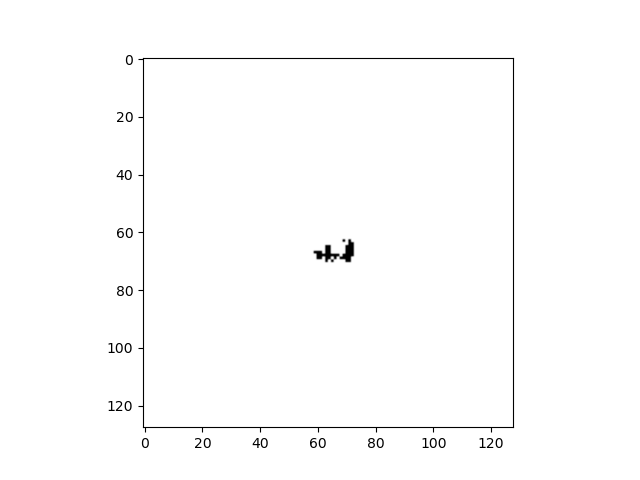}\hfill
    \includegraphics[width=0.2\textwidth]{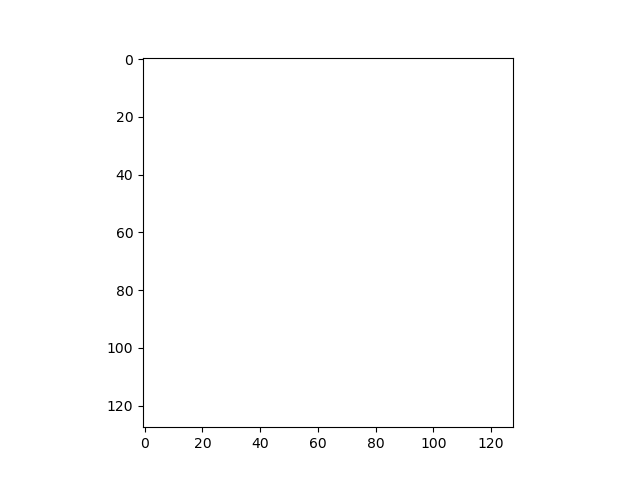}\hfill
    \includegraphics[width=0.2\textwidth]{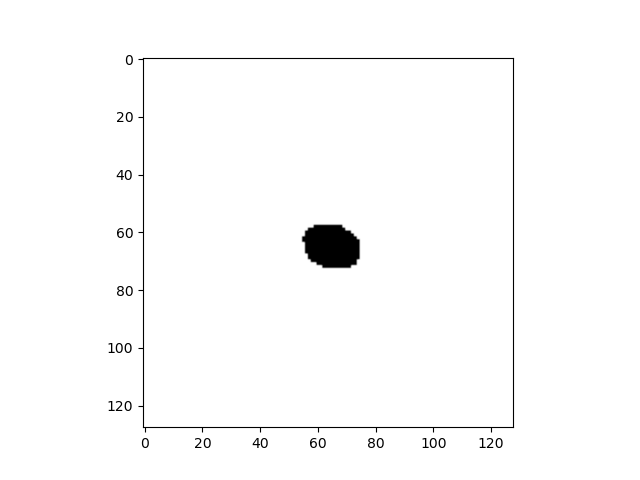}\hfill
    \includegraphics[width=0.2\textwidth]{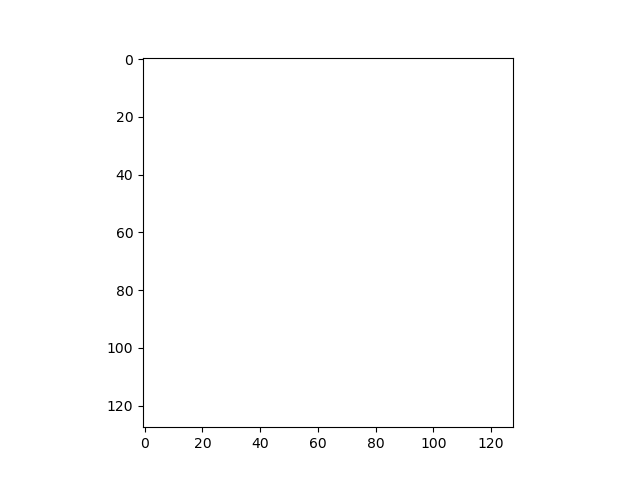}\hfill

    \hrule

    \includegraphics[width=0.2\textwidth]{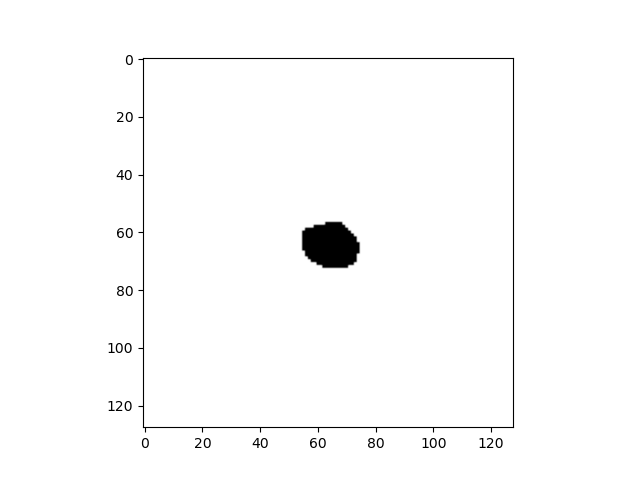}\hfill
    \includegraphics[width=0.2\textwidth]{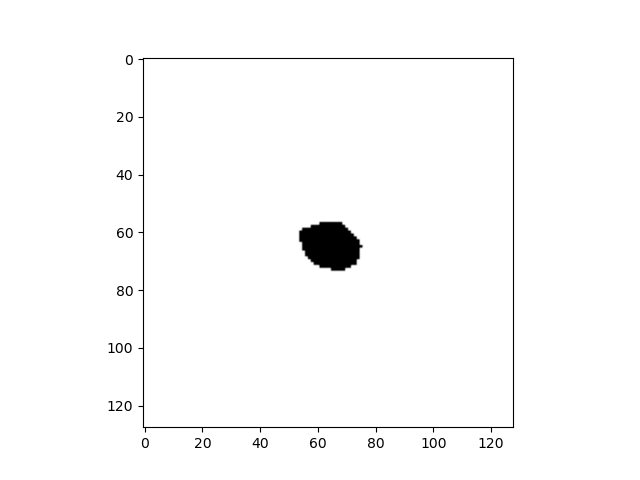}\hfill
    \includegraphics[width=0.2\textwidth]{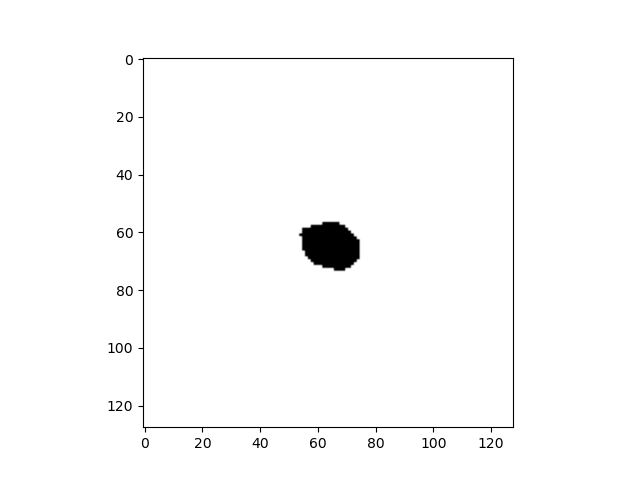}\hfill
    \includegraphics[width=0.2\textwidth]{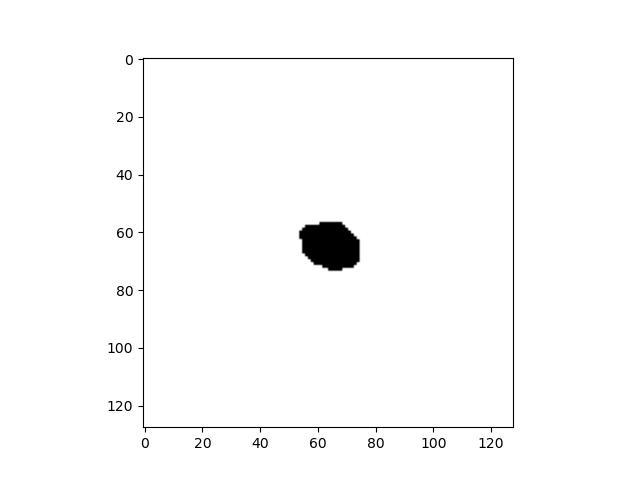}\hfill
    \includegraphics[width=0.2\textwidth]{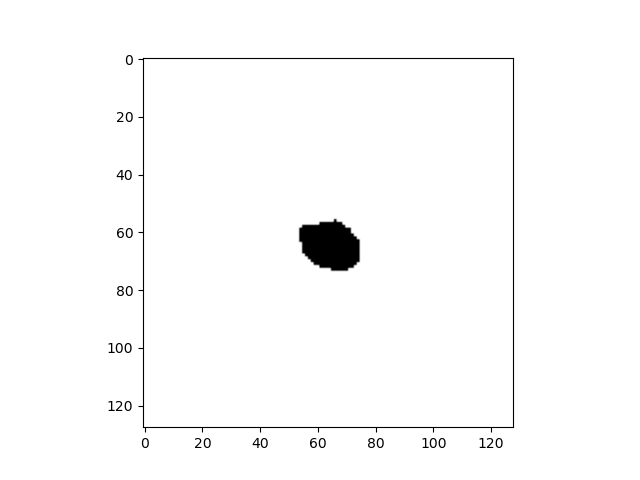}\hfill

    \includegraphics[width=0.2\textwidth]{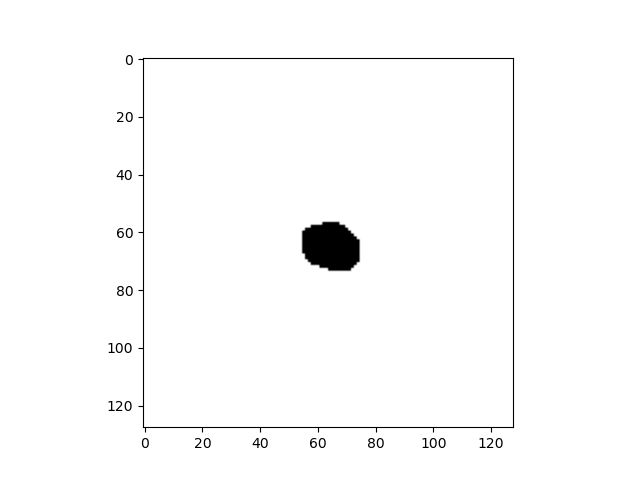}\hfill
    \includegraphics[width=0.2\textwidth]{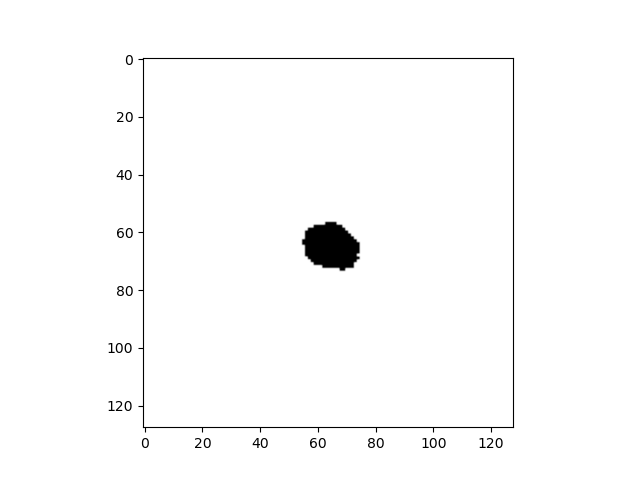}\hfill
    \includegraphics[width=0.2\textwidth]{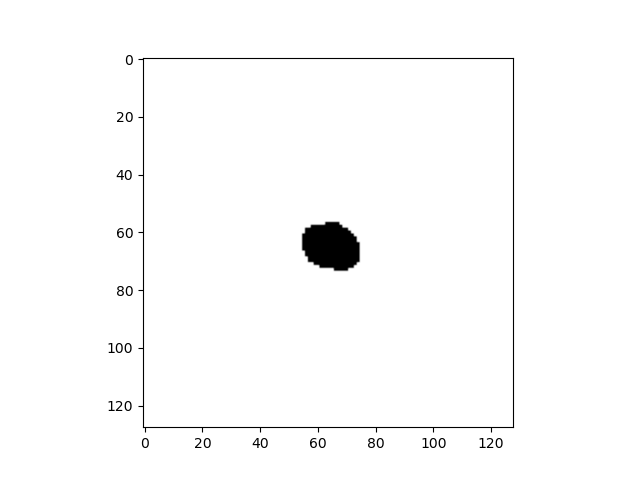}\hfill
    \includegraphics[width=0.2\textwidth]{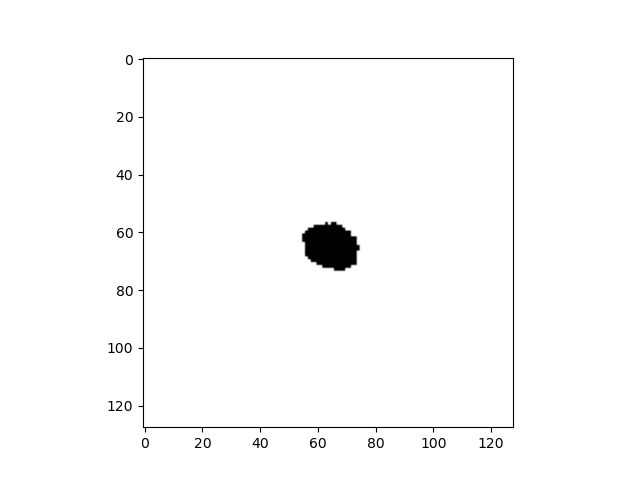}\hfill
    \includegraphics[width=0.2\textwidth]{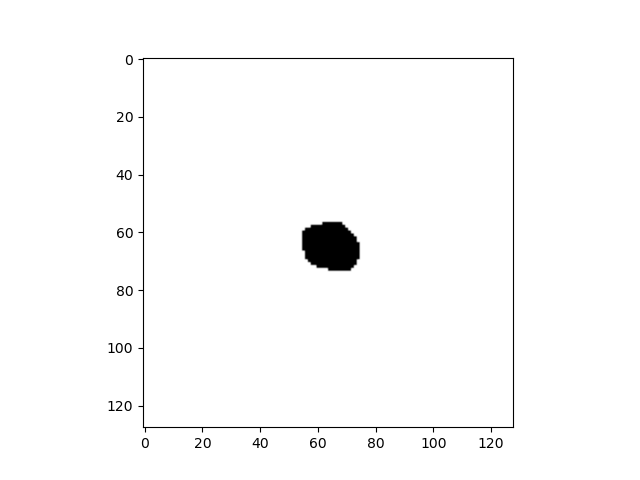}\hfill

    \includegraphics[width=0.2\textwidth]{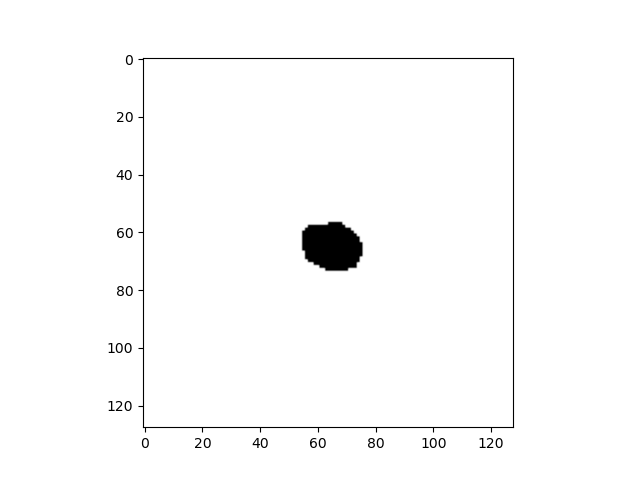}\hfill
    \includegraphics[width=0.2\textwidth]{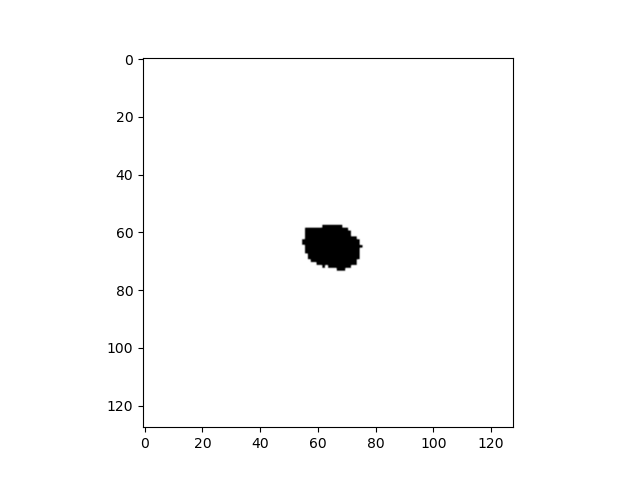}\hfill
    \includegraphics[width=0.2\textwidth]{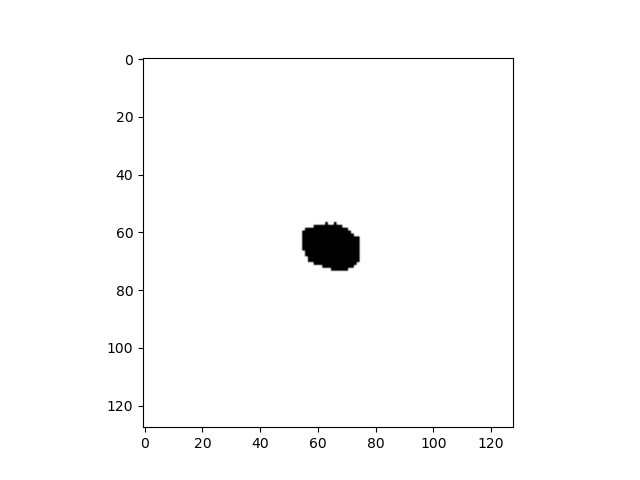}\hfill
    \includegraphics[width=0.2\textwidth]{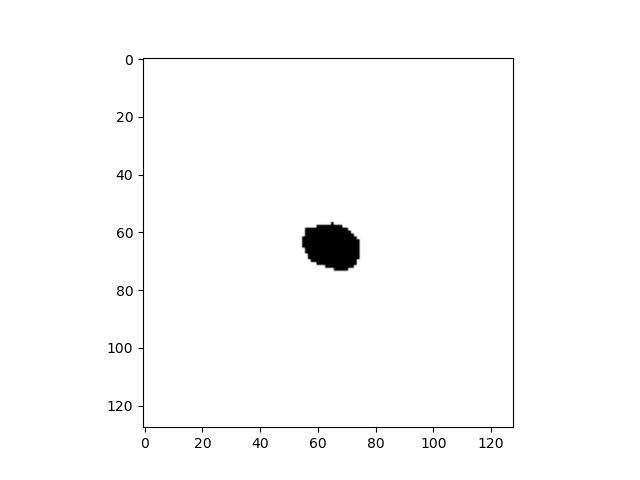}\hfill
    \includegraphics[width=0.2\textwidth]{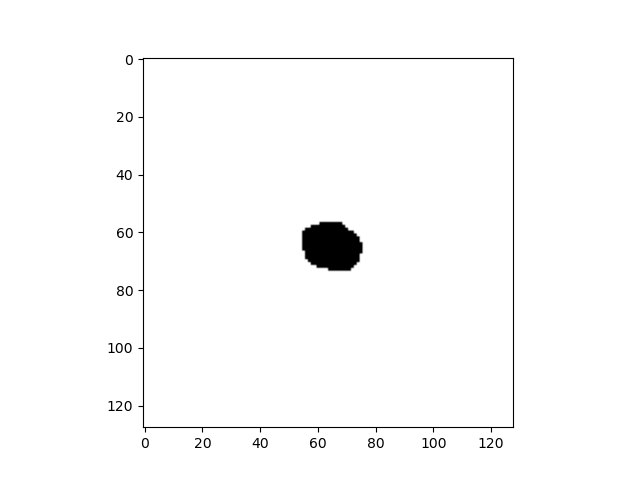}\hfill
    \caption{\textbf{Topleft}: Input image 6 from LIDC data. \textbf{Topright}: Ground truth segmentation. \textbf{2-4 rows}: Segmentation samples from original Probabilistic U-Net. \textbf{5-7 rows}: Segmentation samples from Kendall Shape Probabilistic U-Net. Each row shares the same seed.}
    \label{fig_img6}
\end{figure}

\begin{figure}
    \centering
    \includegraphics[width=0.49\textwidth]{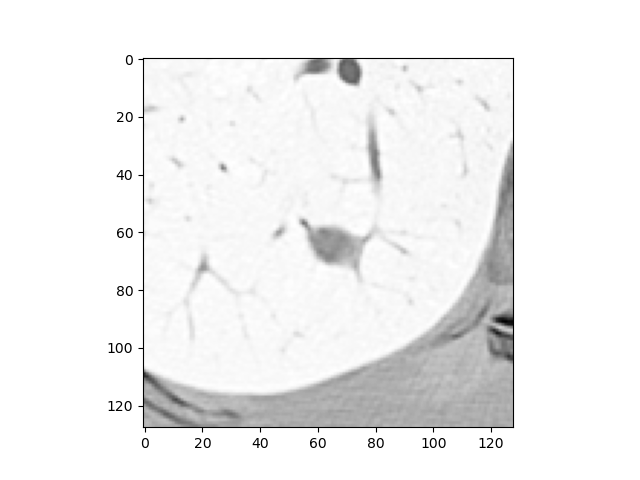}\hfill
    \includegraphics[width=0.49\textwidth]{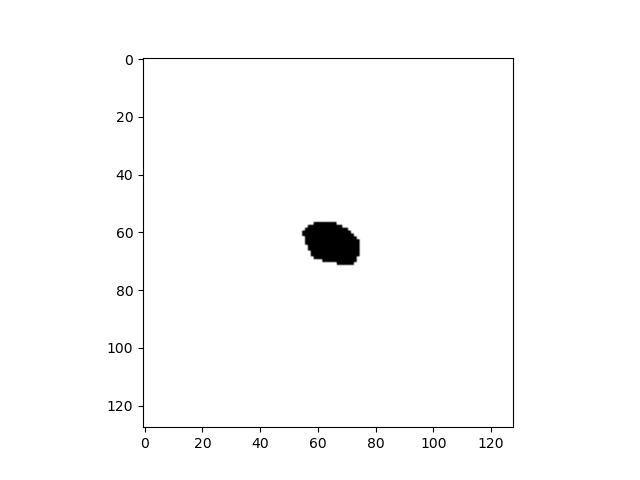}\hfill

    \hrule
    
    \includegraphics[width=0.2\textwidth]{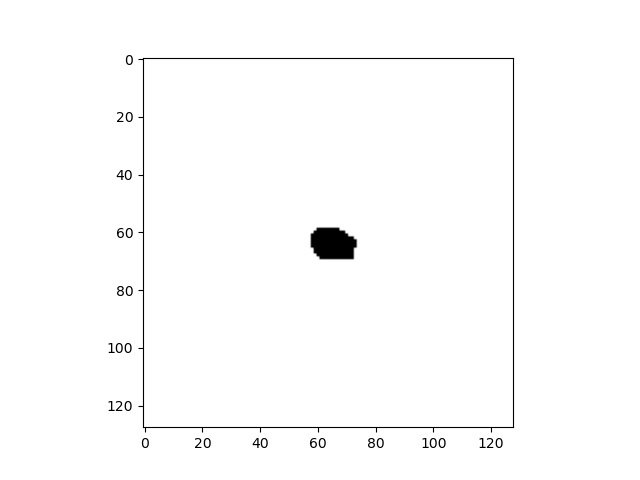}\hfill
    \includegraphics[width=0.2\textwidth]{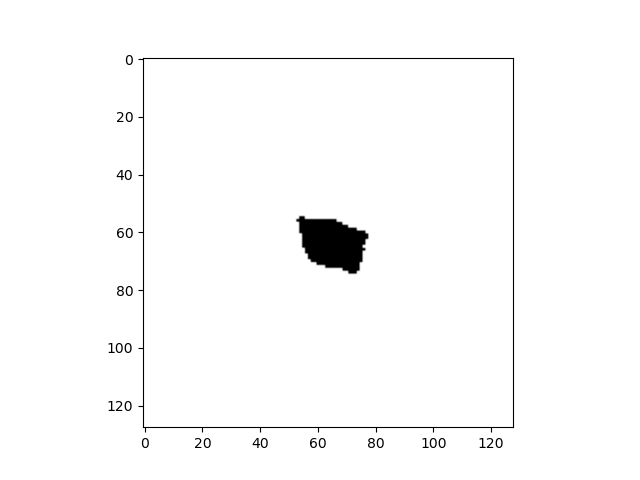}\hfill
    \includegraphics[width=0.2\textwidth]{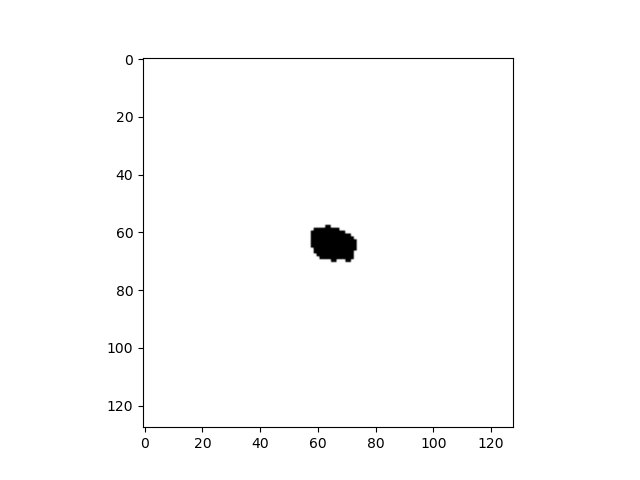}\hfill
    \includegraphics[width=0.2\textwidth]{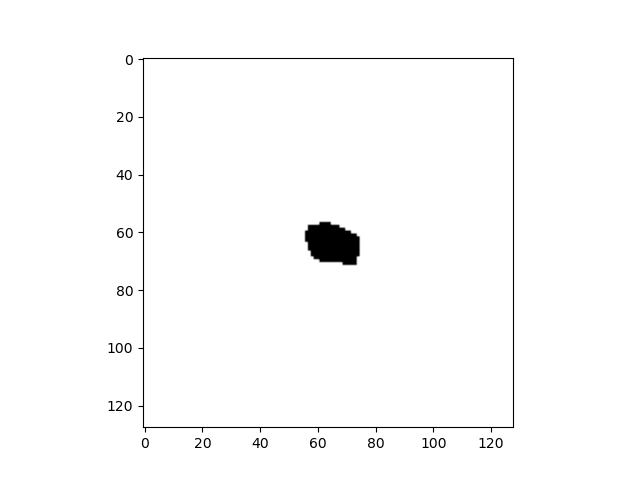}\hfill
    \includegraphics[width=0.2\textwidth]{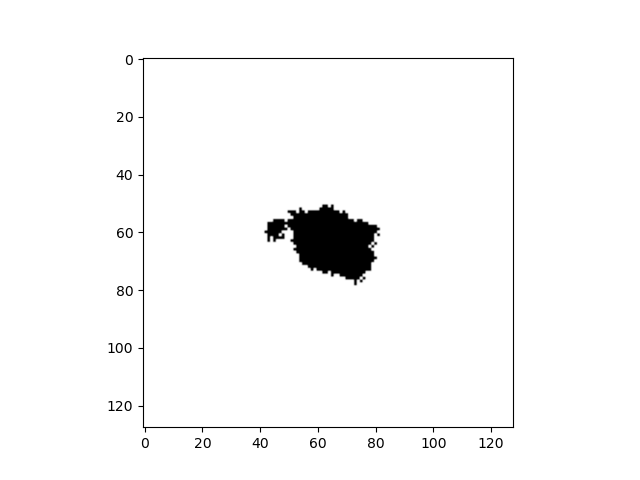}\hfill

    \includegraphics[width=0.2\textwidth]{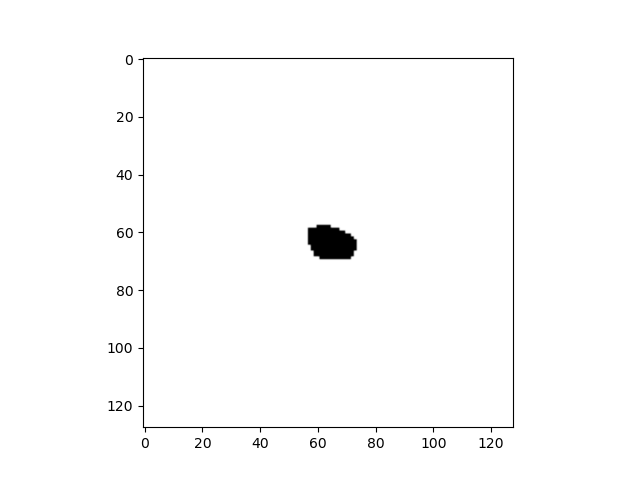}\hfill
    \includegraphics[width=0.2\textwidth]{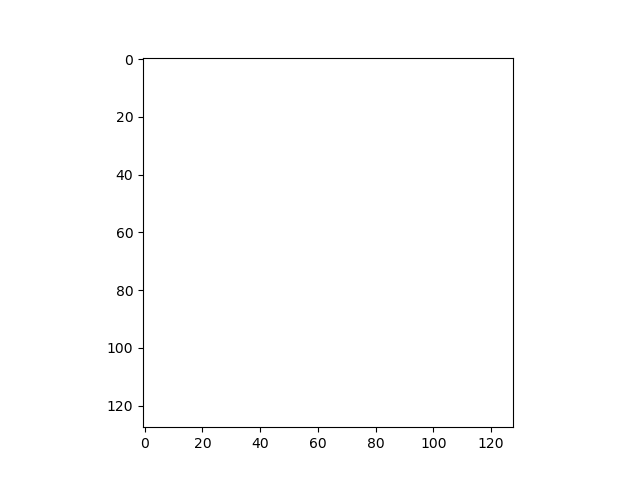}\hfill
    \includegraphics[width=0.2\textwidth]{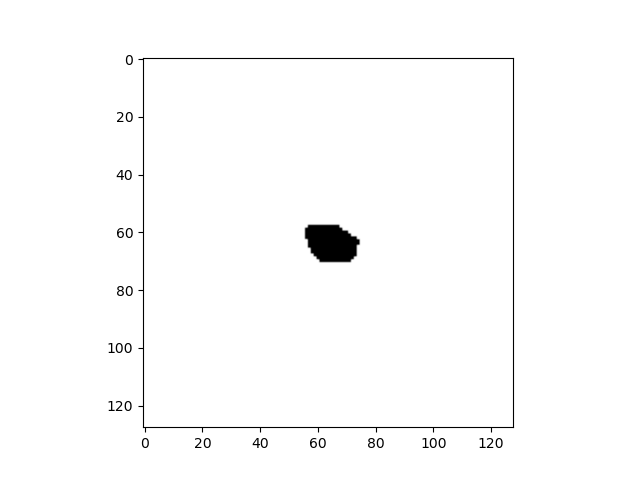}\hfill
    \includegraphics[width=0.2\textwidth]{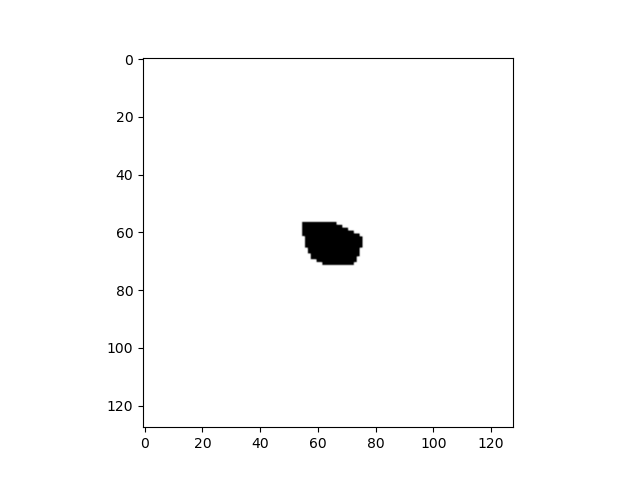}\hfill
    \includegraphics[width=0.2\textwidth]{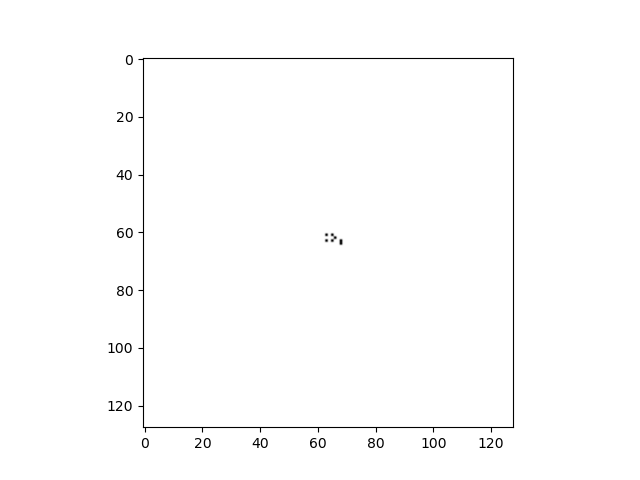}\hfill

    \includegraphics[width=0.2\textwidth]{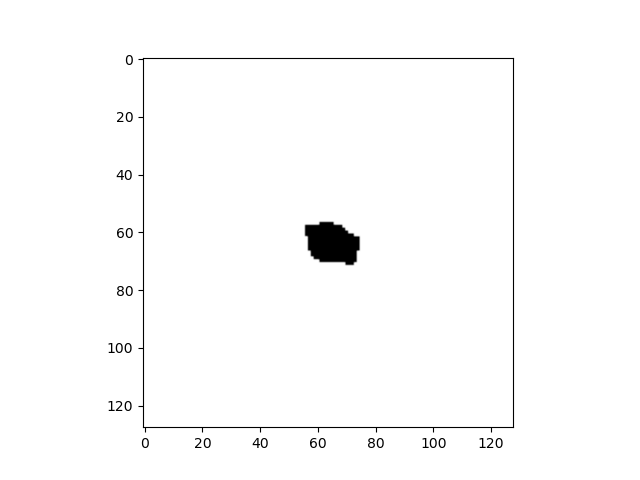}\hfill
    \includegraphics[width=0.2\textwidth]{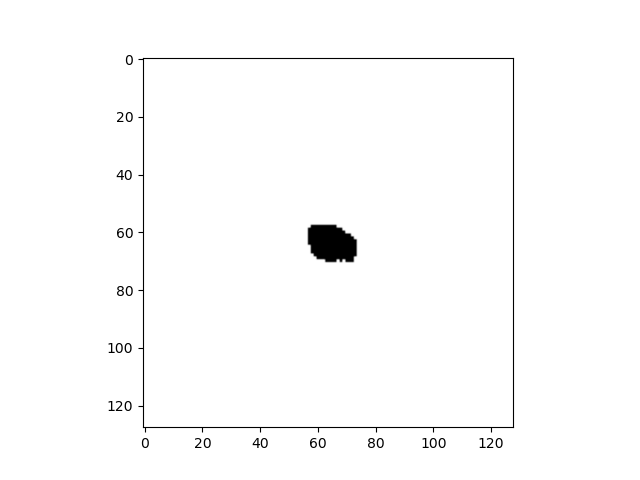}\hfill
    \includegraphics[width=0.2\textwidth]{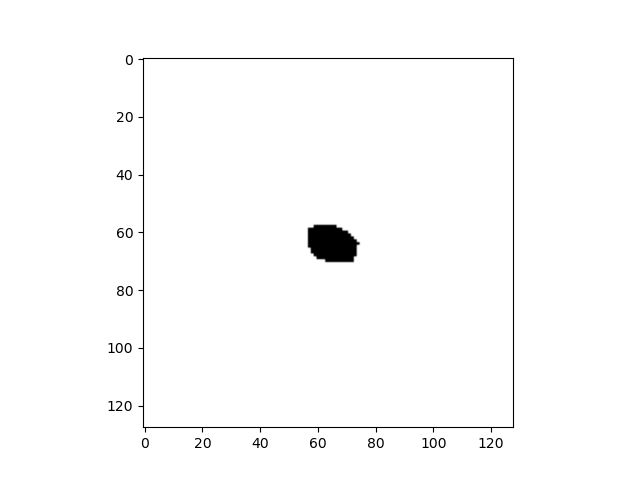}\hfill
    \includegraphics[width=0.2\textwidth]{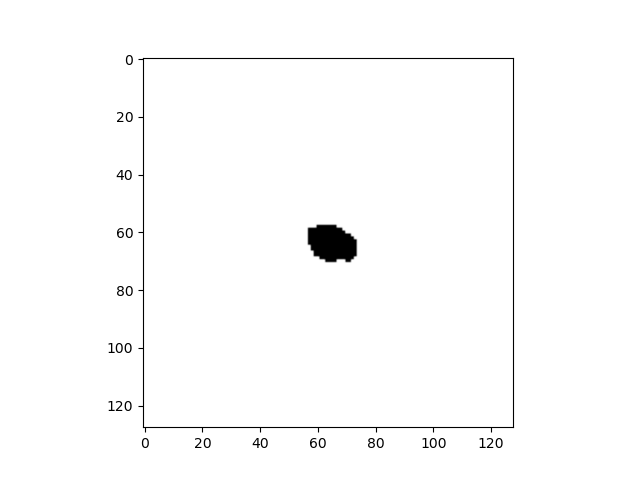}\hfill
    \includegraphics[width=0.2\textwidth]{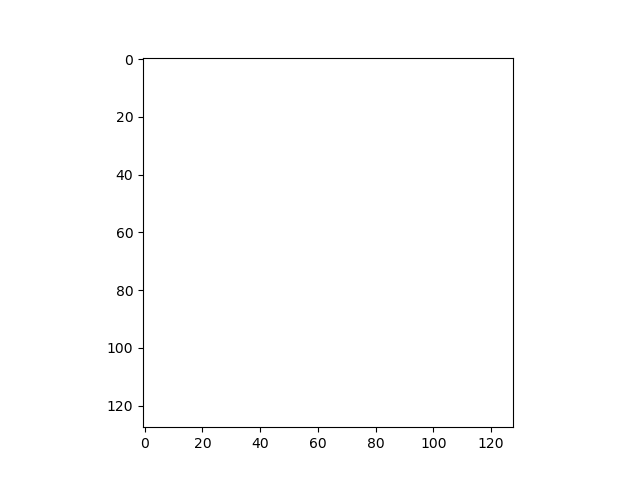}\hfill

    \hrule

    \includegraphics[width=0.2\textwidth]{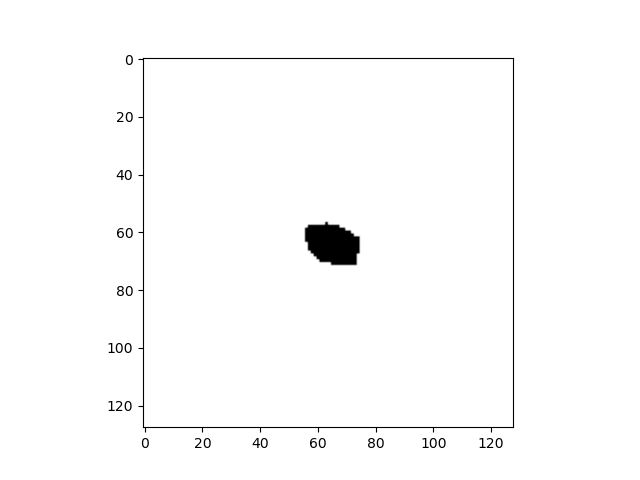}\hfill
    \includegraphics[width=0.2\textwidth]{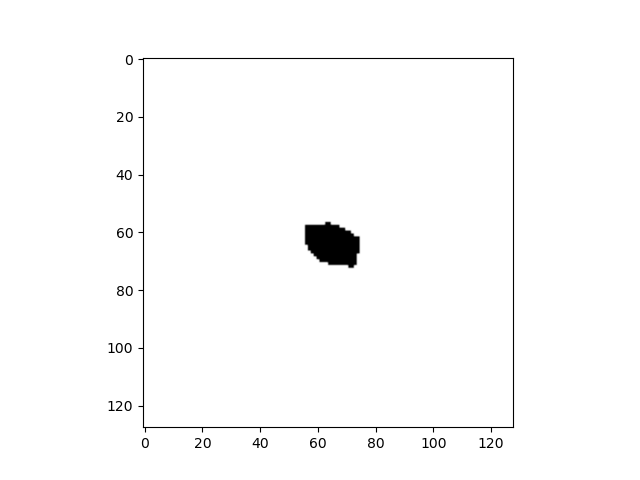}\hfill
    \includegraphics[width=0.2\textwidth]{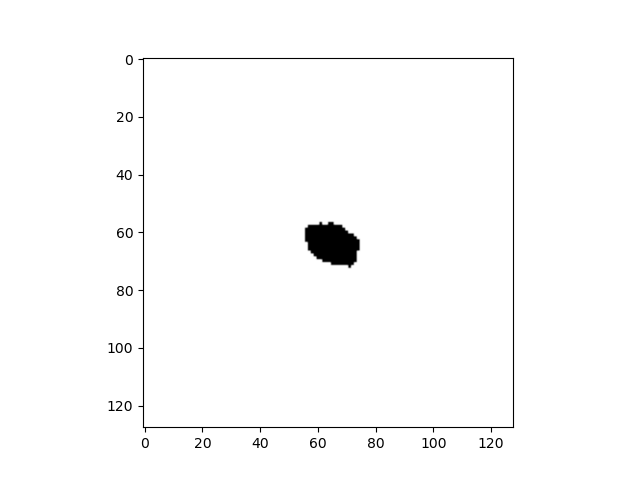}\hfill
    \includegraphics[width=0.2\textwidth]{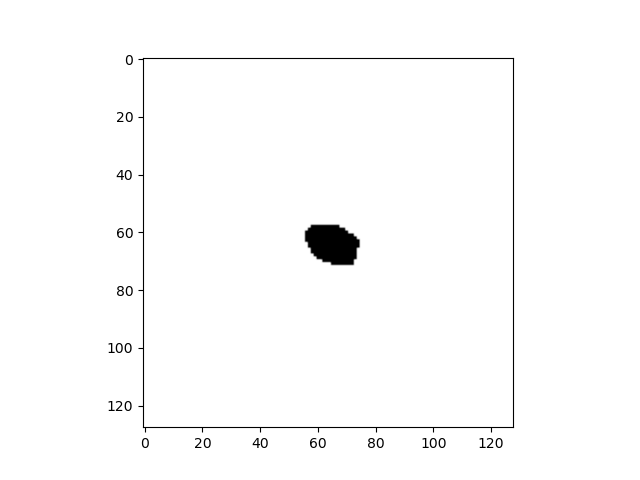}\hfill
    \includegraphics[width=0.2\textwidth]{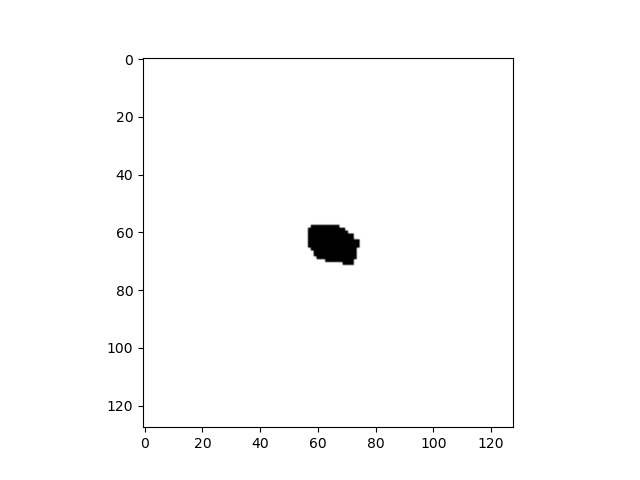}\hfill

    \includegraphics[width=0.2\textwidth]{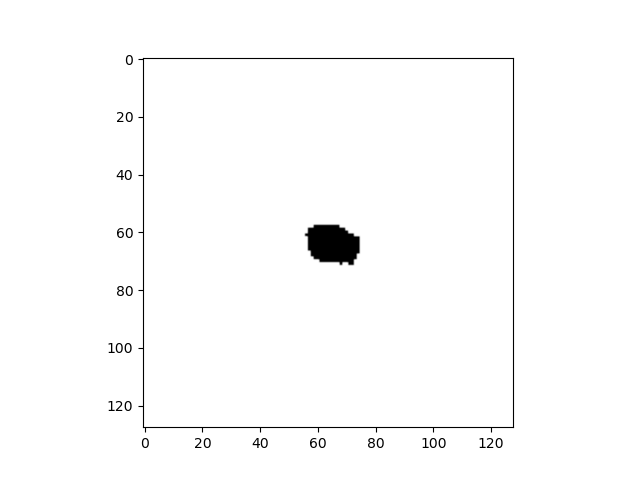}\hfill
    \includegraphics[width=0.2\textwidth]{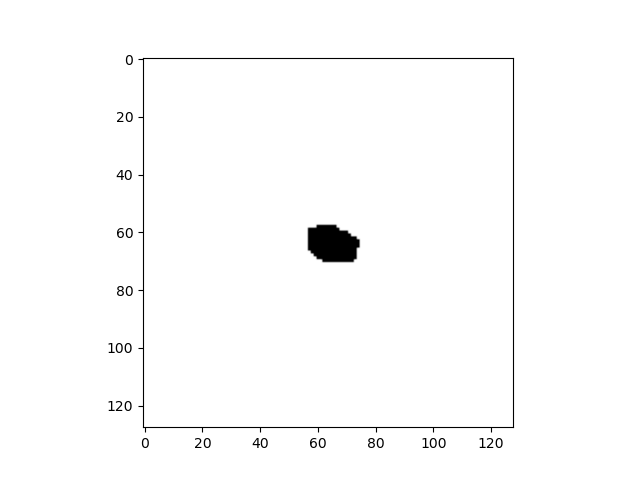}\hfill
    \includegraphics[width=0.2\textwidth]{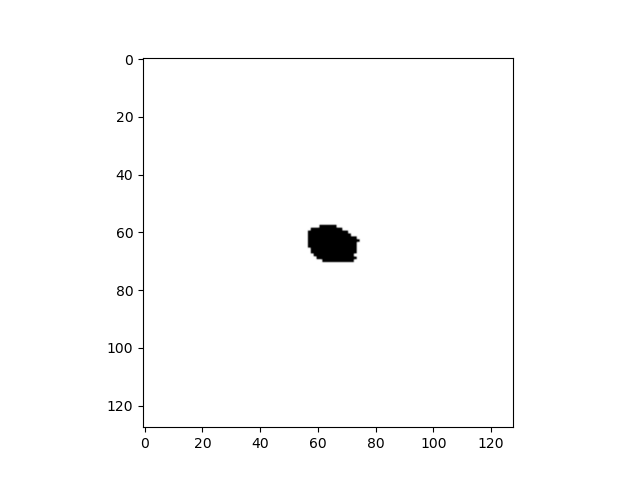}\hfill
    \includegraphics[width=0.2\textwidth]{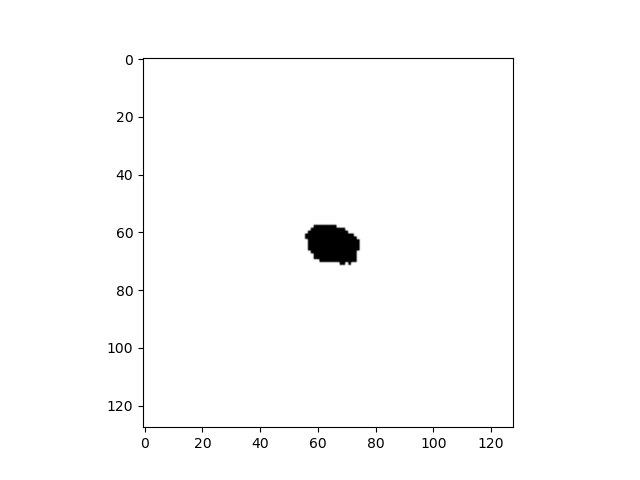}\hfill
    \includegraphics[width=0.2\textwidth]{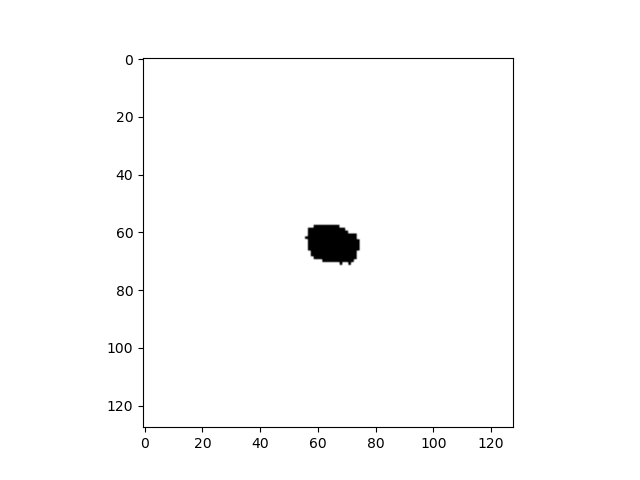}\hfill

    \includegraphics[width=0.2\textwidth]{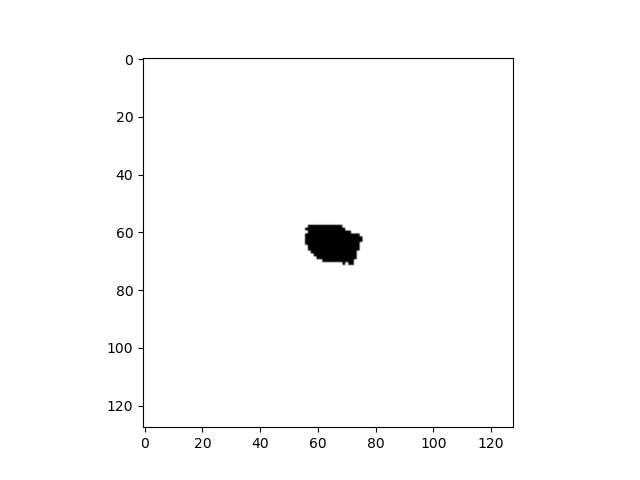}\hfill
    \includegraphics[width=0.2\textwidth]{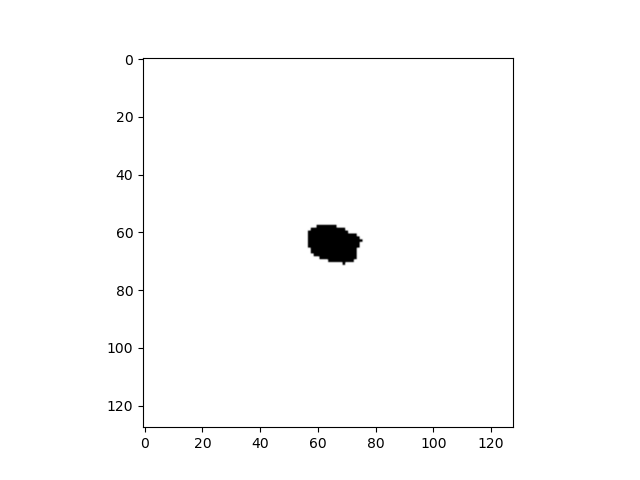}\hfill
    \includegraphics[width=0.2\textwidth]{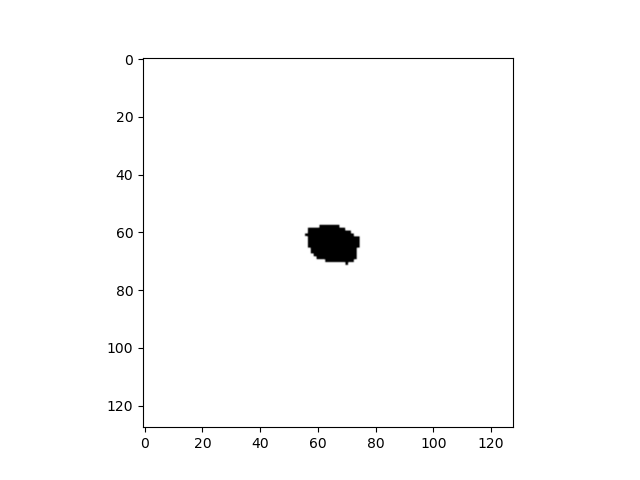}\hfill
    \includegraphics[width=0.2\textwidth]{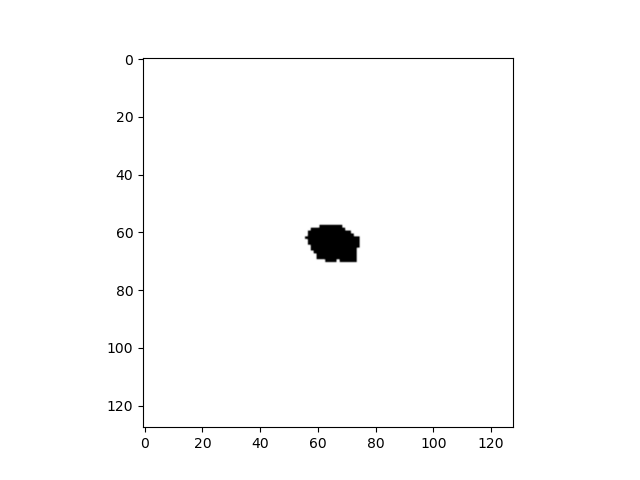}\hfill
    \includegraphics[width=0.2\textwidth]{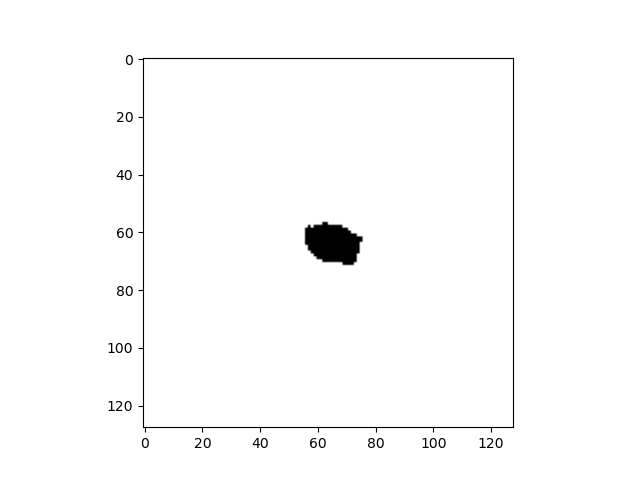}\hfill
    \caption{\textbf{Topleft}: Input image 7 from LIDC data. \textbf{Topright}: Ground truth segmentation. \textbf{2-4 rows}: Segmentation samples from original Probabilistic U-Net. \textbf{5-7 rows}: Segmentation samples from Kendall Shape Probabilistic U-Net. Each row shares the same seed.}
    \label{fig_img7}
\end{figure}

\begin{figure}
    \centering
    \includegraphics[width=0.49\textwidth]{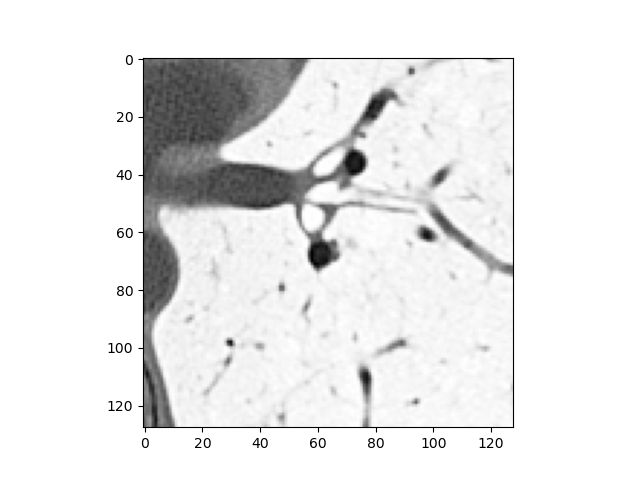}\hfill
    \includegraphics[width=0.49\textwidth]{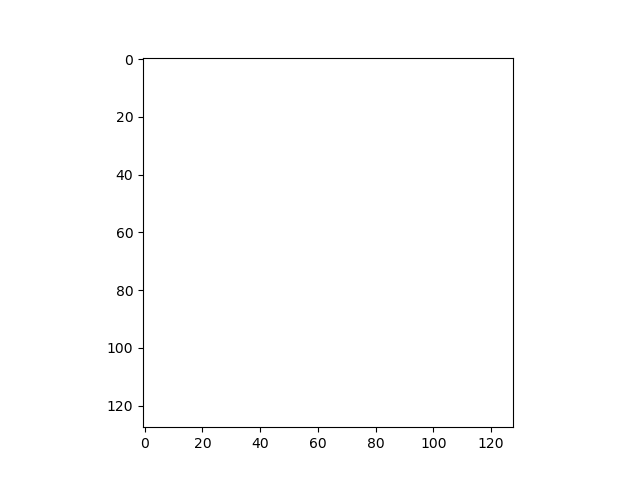}\hfill

    \hrule
    
    \includegraphics[width=0.2\textwidth]{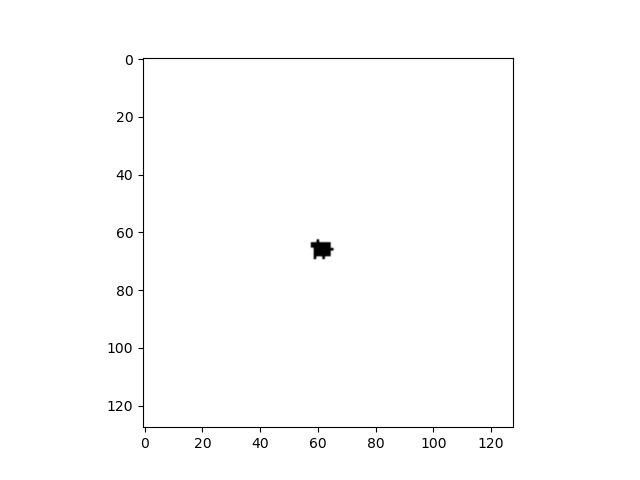}\hfill
    \includegraphics[width=0.2\textwidth]{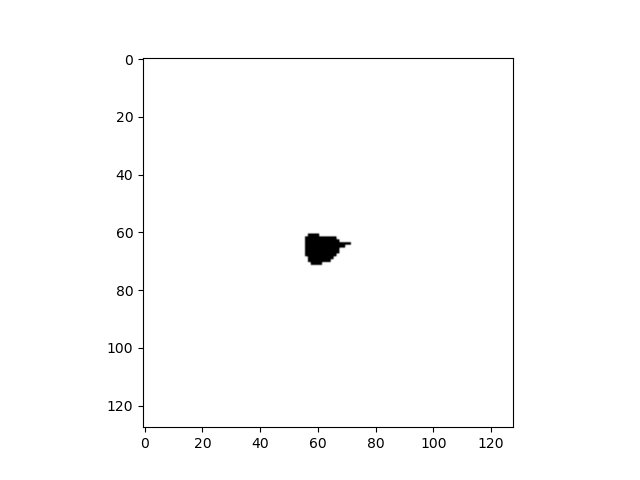}\hfill
    \includegraphics[width=0.2\textwidth]{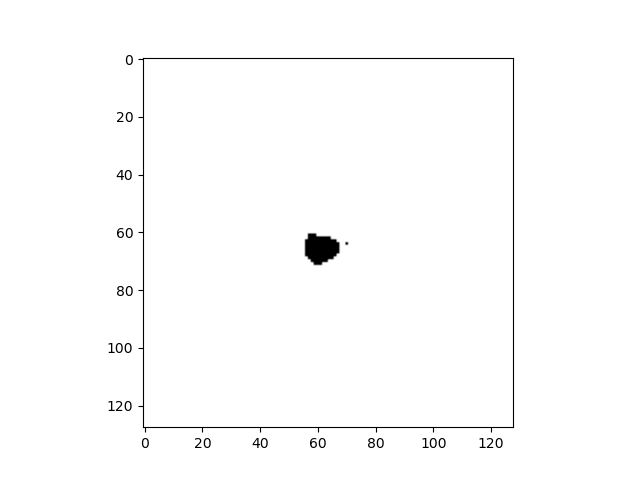}\hfill
    \includegraphics[width=0.2\textwidth]{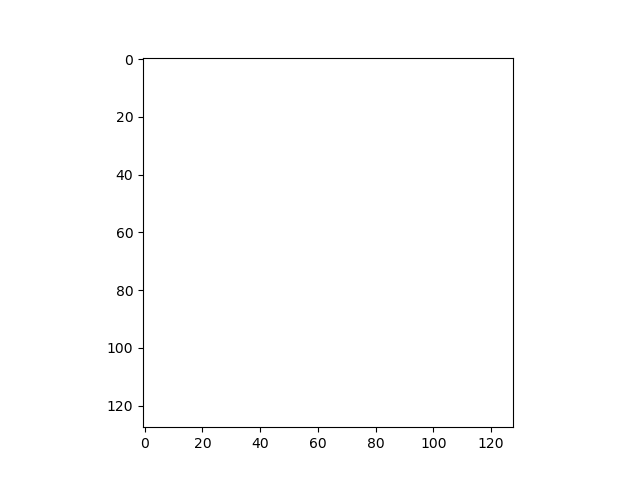}\hfill
    \includegraphics[width=0.2\textwidth]{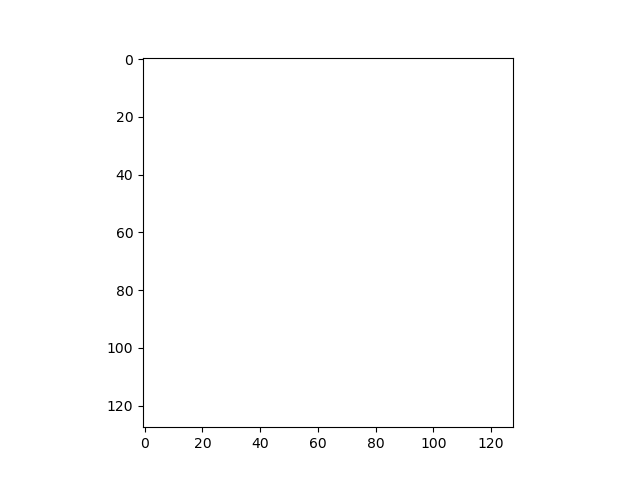}\hfill

    \includegraphics[width=0.2\textwidth]{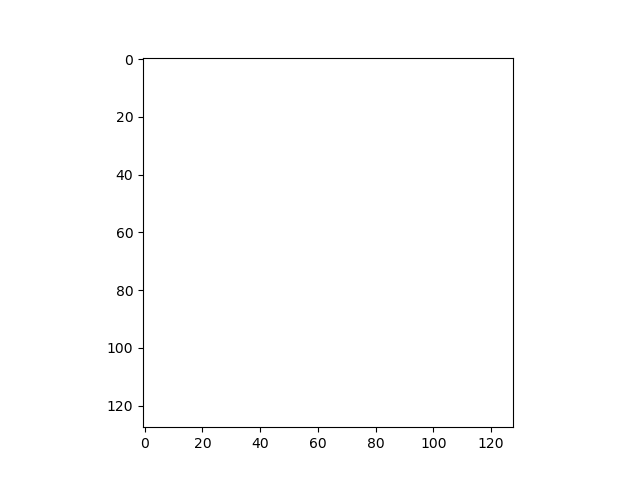}\hfill
    \includegraphics[width=0.2\textwidth]{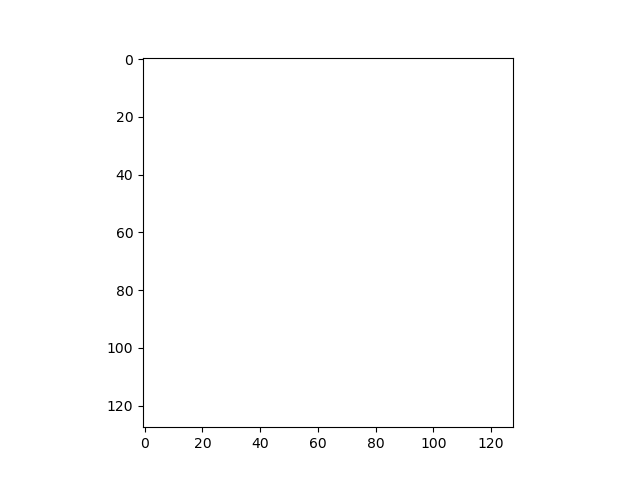}\hfill
    \includegraphics[width=0.2\textwidth]{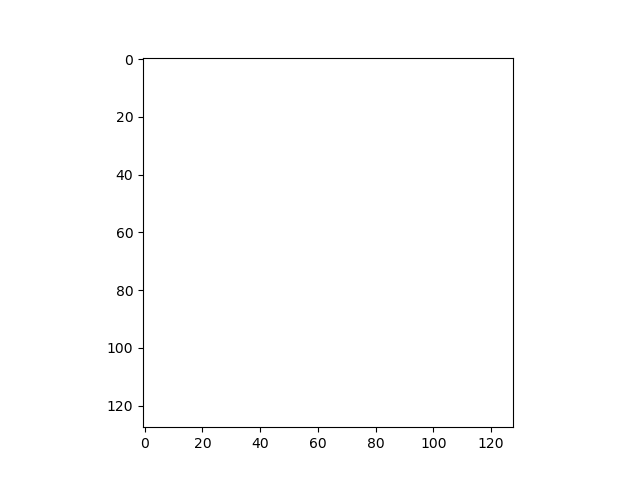}\hfill
    \includegraphics[width=0.2\textwidth]{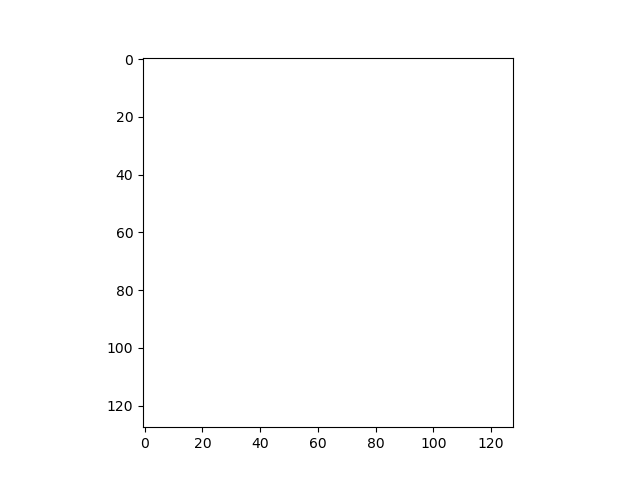}\hfill
    \includegraphics[width=0.2\textwidth]{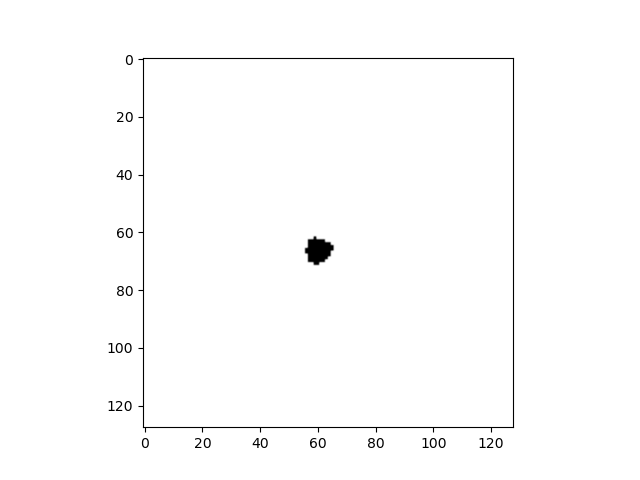}\hfill

    \includegraphics[width=0.2\textwidth]{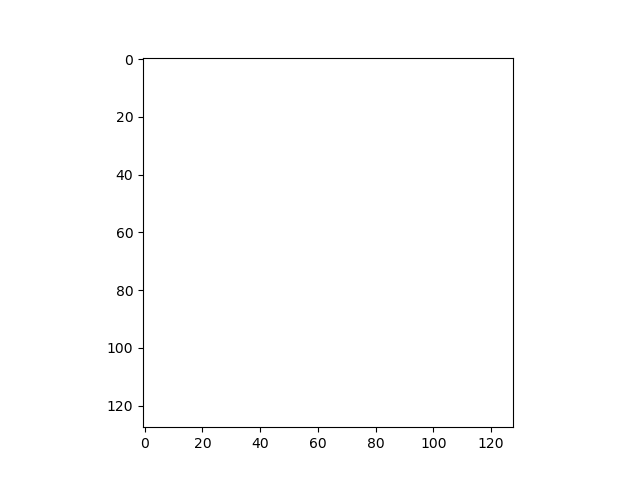}\hfill
    \includegraphics[width=0.2\textwidth]{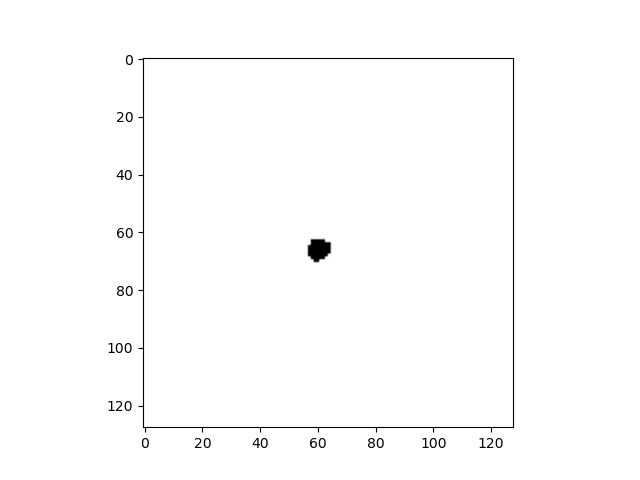}\hfill
    \includegraphics[width=0.2\textwidth]{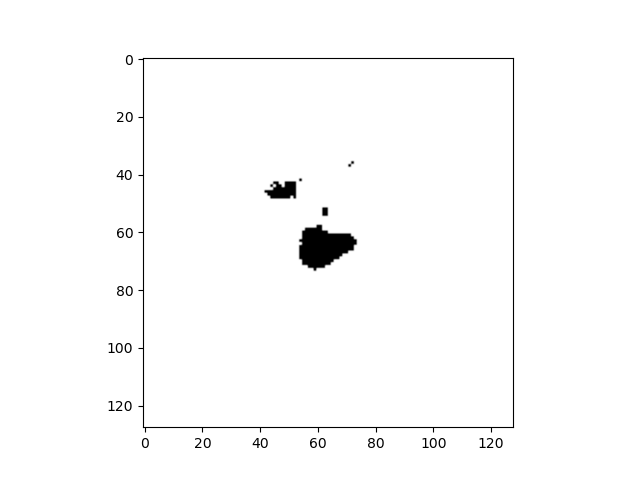}\hfill
    \includegraphics[width=0.2\textwidth]{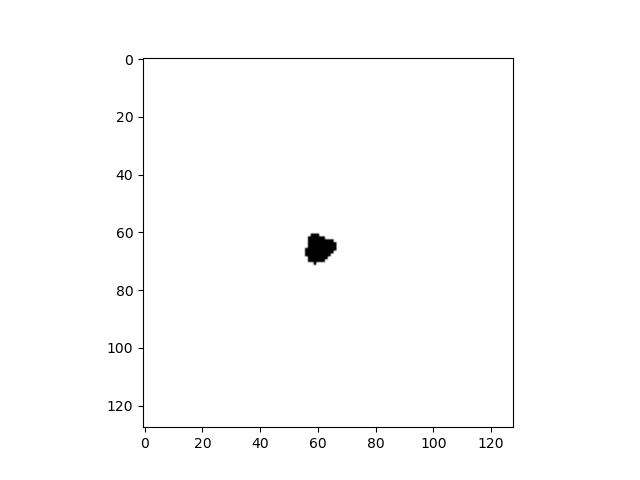}\hfill
    \includegraphics[width=0.2\textwidth]{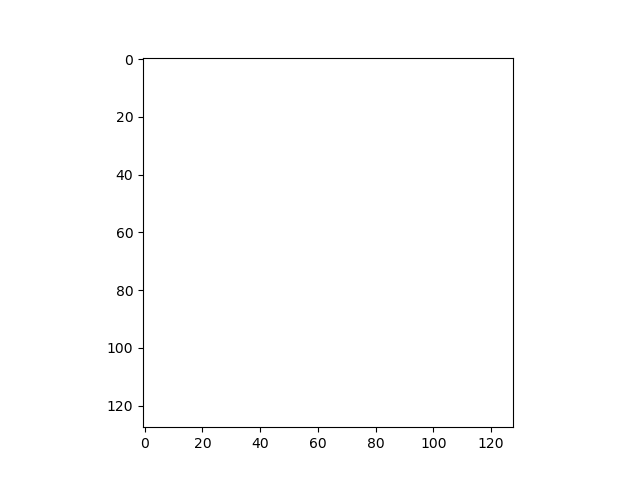}\hfill

    \hrule

    \includegraphics[width=0.2\textwidth]{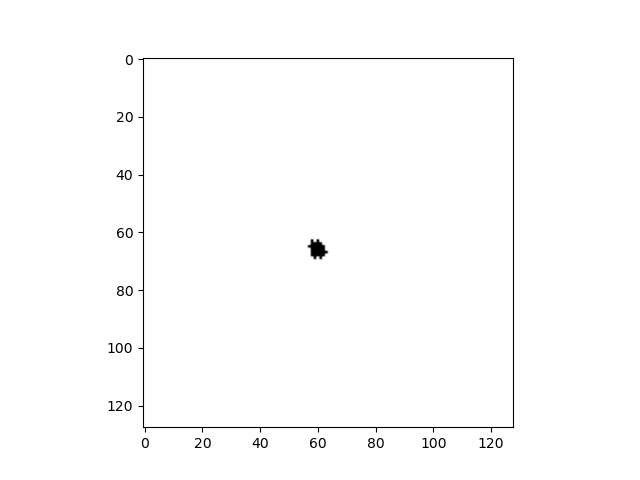}\hfill
    \includegraphics[width=0.2\textwidth]{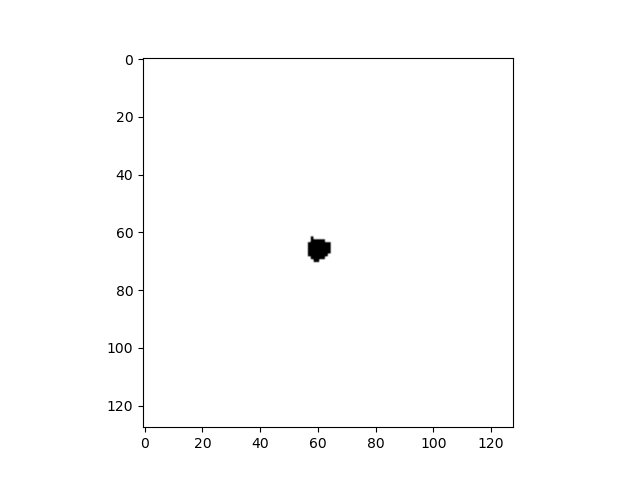}\hfill
    \includegraphics[width=0.2\textwidth]{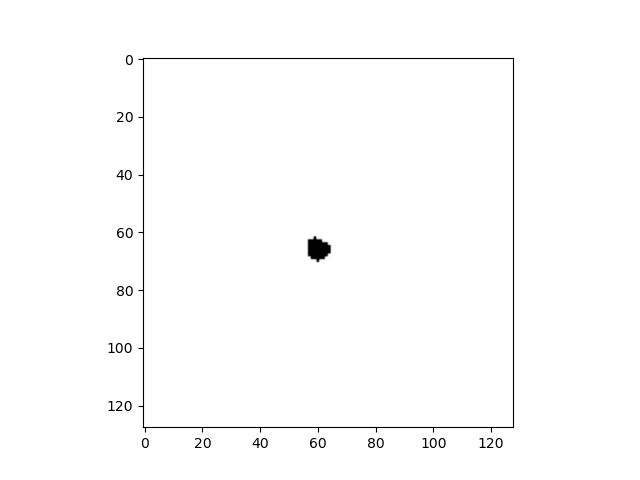}\hfill
    \includegraphics[width=0.2\textwidth]{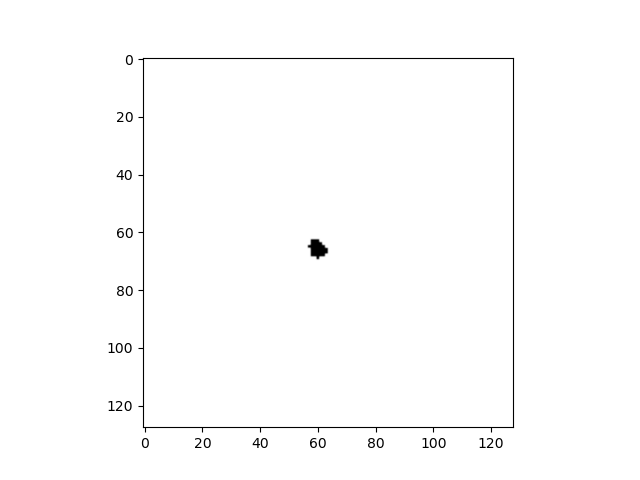}\hfill
    \includegraphics[width=0.2\textwidth]{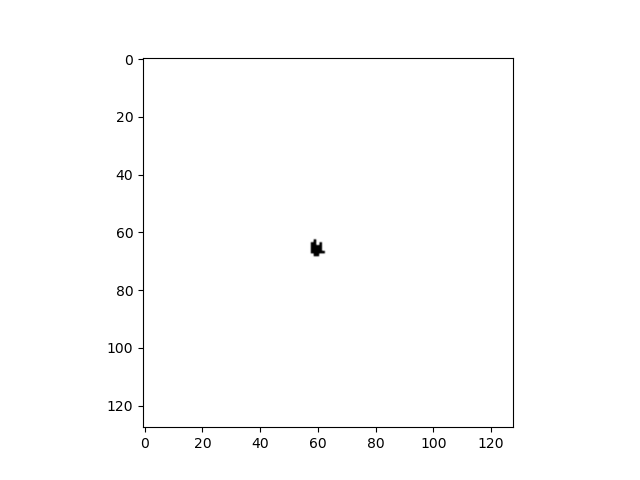}\hfill

    \includegraphics[width=0.2\textwidth]{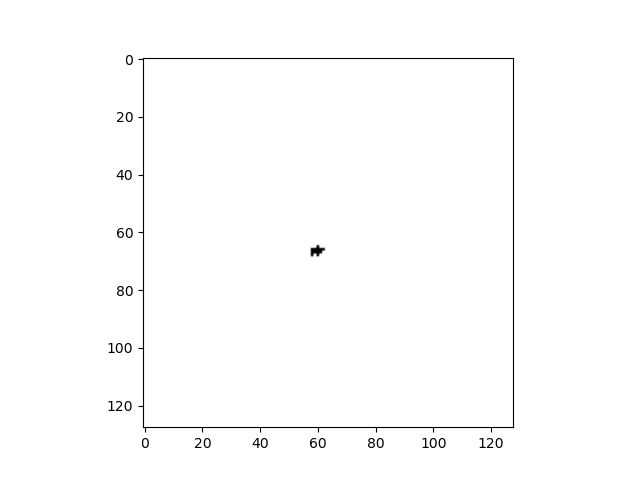}\hfill
    \includegraphics[width=0.2\textwidth]{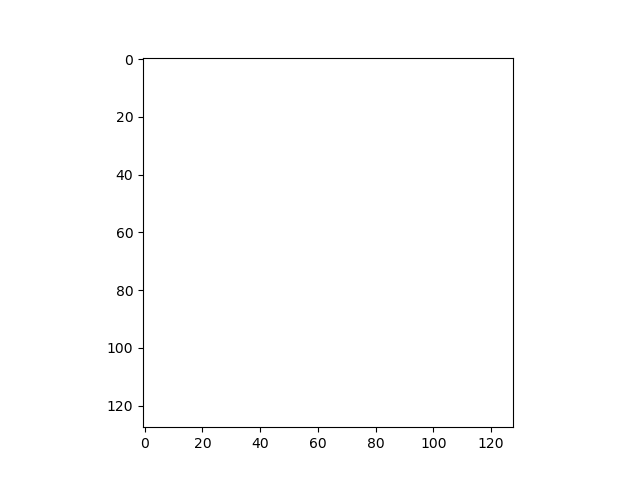}\hfill
    \includegraphics[width=0.2\textwidth]{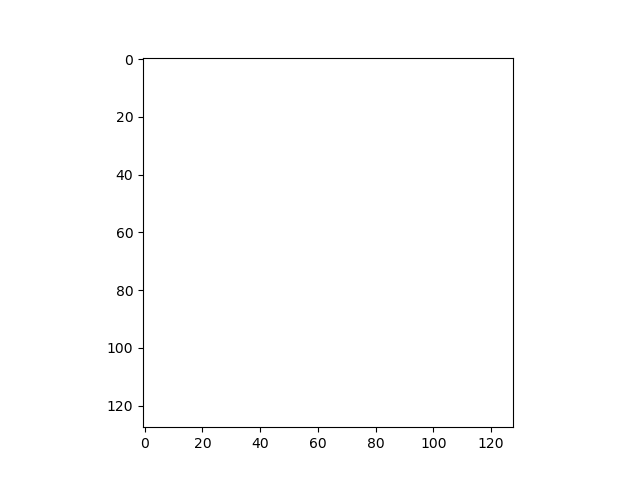}\hfill
    \includegraphics[width=0.2\textwidth]{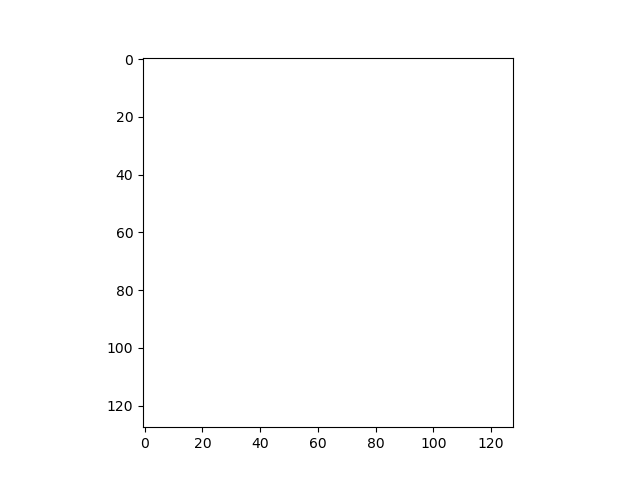}\hfill
    \includegraphics[width=0.2\textwidth]{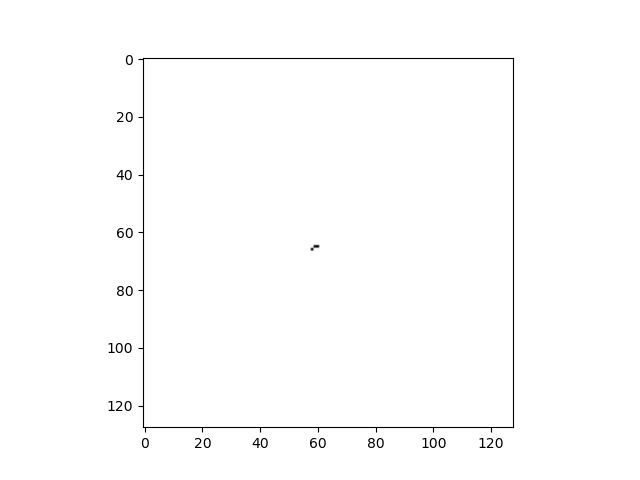}\hfill

    \includegraphics[width=0.2\textwidth]{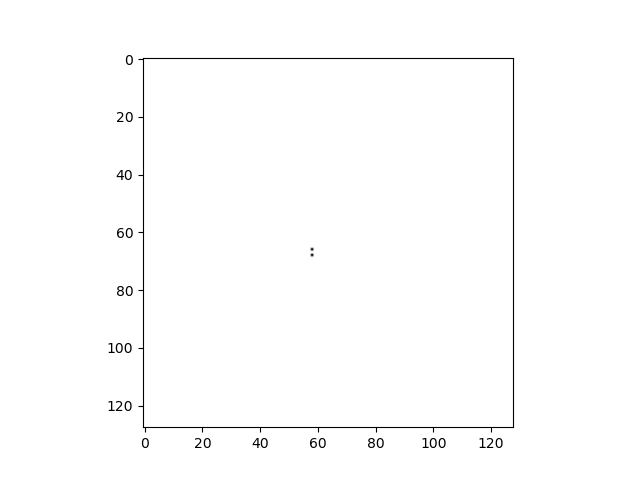}\hfill
    \includegraphics[width=0.2\textwidth]{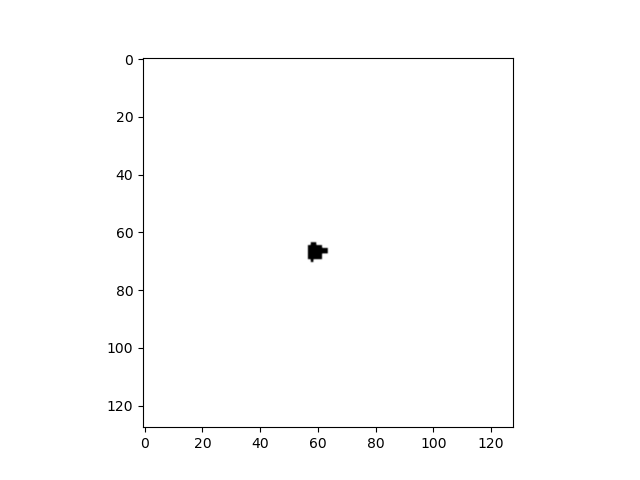}\hfill
    \includegraphics[width=0.2\textwidth]{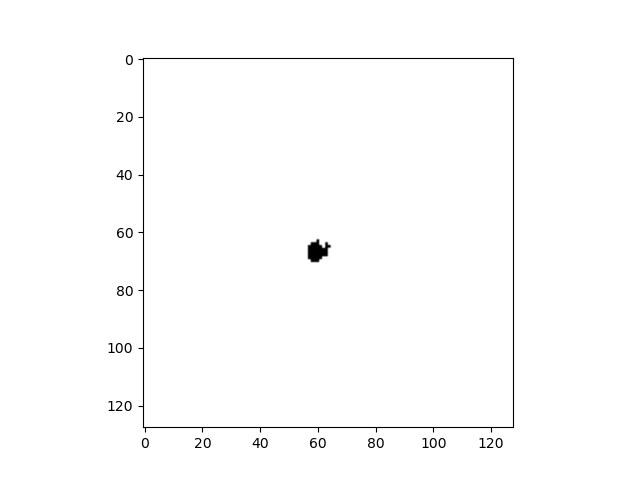}\hfill
    \includegraphics[width=0.2\textwidth]{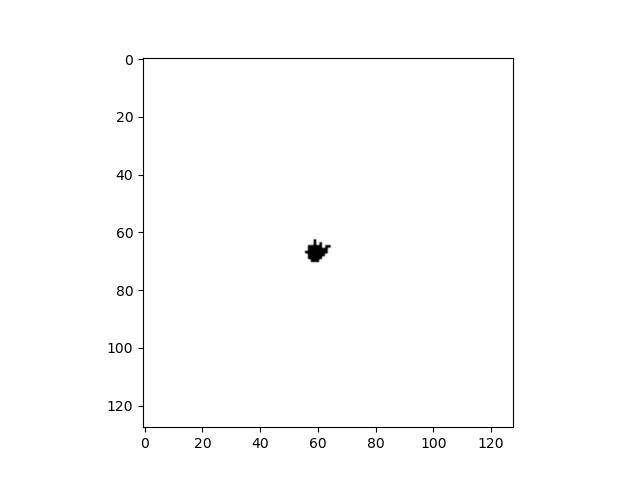}\hfill
    \includegraphics[width=0.2\textwidth]{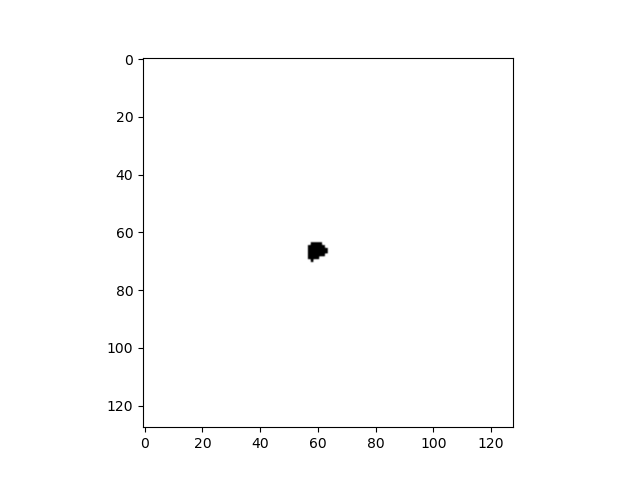}\hfill
    \caption{\textbf{Topleft}: Input image 8 from LIDC data. \textbf{Topright}: Ground truth segmentation. \textbf{2-4 rows}: Segmentation samples from original Probabilistic U-Net. \textbf{5-7 rows}: Segmentation samples from Kendall Shape Probabilistic U-Net. Each row shares the same seed.}
    \label{fig_img8}
\end{figure}

\begin{figure}
    \centering
    \includegraphics[width=0.49\textwidth]{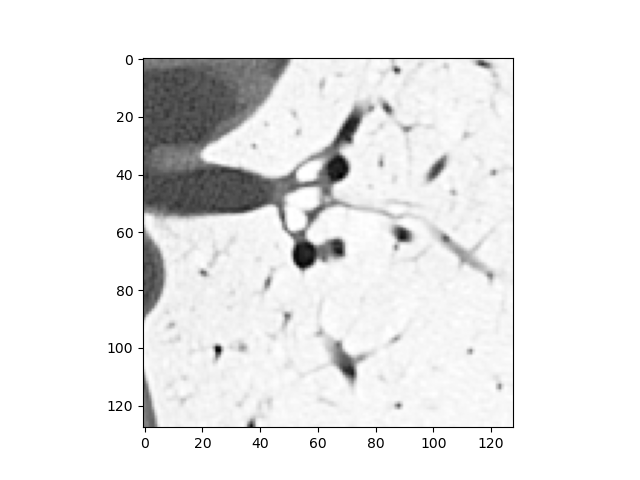}\hfill
    \includegraphics[width=0.49\textwidth]{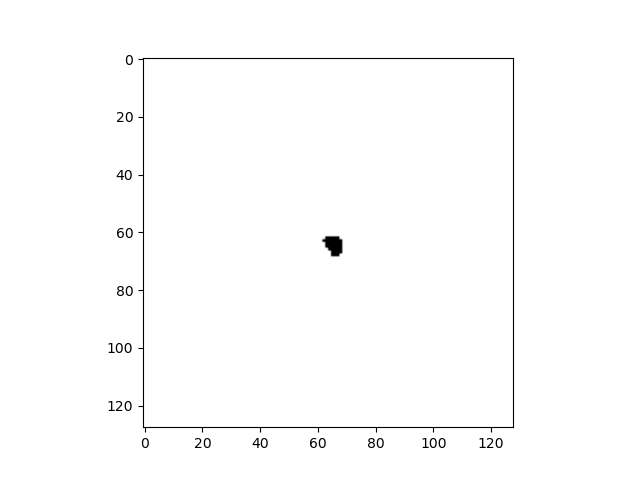}\hfill

    \hrule
    
    \includegraphics[width=0.2\textwidth]{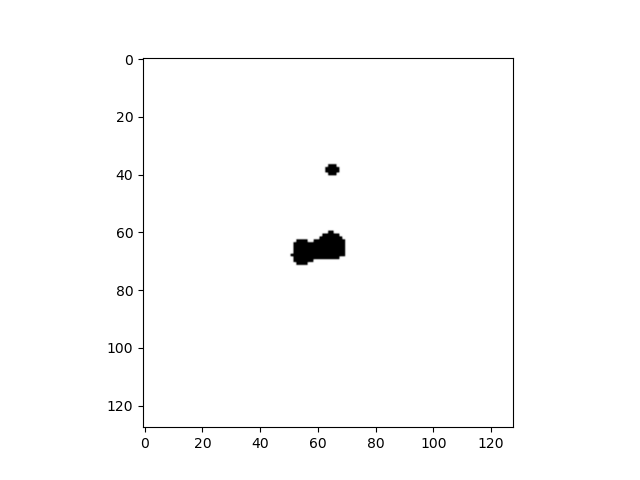}\hfill
    \includegraphics[width=0.2\textwidth]{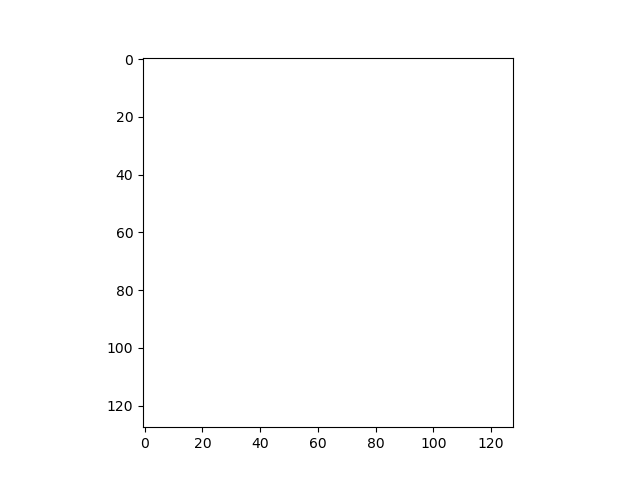}\hfill
    \includegraphics[width=0.2\textwidth]{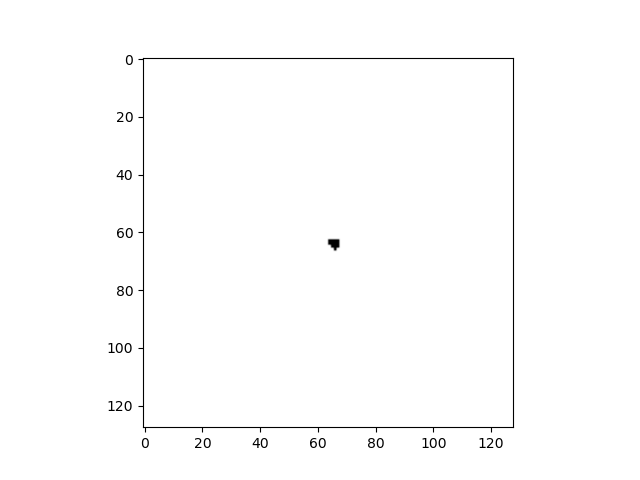}\hfill
    \includegraphics[width=0.2\textwidth]{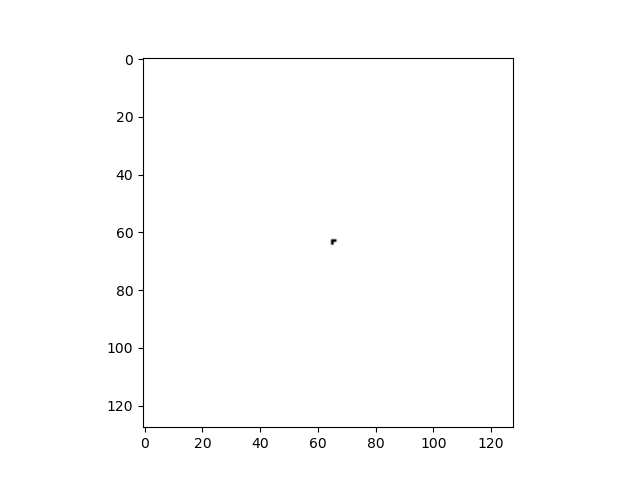}\hfill
    \includegraphics[width=0.2\textwidth]{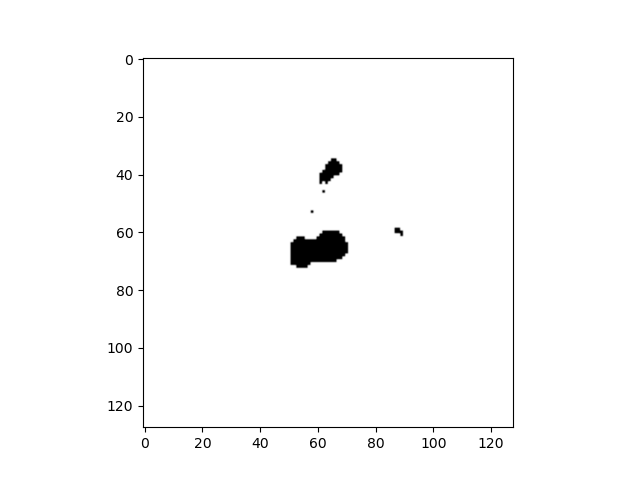}\hfill

    \includegraphics[width=0.2\textwidth]{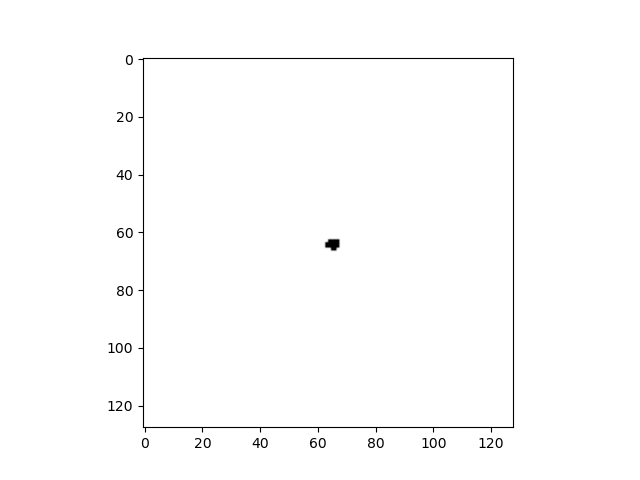}\hfill
    \includegraphics[width=0.2\textwidth]{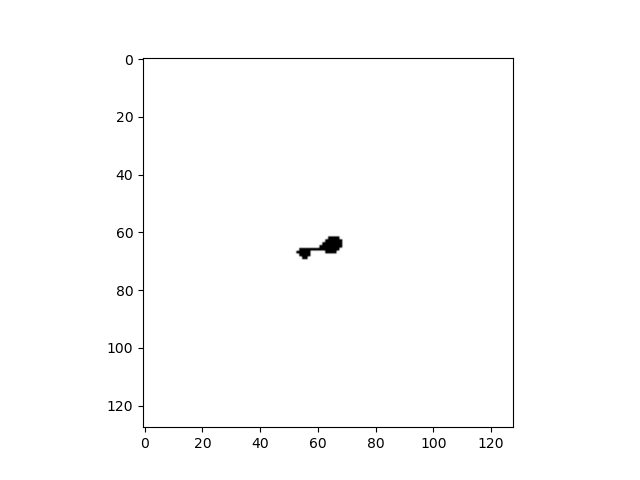}\hfill
    \includegraphics[width=0.2\textwidth]{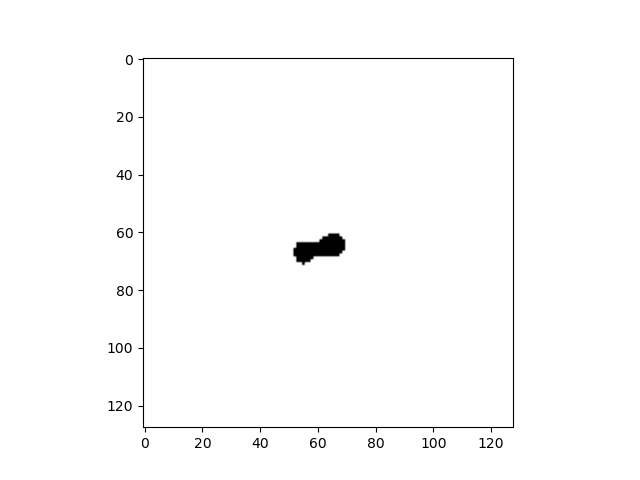}\hfill
    \includegraphics[width=0.2\textwidth]{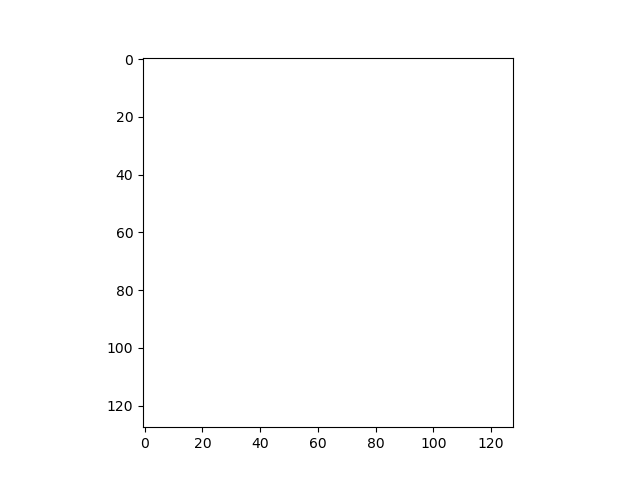}\hfill
    \includegraphics[width=0.2\textwidth]{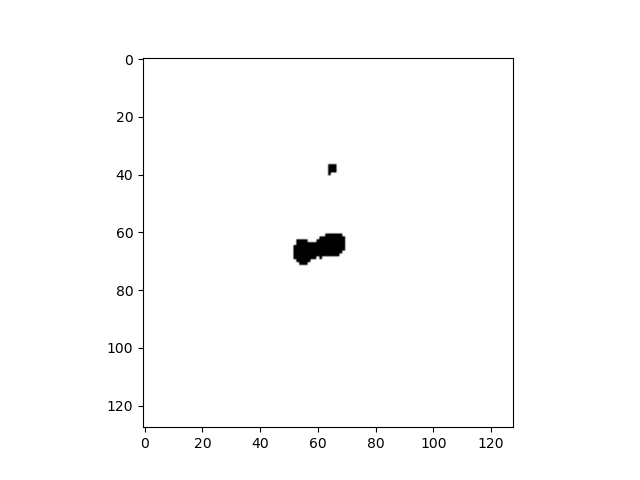}\hfill

    \includegraphics[width=0.2\textwidth]{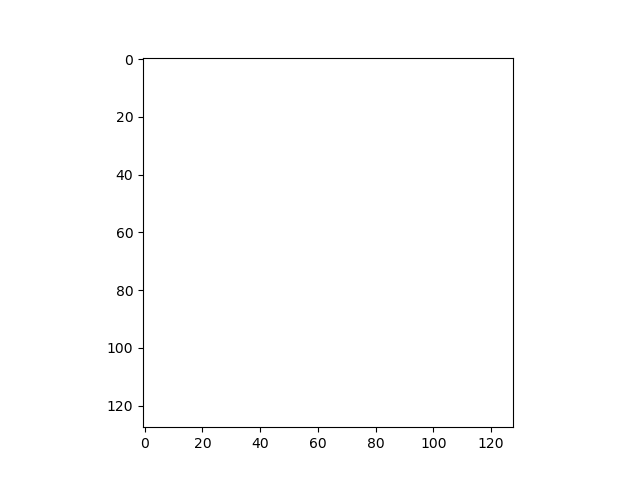}\hfill
    \includegraphics[width=0.2\textwidth]{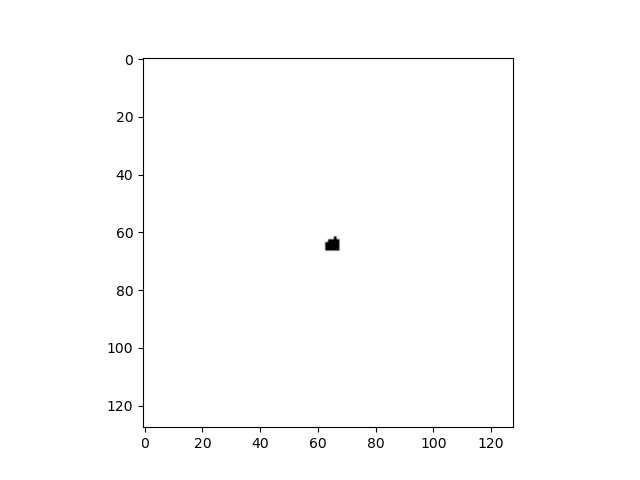}\hfill
    \includegraphics[width=0.2\textwidth]{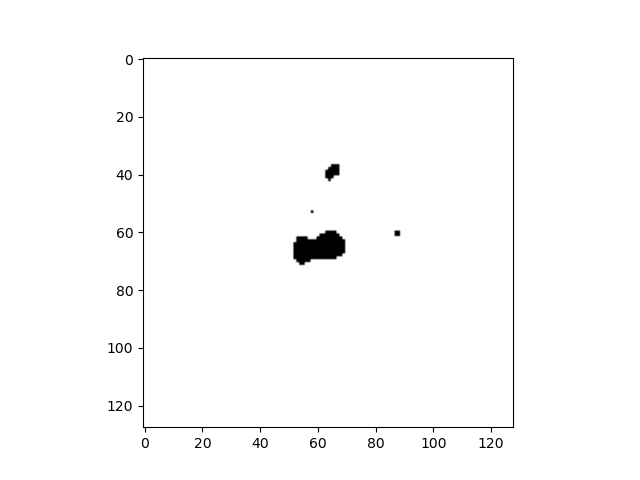}\hfill
    \includegraphics[width=0.2\textwidth]{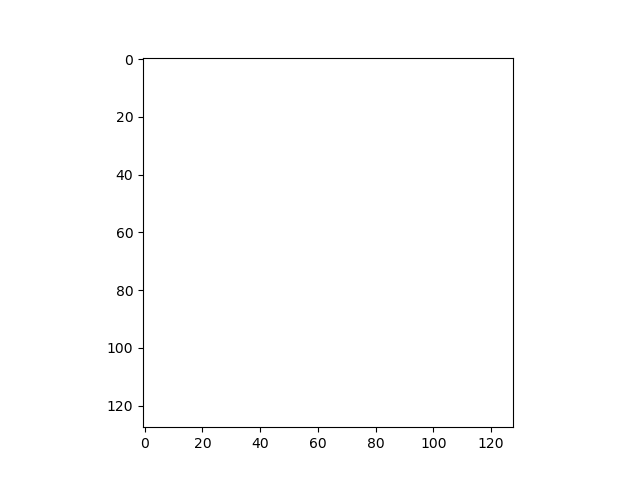}\hfill
    \includegraphics[width=0.2\textwidth]{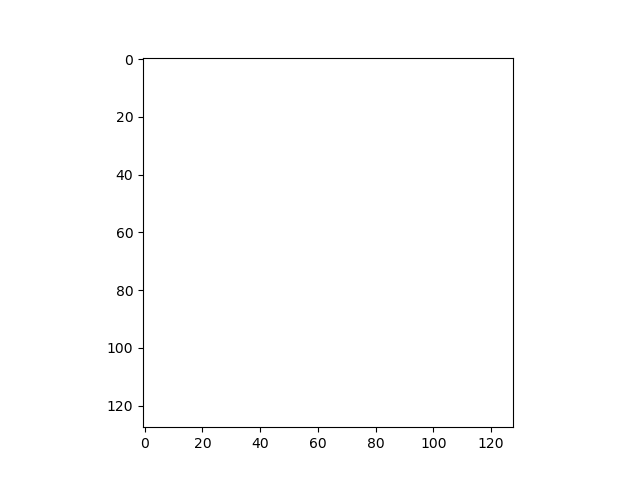}\hfill

    \hrule

    \includegraphics[width=0.2\textwidth]{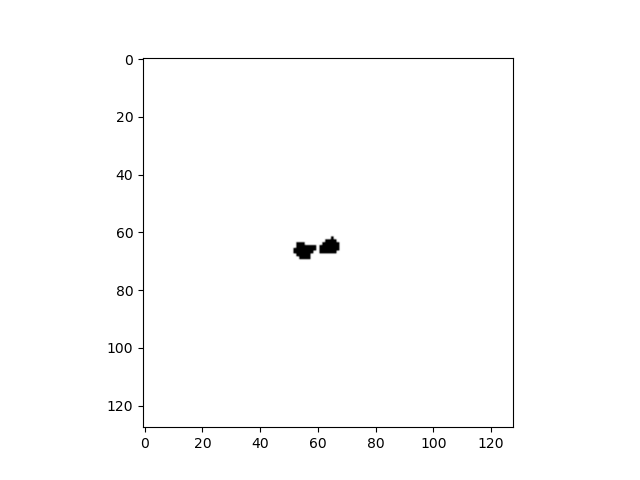}\hfill
    \includegraphics[width=0.2\textwidth]{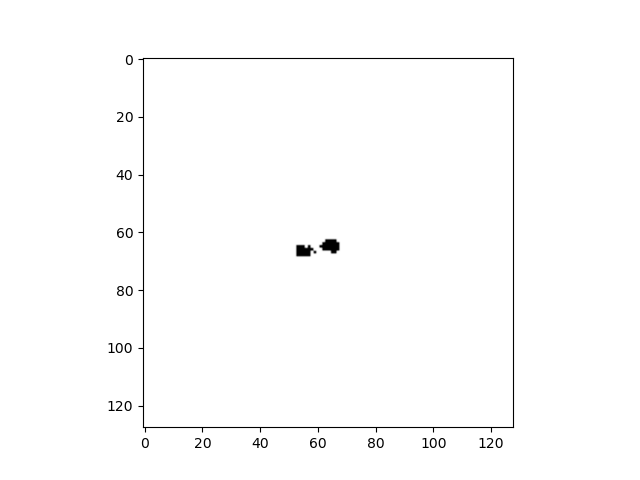}\hfill
    \includegraphics[width=0.2\textwidth]{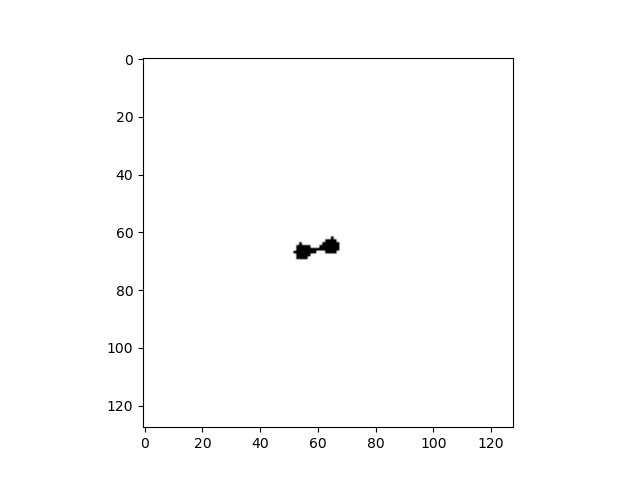}\hfill
    \includegraphics[width=0.2\textwidth]{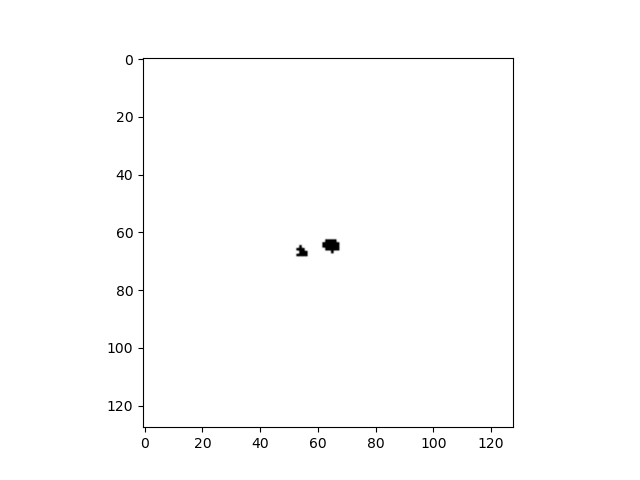}\hfill
    \includegraphics[width=0.2\textwidth]{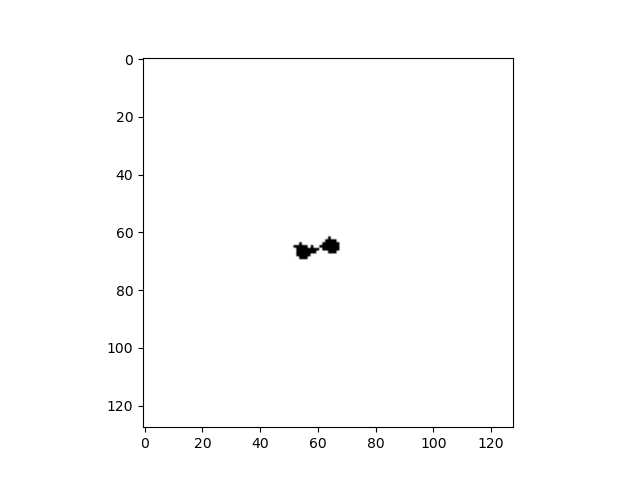}\hfill

    \includegraphics[width=0.2\textwidth]{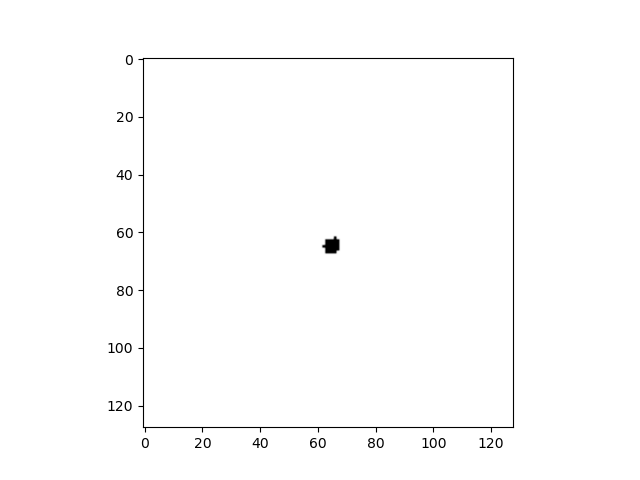}\hfill
    \includegraphics[width=0.2\textwidth]{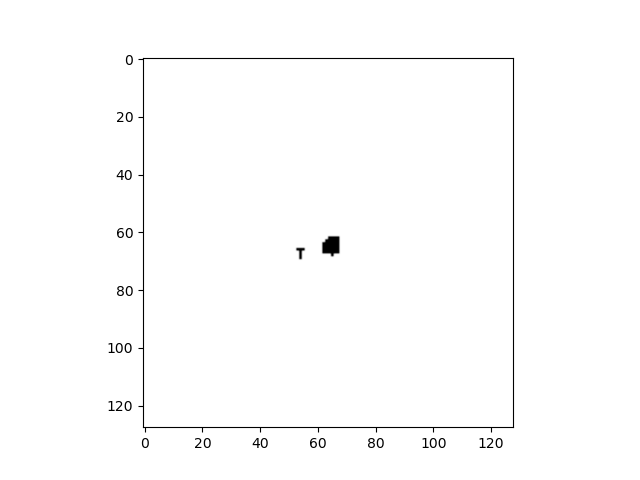}\hfill
    \includegraphics[width=0.2\textwidth]{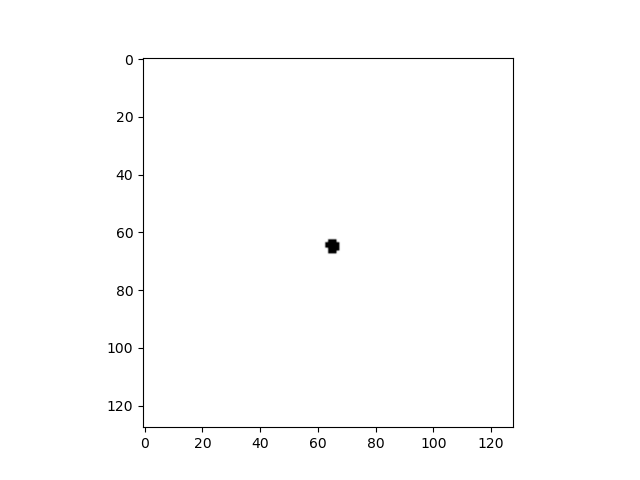}\hfill
    \includegraphics[width=0.2\textwidth]{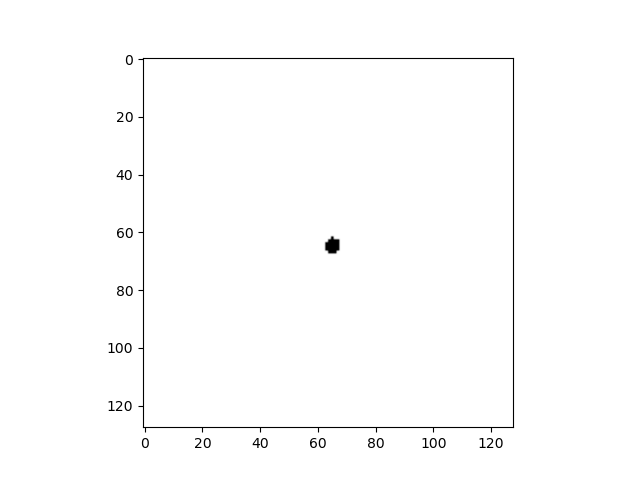}\hfill
    \includegraphics[width=0.2\textwidth]{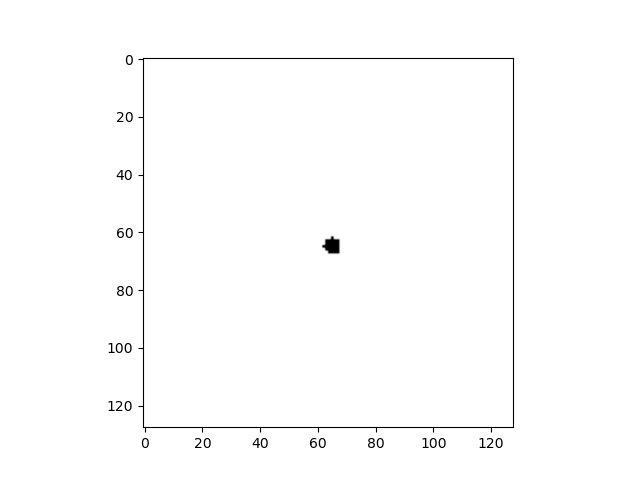}\hfill

    \includegraphics[width=0.2\textwidth]{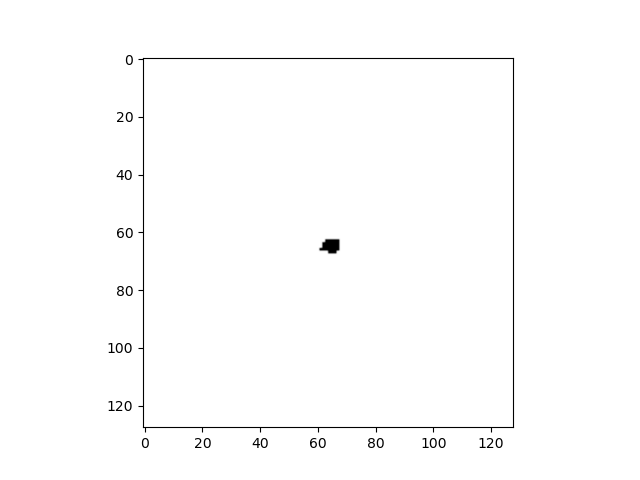}\hfill
    \includegraphics[width=0.2\textwidth]{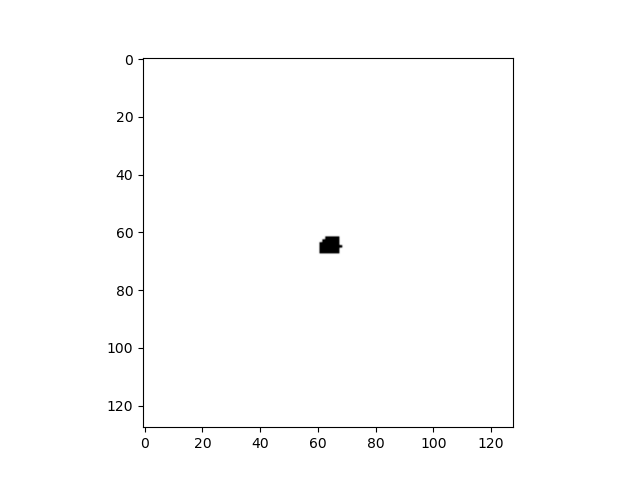}\hfill
    \includegraphics[width=0.2\textwidth]{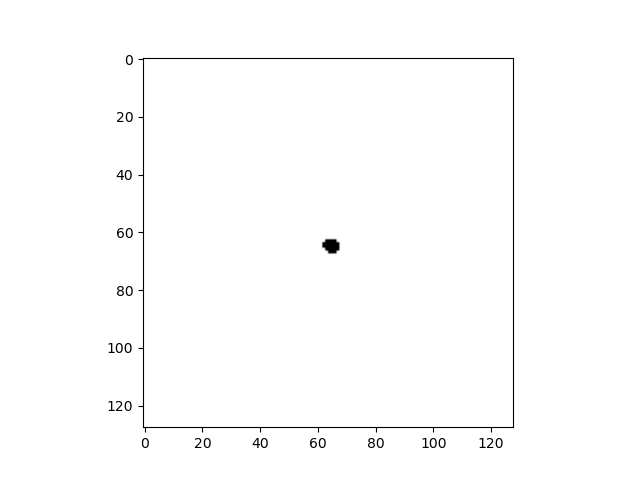}\hfill
    \includegraphics[width=0.2\textwidth]{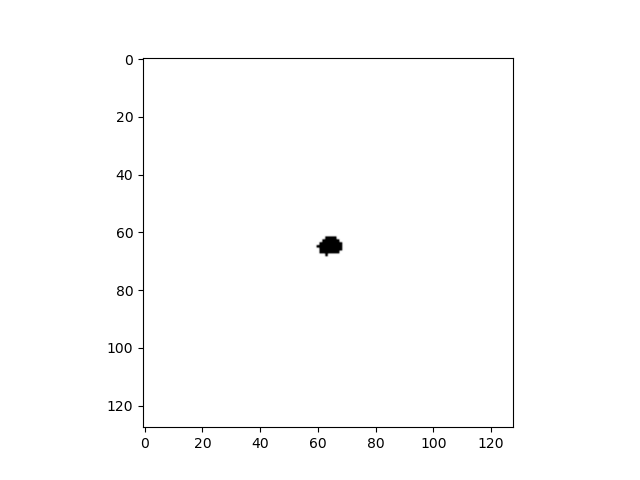}\hfill
    \includegraphics[width=0.2\textwidth]{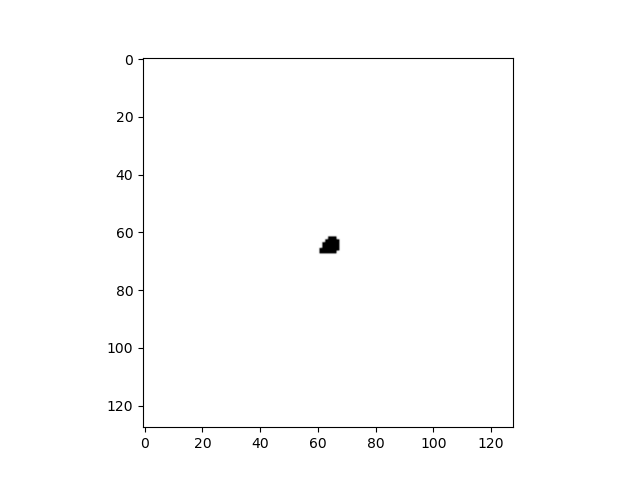}\hfill
    \caption{\textbf{Topleft}: Input image 9 from LIDC data. \textbf{Topright}: Ground truth segmentation. \textbf{2-4 rows}: Segmentation samples from original Probabilistic U-Net. \textbf{5-7 rows}: Segmentation samples from Kendall Shape Probabilistic U-Net. Each row shares the same seed.}
    \label{fig_img9}
\end{figure}

\begin{figure}
    \centering
    \includegraphics[width=0.49\textwidth]{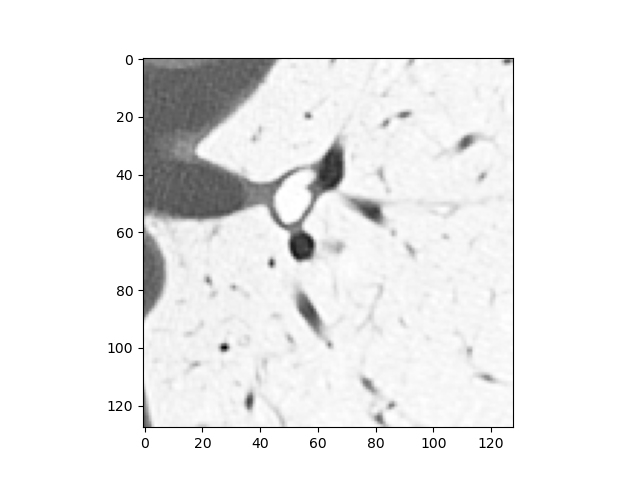}\hfill
    \includegraphics[width=0.49\textwidth]{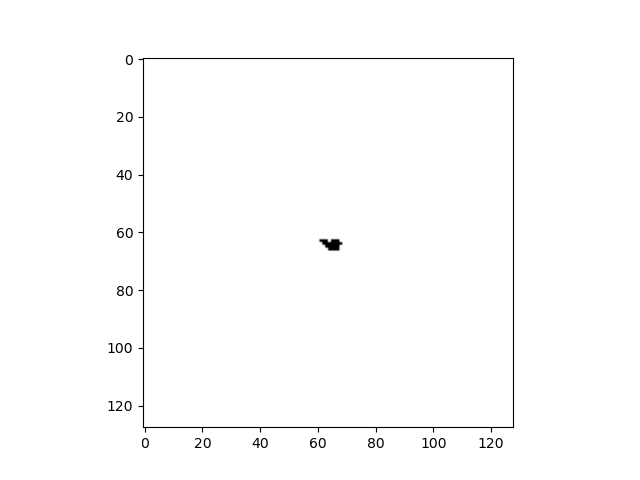}\hfill

    \hrule
    
    \includegraphics[width=0.2\textwidth]{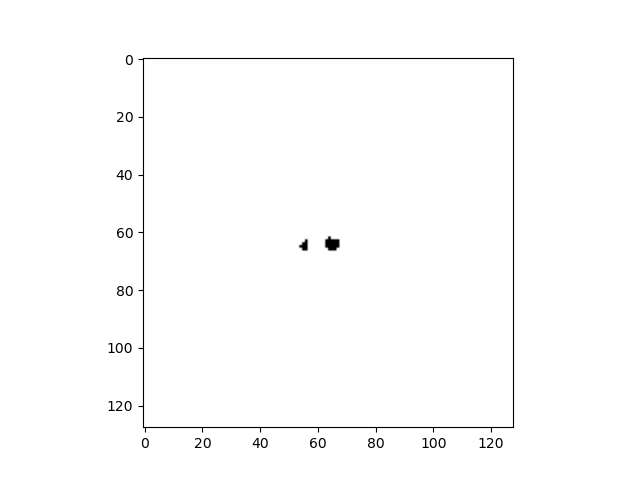}\hfill
    \includegraphics[width=0.2\textwidth]{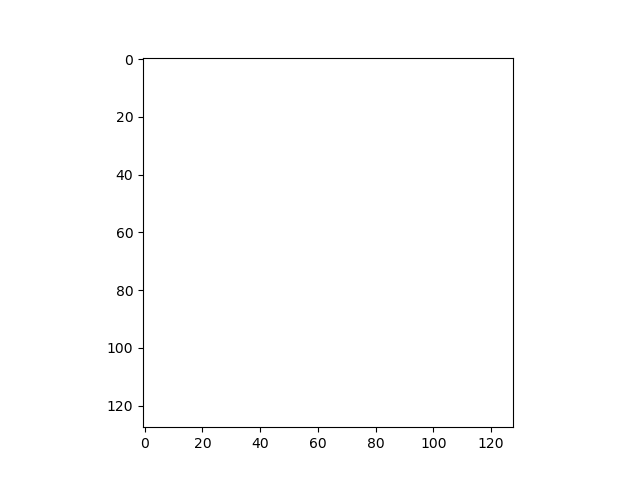}\hfill
    \includegraphics[width=0.2\textwidth]{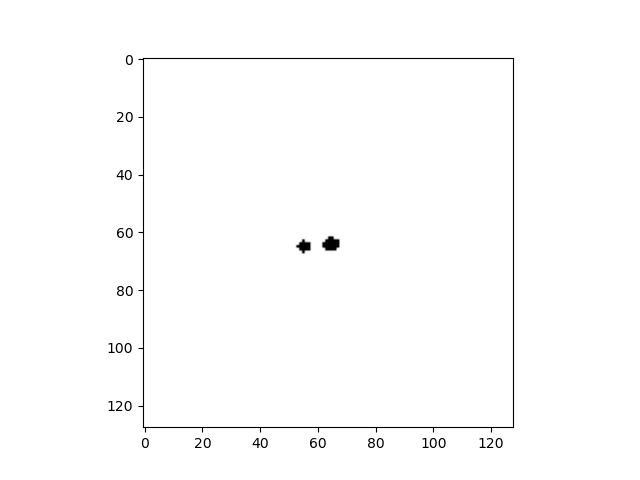}\hfill
    \includegraphics[width=0.2\textwidth]{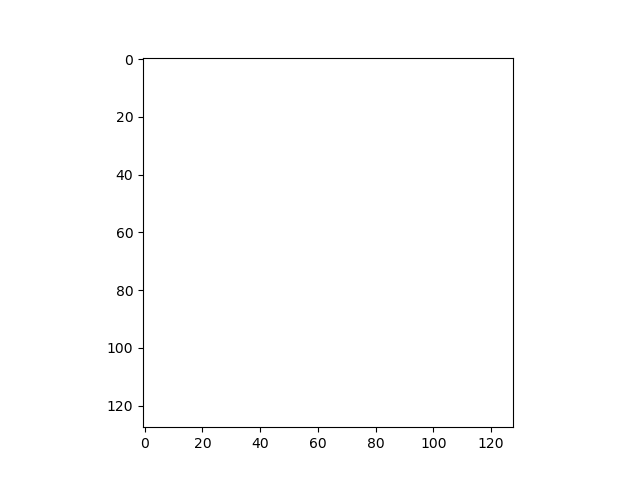}\hfill
    \includegraphics[width=0.2\textwidth]{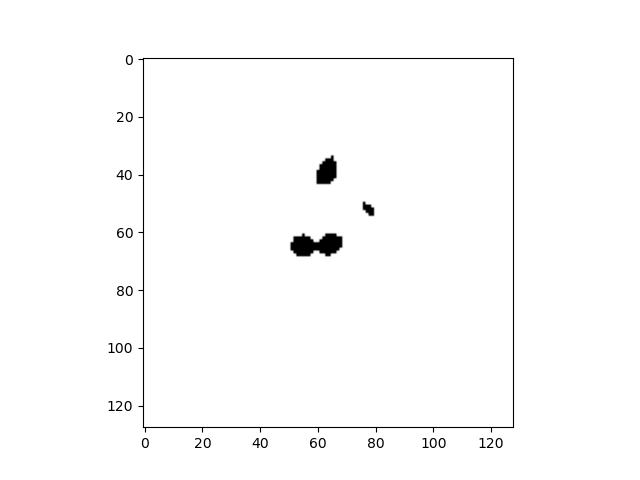}\hfill

    \includegraphics[width=0.2\textwidth]{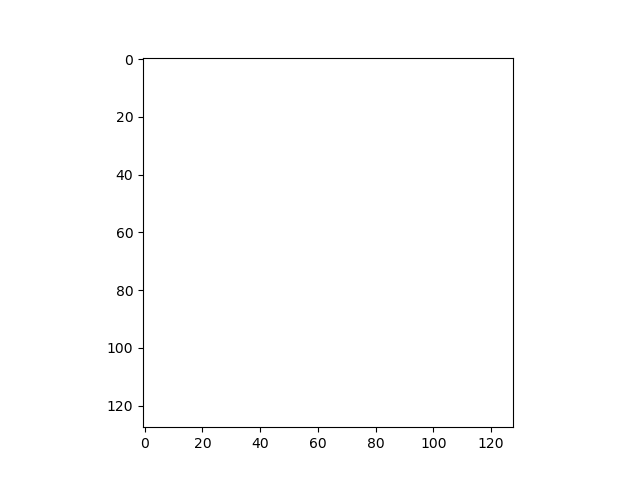}\hfill
    \includegraphics[width=0.2\textwidth]{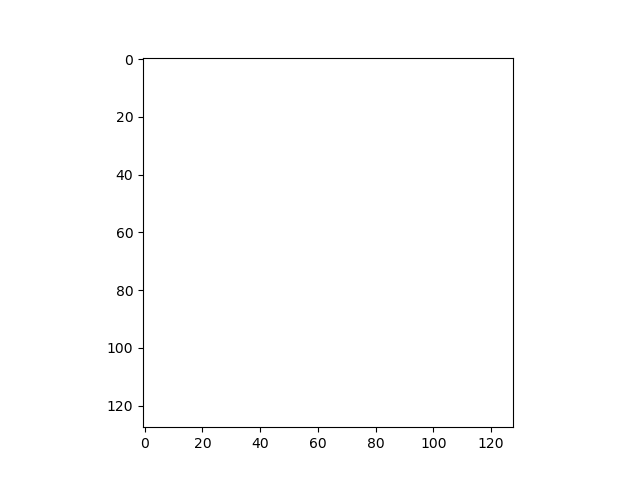}\hfill
    \includegraphics[width=0.2\textwidth]{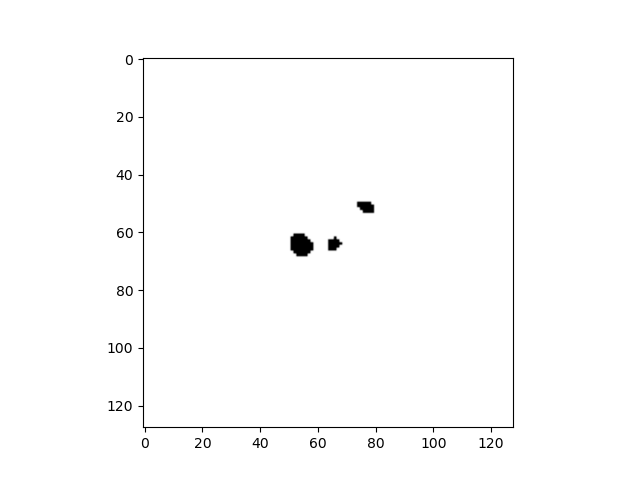}\hfill
    \includegraphics[width=0.2\textwidth]{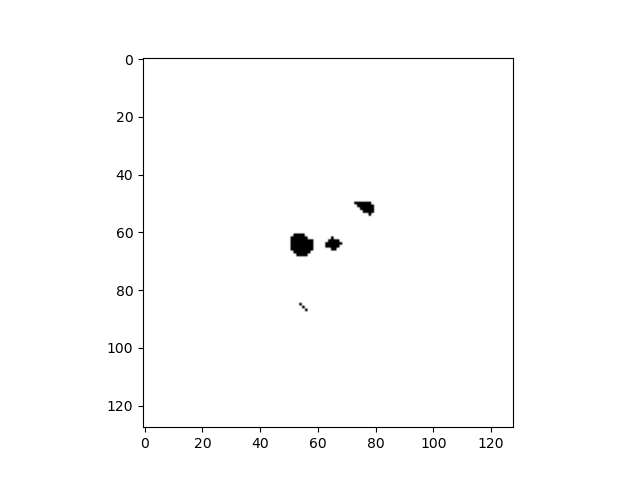}\hfill
    \includegraphics[width=0.2\textwidth]{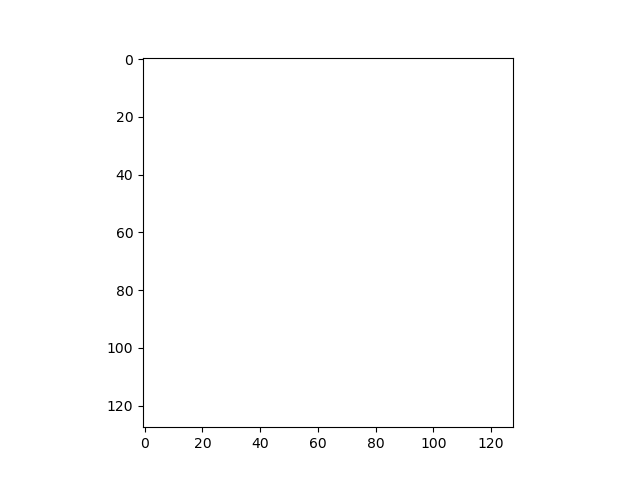}\hfill

    \includegraphics[width=0.2\textwidth]{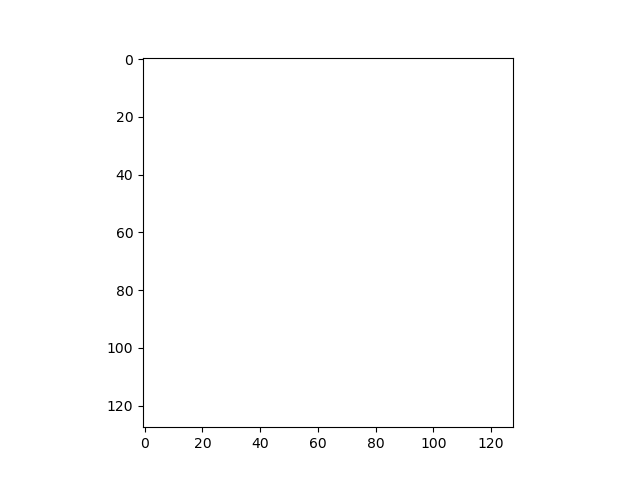}\hfill
    \includegraphics[width=0.2\textwidth]{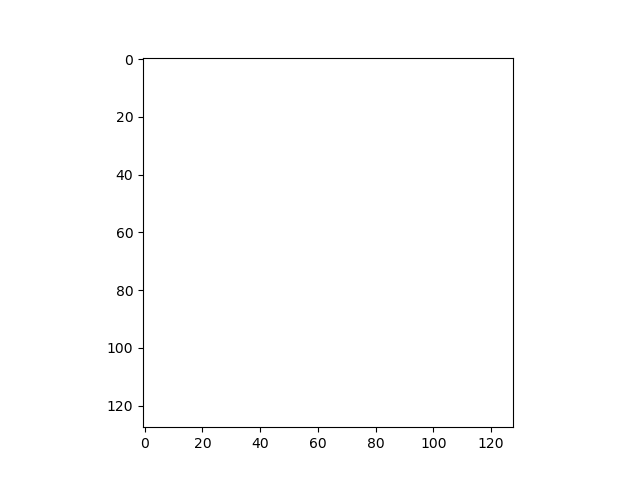}\hfill
    \includegraphics[width=0.2\textwidth]{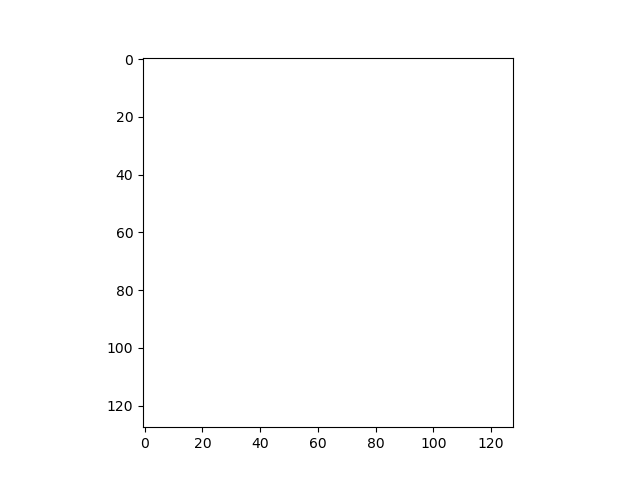}\hfill
    \includegraphics[width=0.2\textwidth]{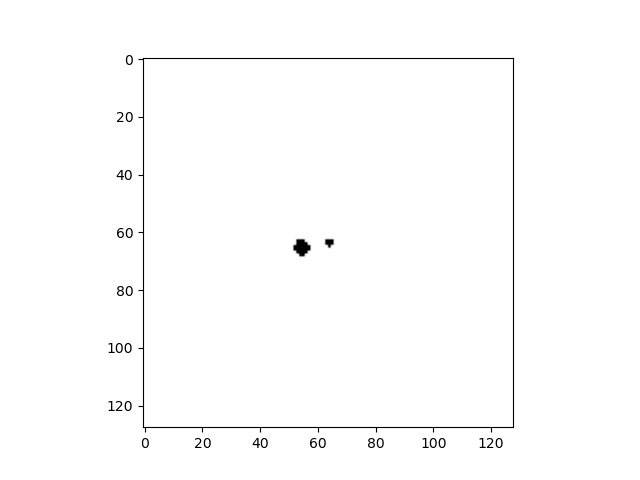}\hfill
    \includegraphics[width=0.2\textwidth]{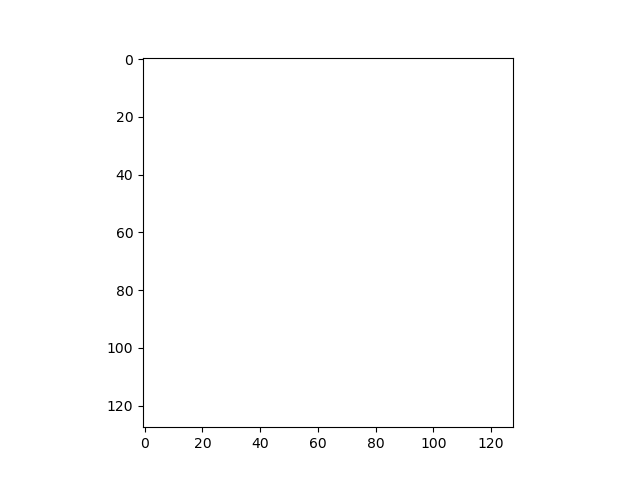}\hfill

    \hrule

    \includegraphics[width=0.2\textwidth]{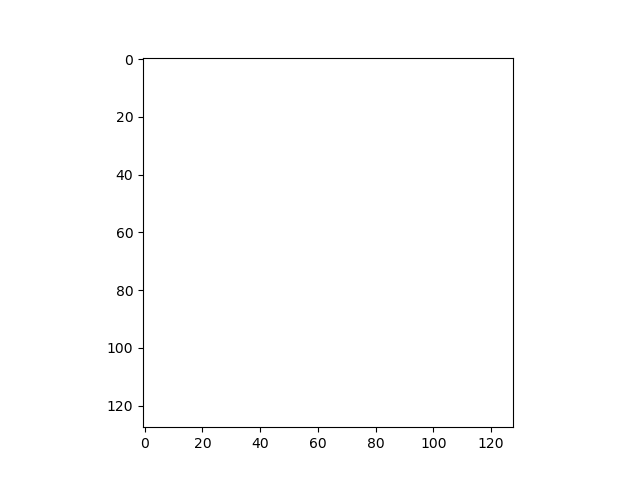}\hfill
    \includegraphics[width=0.2\textwidth]{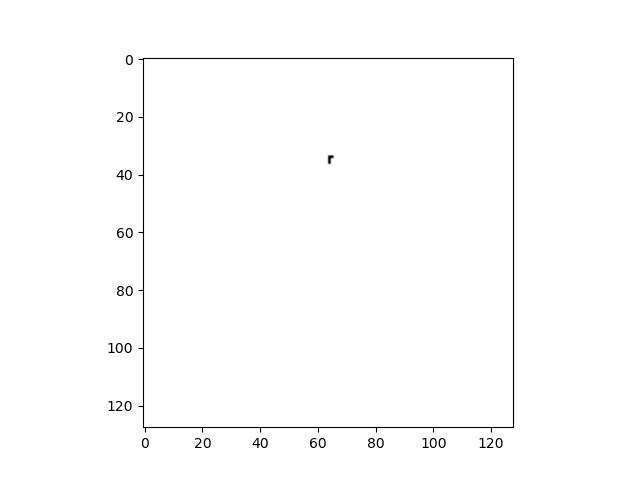}\hfill
    \includegraphics[width=0.2\textwidth]{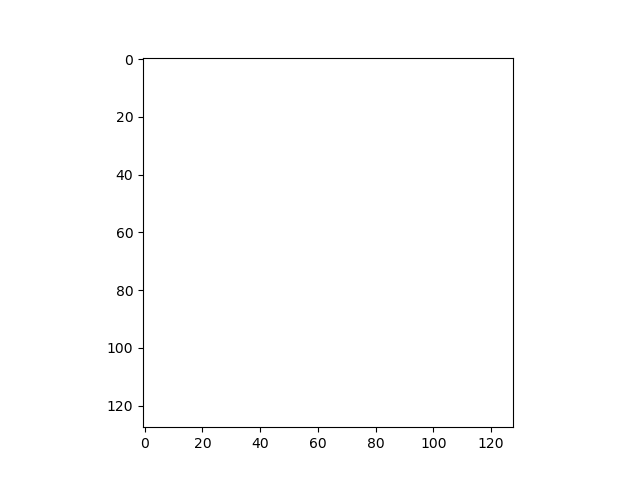}\hfill
    \includegraphics[width=0.2\textwidth]{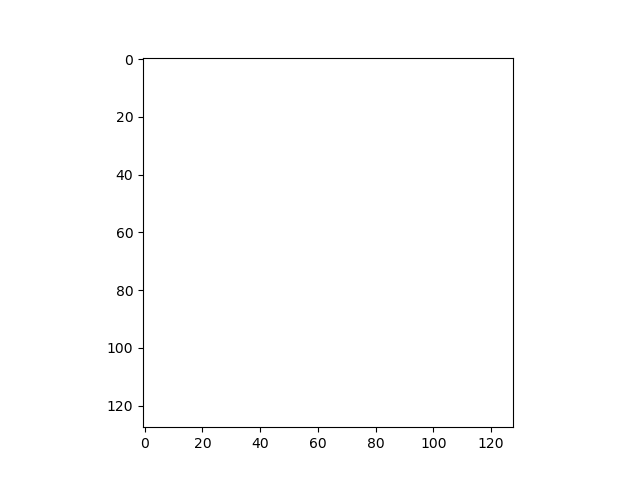}\hfill
    \includegraphics[width=0.2\textwidth]{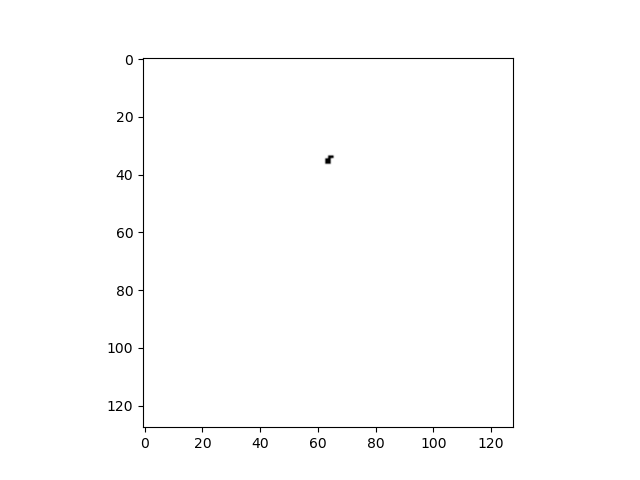}\hfill

    \includegraphics[width=0.2\textwidth]{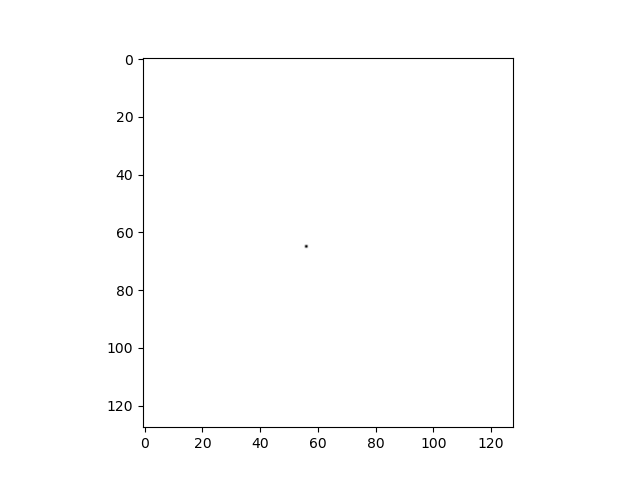}\hfill
    \includegraphics[width=0.2\textwidth]{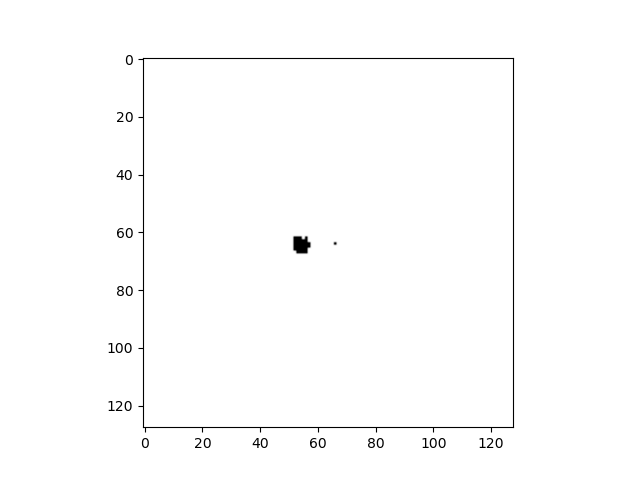}\hfill
    \includegraphics[width=0.2\textwidth]{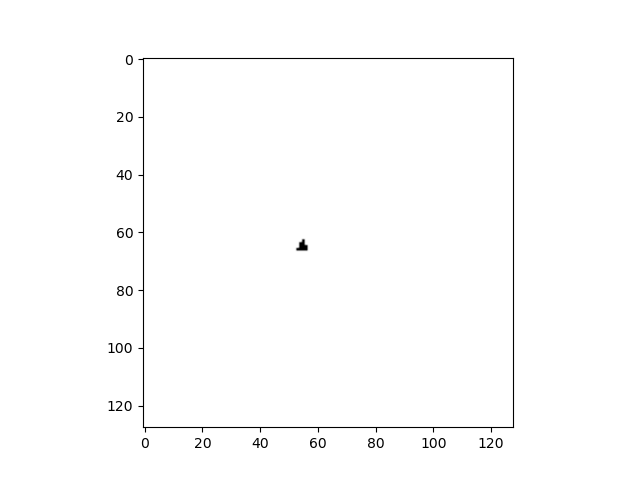}\hfill
    \includegraphics[width=0.2\textwidth]{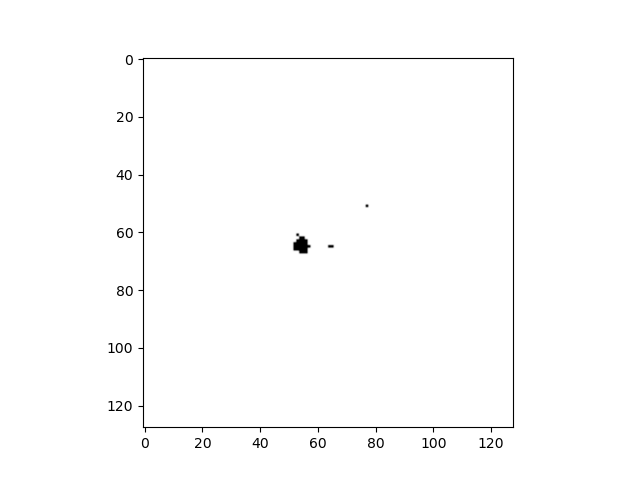}\hfill
    \includegraphics[width=0.2\textwidth]{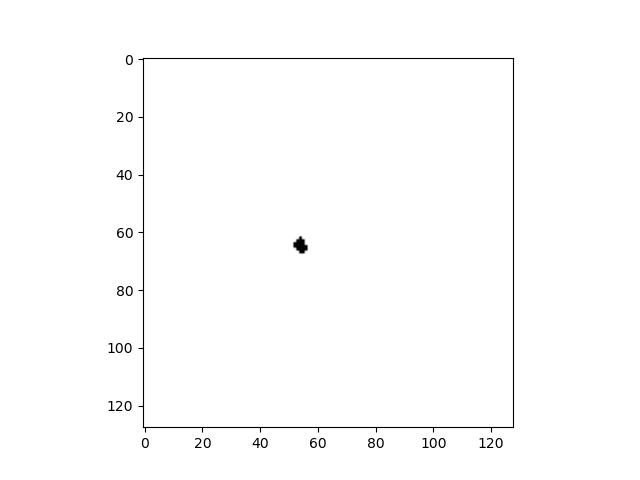}\hfill

    \includegraphics[width=0.2\textwidth]{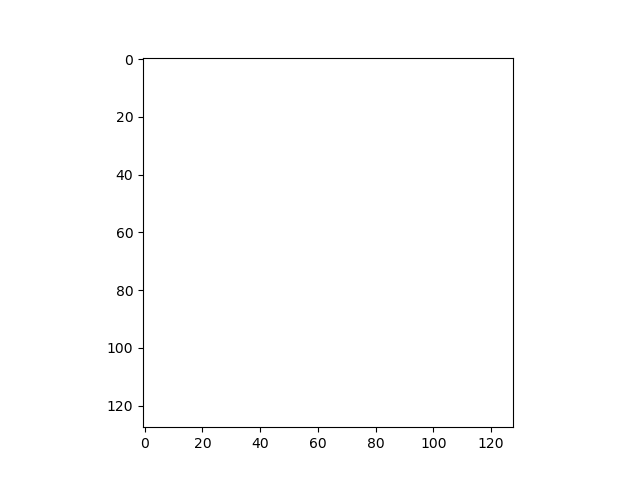}\hfill
    \includegraphics[width=0.2\textwidth]{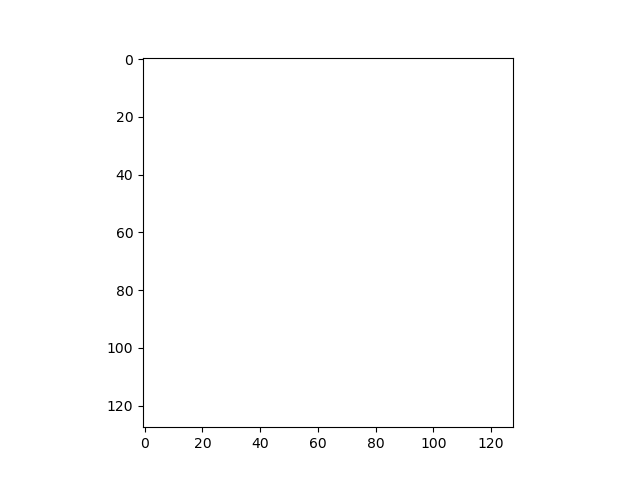}\hfill
    \includegraphics[width=0.2\textwidth]{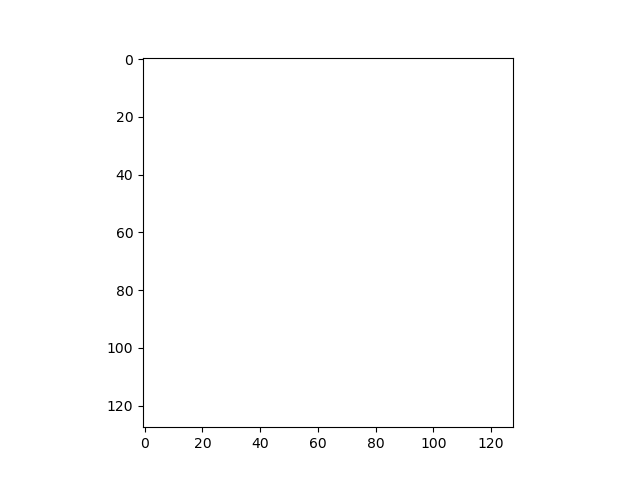}\hfill
    \includegraphics[width=0.2\textwidth]{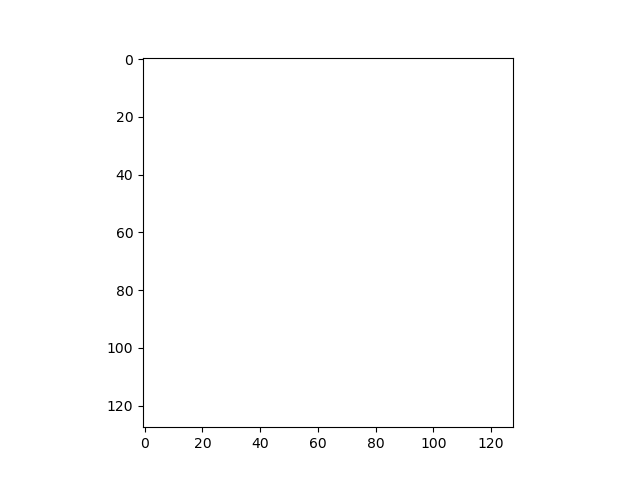}\hfill
    \includegraphics[width=0.2\textwidth]{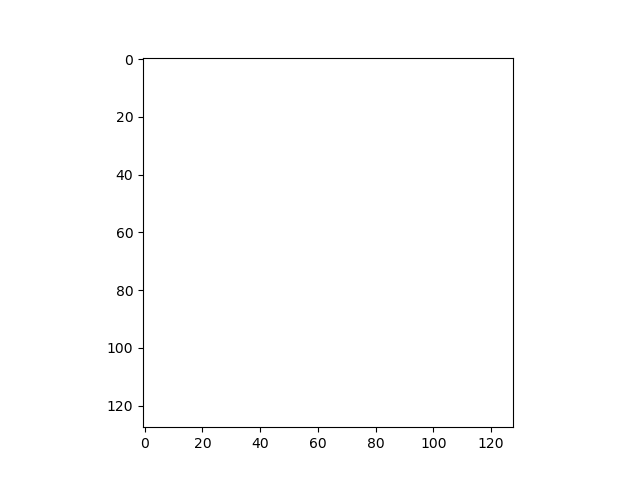}\hfill
    \caption{\textbf{Topleft}: Input image 10 from LIDC data. \textbf{Topright}: Ground truth segmentation. \textbf{2-4 rows}: Segmentation samples from original Probabilistic U-Net. \textbf{5-7 rows}: Segmentation samples from Kendall Shape Probabilistic U-Net. Each row shares the same seed.}
    \label{fig_img10}
\end{figure}

We chose 11 arbitrary input images from the LIDC dataset for illustration and comparison of the two U-Net models. We denote these input images 0 - 10. Then we produced 5 segmentation samples for each input image from both our model and the original Probabilistic U-Net, using the same hyperparameters and using the same seed for randomization. We conducted this experiment for three different random seed numbers, which resulted 15 segmentation samples for each input image. We display all the results in Figures \ref{fig_img0}, \ref{fig_img1}, \ref{fig_img2}, \ref{fig_img3}, \ref{fig_img4}, \ref{fig_img5}, \ref{fig_img6}, \ref{fig_img7}, \ref{fig_img8}, \ref{fig_img9}, and \ref{fig_img10}.
\end{document}